\pdfoutput=1
\documentclass[10pt,letterpaper]{article}
\usepackage{arxiv-style}
\usepackage{multicol}

\newcommand{\manuscripttitle}[1]{%
  ChemoMAE: Self-Supervised Spectral Representation Learning#1%
  Reveals Chemical Heterogeneity in Historical Wood}
\icmltitlerunning{\manuscripttitle{ }}
\hypersetup{
  pdftitle={\manuscripttitle{ }},
  pdfauthor={Ryuji Yamaguchi, Miyuki Matsuo}
}

\begin{document}
\twocolumn[
  \icmltitle{\manuscripttitle{\texorpdfstring{\\}{ }}}
  \begin{icmlauthorlist}
    \icmlsetsymbol{corresponding}{*}
    \icmlauthor{Ryuji Yamaguchi}{corresponding}
    \quad
    \icmlauthor{Miyuki Matsuo}{corresponding}
    \par\vspace{0.4em}
    Kyoto University
    \par\vspace{0.6em}
    {\small
      \href{https://pypi.org/project/chemomae/}{\raisebox{-0.18em}{\includegraphics[height=1.2em]{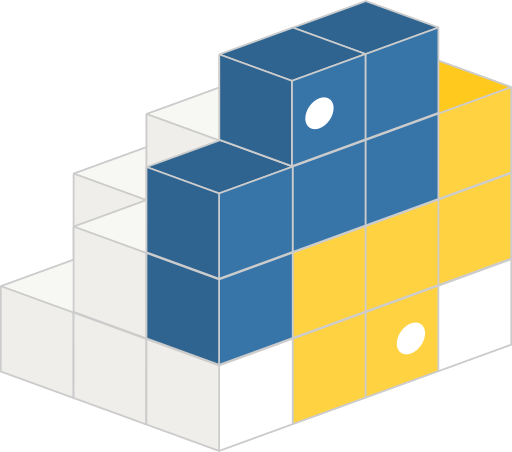}}\enspace\textbf{chemomae}}
      \qquad
      \href{https://github.com/Mantis-Ryuji/WoodDegradationMap}{\textcolor{black}{\simpleicon{github}}\enspace\textbf{WoodDegradationMap}}
    }
  \end{icmlauthorlist}
  \vskip 0.18in
  \begin{abstract}
In historical timber buildings, multiple environmental factors act over centuries on wood whose chemical composition varies across anatomical regions,
producing complex patterns of degradation.
Understanding how timber members degrade is important for the effective conservation and restoration of historical buildings,
yet the mechanisms underlying these patterns remain unclear.
To investigate these mechanisms, we used self-supervised learning to establish a common basis for comparing specimens
from timber members that differ in wood species and service history.
We examined transverse sections of 49 wood specimens from several historical buildings in Japan using near-infrared hyperspectral imaging.
ChemoMAE, a self-supervised Transformer, learns pixelwise spectral representations through masked denoising that exploits multicollinearity among spectral variables.
Clustering these representations yields spatial maps without predefined chemical-state categories.
Because anatomy and degradation can produce spatially continuous regions, we formulated Local Label Agreement (LLA)
to quantify agreement among labels assigned to neighboring pixels.
Using spectra alone, ChemoMAE yielded higher mean LLA than baseline methods on validation specimens.
We then trained ChemoMAE anew and fitted cluster centers across all specimens.
Spatial structures consistent with wood anatomy and the expected effects of surface exposure emerged in the resulting maps.
By cross-referencing regional spectra and spatial distributions across specimens in light of wood anatomy, service histories, and prior studies,
we proposed models of internal and surface degradation as mechanistic hypotheses.
This exploratory AI for Science framework gives researchers a basis for linking latent patterns learned through self-supervision
with domain knowledge when investigating how material states form.

  \end{abstract}
]

\begingroup
  \renewcommand{\thefootnote}{\fnsymbol{footnote}}
  \footnotetext[1]{\raggedright Corresponding authors:\\
    Ryuji Yamaguchi (\href{mailto:ryuji0801yama@icloud.com}{ryuji0801yama@icloud.com})\\
    Miyuki Matsuo (\href{mailto:matsuo.miyuki.4d@kyoto-u.ac.jp}{matsuo.miyuki.4d@kyoto-u.ac.jp})}
\endgroup

\section{Introduction}
\label{sec:introduction}

Understanding how timber members degrade during service is important for the conservation and restoration of historical buildings.
Wood species with different anatomical features and chemical compositions have long been used in these buildings.
Even within a single member, centuries of environmental exposure can add to chemical differences already present among anatomical regions.
During service, exposed surfaces undergo weathering through the action of sunlight and rainwater.
Sunlight preferentially degrades lignin, and rainwater can wash away soluble degradation products~\citep[pp.~274--275]{feist1990weathering}.
Decay and insect damage may also occur depending on the service environment, and a member's history of wetting and drying is often only partly known~\citep[pp.~67--68]{nilsson1990aging}.
Even in interior wood beyond the direct effects of weathering, chemical changes can occur with age~\citep{kranitz2016aging}.
Degradation can be localized within a cross-section, and mechanical properties can vary with distance from the exposed surface~\citep[pp.~105--106]{schniewind1990properties}.
Understanding how the service environment influences the type, severity, and depth of degradation could help estimate how far damage extends into a member and inform assessments of its effective load-bearing cross-section.
Such estimates matter when deciding whether a member can remain in service or requires restoration.
Point measurements can reveal local chemical differences between outer and inner wood, but leave the extent and continuity of altered regions unresolved.
Tracing these regions across a transverse section in relation to the exposed surface and wood anatomy offers clues to the processes that shaped their distribution.

To examine these distributions, we use near-infrared hyperspectral imaging (NIR-HSI) to record a spectrum at each pixel on a transverse section of each specimen.
We relate the resulting spectral variation to wood anatomy and, where an edge of the section can be linked to a surface exposed during service, examine how far the corresponding regions extend inward from that edge.
Extending the comparison to timber members that differ in wood species and service history allows us to ask which spatial patterns they share and how those patterns vary in extent.
Principal component analysis (PCA), a standard method in chemometrics, summarizes spectral variation by linear projection~\citep{wold1987pca}, which limits its ability to describe complex nonlinear relationships.
For these comparisons, we learn a common nonlinear representation of pixel spectra through self-supervised \emph{masked spectral reconstruction}, without assigning chemical-state labels in advance.
We perform unsupervised segmentation by clustering the learned spectral representations using centers shared across specimens.
Plotting each pixel's cluster label at its original location produces a spatial map showing where each cluster occurs on the transverse section.
\citet{wang2023discovery} describe how AI can help scientists uncover structure in unlabeled data and formulate hypotheses for experimental testing.
We cross-reference the spatial distribution of each cluster with its mean NIR spectrum across specimens.
We interpret shared patterns and differences in their spatial extent in light of wood anatomy and the service histories of the timber members to formulate hypotheses about the mechanisms of internal and surface degradation.

We apply this exploratory framework to 49 historical wood specimens representing seven wood taxa from several historical buildings in Japan.
To learn a common spectral representation for these specimens, we introduce ChemoMAE, a self-supervised method that uses a Transformer to learn relationships among wavelength bands in individual pixel spectra.
In NIR spectra, information associated with the same chemical bond appears in multiple bands, contributing to multicollinearity among spectral variables.
ChemoMAE exploits these dependencies through masked reconstruction and combines this task with denoising to reduce sensitivity to measurement-related variations in spectral shape.
In \emph{masked denoising}, we apply perturbations inspired by signal noise and wavelength shifts to the full spectrum before masking, then reconstruct the pre-perturbation values of the masked bands from the remaining bands.
A low-dimensional bottleneck concentrates the features used for comparison across specimens.
We design the spectral augmentations to preserve the zero mean and constant norm imposed by the standard normal variate (SNV) transformation.
We also normalize the latent representations onto the unit sphere so that cosine similarity provides a common measure in the input and latent spaces.

When no ground-truth segmentation can be specified in advance, what evidence would support using these spatial maps for chemical interpretation? Based on prior knowledge of wood anatomy and degradation, we expect regions of similar chemical state to span neighboring pixels at the scale of observation.
If the learned spectral features capture chemical differences among these regions, the resulting segmentation should reflect this spatial structure even when neither representation learning nor clustering uses spatial information.
If these features are learned across specimens, we also expect spatially coherent segmentations in unseen specimens.
We formulated Local Label Agreement (LLA) to measure agreement among neighboring cluster labels and evaluated this expectation using specimen-level cross-validation.
Across all clustering granularities examined, ChemoMAE with masked denoising yielded higher mean LLA on held-out specimens than direct use of the input spectra, PCA, a conventional autoencoder, a denoising autoencoder, or a masked autoencoder trained without additional perturbations.
When prior knowledge suggests that a material's chemical states have such spatial structure, observing spatial coherence in held-out specimens provides one basis for using the maps for chemical interpretation.
The chemical meaning of individual regions can then be examined in light of their spectra and knowledge of the material.

Following this evaluation, we learned the spectral representation and fitted the cluster centers anew using spectra drawn from all 49 specimens.
This \emph{full-dataset fitting} placed the specimens on a common basis for comparison.
Spatial structure consistent with wood anatomy and with the degradation patterns expected from the members' service environments emerged in the resulting maps.
Interpreting these distributions and the corresponding spectral features in light of the members' service histories led us to propose models of internal and surface degradation.
In the internal degradation model, aging alters the hydrogen-bonding environment around cellulose microfibrils.
Where water enters the wood, leaching can alter the composition of the remaining cell wall, and repeated wetting and drying may promote local rearrangement of cellulose chains and microfibrils.
In the surface degradation model, sunlight alters lignin in the outermost layer, and rainwater leaches soluble products.
Water carries some acidic degradation products into the wood, where those retained in cell walls may promote secondary reactions upon rewetting.
Anisotropic water transport may contribute to variation in the depth of the resulting transition region.
In this exploratory AI for Science framework, the common representation enables researchers to draw on evidence across diverse specimens when formulating hypotheses about how material states arise.

\section{Related Work}
\label{sec:related}

Studies of historical wood have generally reported a decline in hemicellulose content with age, although findings on cellulose crystallinity vary~\citep{kranitz2016aging}.
In a poplar member from a historic Chinese temple, Guo et al. examined the wood from the surface inward using imaging FT-IR spectroscopy and complementary measurements.
They found lower cellulose crystallinity and higher equilibrium moisture content in the historical wood than in the reference poplar~\citep{guo2018ageing}.
In wood preserved under dry conditions in a Japanese temple, NIR spectroscopy with deuterium exchange suggested a loss of less ordered cellulose relative to modern wood of the same species~\citep{tsuchikawa2005ageing}.
Controlled irradiation experiments have also characterized chemical changes near exposed surfaces.
Ultraviolet irradiation degraded aromatic structures in lignin and produced oxidation products that could be extracted with water~\citep{tolvaj1995photodegradation}.
In Japanese cedar earlywood exposed to simulated sunlight for 1,500~h, FT-IR depth profiles showed increased carbonyl absorption as far as 600--700~$\mu$m below the exposed surface~\citep{kataoka2001depth}.

Hyperspectral imaging has also been used to relate spectral variation in wood to known exposure conditions.
Agresti et al. applied PCA to hyperspectral images of artificially irradiated poplar surfaces and found that their shortwave infrared spectra varied with irradiation time~\citep{agresti2013photodegraded}.
Reconstruction-based methods offer another way to extract features from unlabeled spectra.
Denoising autoencoders learn by reconstructing observations from corrupted inputs, while masked autoencoders (MAEs) learn by predicting withheld portions of the input~\citep{vincent2010stacked,he2022mae}.
In Raman spectroscopy, Jensen et al. trained a variational autoencoder to reconstruct spectra from perturbed inputs, and Ren et al. used masked spectral reconstruction in SMAE to learn features for clustering and classifying bacterial spectra~\citep{jensen2024ramanssl,ren2025smae}.
Learned spectral representations have also supported spatial interpretation and comparison across datasets.
GyPSUM clustered representations of NIR pixel spectra and used cluster mean spectra and expert knowledge to interpret mineral distributions~\citep{gao2021gypsum}.
RamanMAE mapped subcellular compartments by clustering spectral representations learned through masked reconstruction~\citep{paidi2025ramanmae}.
LeafVAE applied a trained encoder and fixed cluster centers to independent corn leaf datasets, allowing spectral signatures to be compared across measurement conditions~\citep{ji2026leafvae}.

Bench et al. asked whether tissue regions containing the same constituents could be distinguished by the spatial arrangement of their spectra.
They constructed synthetic Raman images from measured spectra of porcine muscle and fat, with fat arranged in stripes in some regions and in scattered globules in others.
A convolutional autoencoder trained on image patches learned features that distinguished these arrangements~\citep{bench2022unsupervised}.
Our study starts from a hypothesis about chemical heterogeneity in historical wood.
Based on prior knowledge of wood anatomy and degradation, we expect regions of similar chemical state to span neighboring pixels on the measured transverse sections.
If a spectral representation captures differences between such regions across specimens, clustering should yield spatially coherent segmentations on unseen specimens even when pixel positions and neighborhood relations are excluded from both representation learning and clustering.
Bench et al. used spatial information to learn representations that distinguish tissue regions.
We use spatial information to evaluate representations learned from spectra alone.
In this hypothesis-driven evaluation, we use LLA to assess whether segmentations of held-out specimens exhibit the predicted spatial coherence.

Studies of wood aging, water exposure, and weathering have characterized changes in cell-wall chemistry and structure under a range of conditions~\citep{bardet2002archaeological,tsuchikawa2005ageing,kanbayashi2018weathering}.
Work in hyperspectral imaging has also shown how learned spectral representations can support expert interpretation of spatial maps and comparison across specimens~\citep{gao2021gypsum,ji2026leafvae}.
We bring these lines of work together around a question for hyperspectral studies of materials.
When ground-truth labels for material states are scarce, how can researchers use a learned representation to connect observations across diverse specimens with findings from prior studies and formulate hypotheses about how those states arose?
We examine this question in historical timber members that differ in wood species and service history.
To assess the basis for using the maps for chemical interpretation, we compare ChemoMAE with PCA and alternative autoencoders on held-out specimens, using the spatial coherence expected from wood anatomy and degradation as our criterion.
We then learn a shared spectral representation and fit a common set of cluster centers using spectra from all specimens.
An interpretation suggested by a cluster's mean NIR spectrum and its location in one specimen can thus be examined against the mean spectrum, location, and extent of the same cluster in other specimens.
We use the shared patterns and differences across specimens to ask how processes documented under different conditions might be linked in historical wood.
These comparisons lead us to propose models of internal and surface degradation.

\section{Materials and Data Preparation}
\label{sec:materials}

\begin{figure*}[!t]
\centering
\includegraphics[width=.85\textwidth]{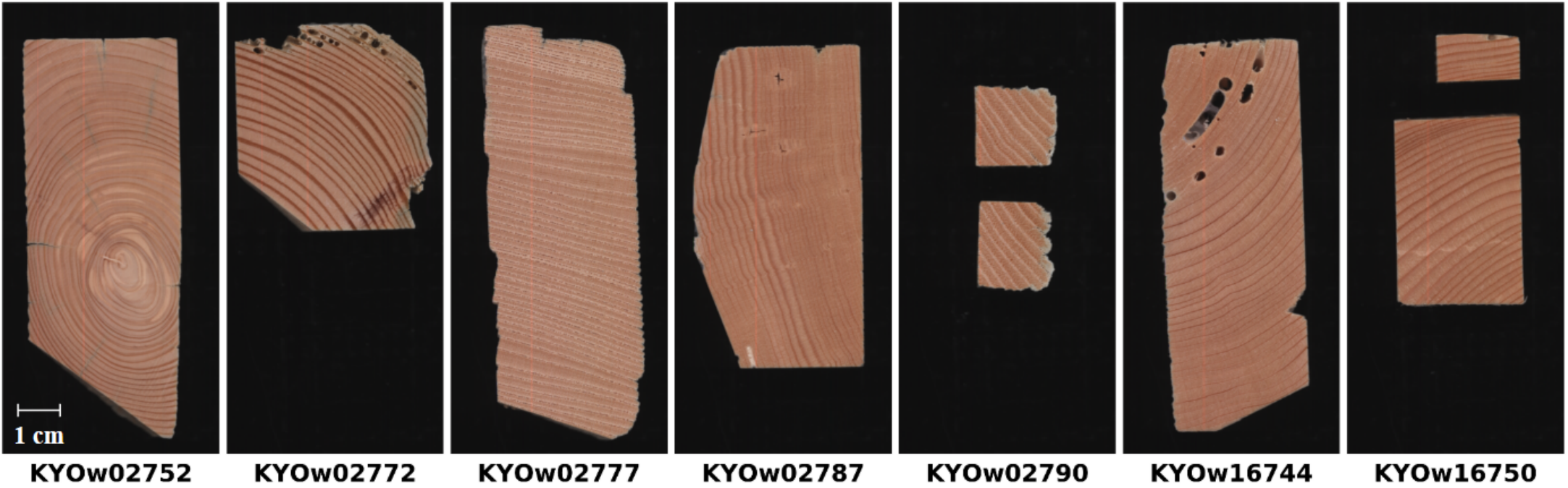}
\caption{Pseudo-RGB images of the measured surfaces of seven specimens, one per taxon.
From left to right: \textit{Chamaecyparis obtusa}, \textit{Pinus} sp., \textit{Zelkova serrata}, \textit{Tsuga sieboldii}, \textit{Castanea crenata}, \textit{Cryptomeria japonica}, and \textit{Abies} sp.
Specimen IDs are shown below each image.}
\label{fig:specimen-surfaces}
\end{figure*}

\subsection{Historical Wood Specimens}
\label{sec:specimens}
We studied 49 historical wood specimens taken from timber members in several historical buildings.
The specimens represented seven wood taxa, five identified to species and two to genus: 14 \textit{Chamaecyparis obtusa}, 23 \textit{Pinus} sp., four \textit{Zelkova serrata},
one \textit{Tsuga sieboldii}, three \textit{Castanea crenata}, three \textit{Cryptomeria japonica}, and one \textit{Abies} sp.
Each specimen was identified by an ID beginning with KYOw;
the associated site, building, timber member type, and wood taxon are listed in Appendix~\ref{app:specimen-metadata}.
The combination of these attributes differed for most specimens.
Neither taxonomic identity nor member type served as a supervisory label for chemical state.

For specimens with matching attributes, visual inspection of the original timber members revealed differences in surface characteristics and degradation state,
from which we inferred differences in the service environments of the sampled locations.
Even within a single member, exposure may have varied with location, for example between the outdoor-facing and indoor-facing sides of a column.
However, no records establish whether these specimens came from the same member
or how they were positioned relative to each other during service.

The specimens were approximately 10\,mm thick. We imaged a transverse section of each specimen.
Where an edge of the section could be linked to a side face exposed during service, we used that edge as a reference for assessing how far a region extended inward within the section.
The specimens were stored at approximately 25\,$^\circ$C and 60\% relative humidity and measured in an air-dry state.
The degree of sanding and surface finishing was not standardized across specimens;
the measured surfaces included untreated surfaces and surfaces with traces of insect damage.

Before examining the analytical results, we selected one illustrative specimen per taxon (Figure~\ref{fig:specimen-surfaces}).
The pseudo-RGB images provide an overview of the shape and anatomical structure of the measured surfaces.
Pseudo-RGB images of all 49 specimens are shown in Appendix~\ref{app:specimen-overviews} (Figure~\ref{fig:all-specimens-pseudo-rgb}).

\subsection{NIR-HSI Measurement}
\label{sec:nir-measurement}
Near-infrared hyperspectral images were acquired using a line-scan Compovision system
(CV-N800-HS, Sumitomo Electric Industries) equipped with an InGaAs sensor.
Two halogen lamps, one on each side, provided illumination.
The measured surfaces of the specimens were visually adjusted to a common height,
and the specimens were scanned on a translation stage at a constant speed.

Measurement conditions were the same for all 49 specimens:
a frame rate of 200\,Hz, an exposure time of 4.5\,ms, and a stage scanning speed of 37.53\,mm/s.
Each line comprised 320 spatial pixels and 256 wavelength bands.
The acquired wavelength range was 913.10--2518.47\,nm,
with intervals of 6.25--6.34\,nm between adjacent bands.
The spatial sampling interval, calculated from the approximately 60\,mm field of view and the scanning speed,
was approximately 0.19\,mm in both directions.

Before measurement, a single line was acquired for each of the white and dark references
using the same frame rate and exposure time as for the specimens.
A barium sulfate (BaSO$_4$) white plate was used for the white reference and a lens cap for the dark reference;
neither reference acquisition involved scanning.

\subsection{Data Preprocessing}
\label{sec:preprocessing}
During preprocessing, specimen masks were used to extract wood regions from the near-infrared hyperspectral images.
Reflectance conversion, wavelength-range selection, resampling
and the standard normal variate (SNV) transformation~\citep{barnes1989snv} were then applied sequentially to each pixel spectrum.
These fixed preprocessing steps were common to all specimens and all comparison conditions.
Excluded pixels were displayed as background.

In the following, $(u,v)$ denotes the row and column of an image, $C=256$ is the number of measured bands,
and $c\in\{1,\ldots,C\}$ indexes the bands.
The intensity spectrum at pixel $(u,v)$ is denoted by
$\mathbf{i}_{u,v}=(i_{u,v,1},\ldots,i_{u,v,C})^{\mathsf{T}}
\in\mathbb{R}^{C}$.

\subsubsection{Specimen Mask}
For each specimen, multi-Otsu thresholding was applied to the $\ell_1$ norms of the pixel intensity spectra $\mathbf{i}_{u,v}$
to partition the pixels into three classes.
The class with the lowest values was treated as background, and the other two classes as candidate wood regions.
To reduce the influence of pixels containing mixtures of wood and background,
the candidate regions were eroded inward by one pixel to obtain the specimen mask.

\subsubsection{Reflectance and Wavelength Range}
The measured intensities were converted to reflectance using the white and dark references.
Let $\mathbf{w}_{v},\mathbf{d}_{v}\in\mathbb{R}^{C}$ denote the white- and dark-reference spectra, respectively,
corresponding to image column $v$.
The reflectance spectrum $\mathbf{r}_{u,v}$ was calculated as
\begin{equation}
  \mathbf{r}_{u,v}
  =
  (\mathbf{i}_{u,v}-\mathbf{d}_{v})
  \oslash
  (\mathbf{w}_{v}-\mathbf{d}_{v})
  \label{eq:reflectance}
\end{equation}
where $\oslash$ denotes elementwise division.
For both references, the values for each column were applied to all rows in that column.

We used the ratio of the mean reference difference to its standard deviation across columns
as a proxy for the stability of the reflectance conversion,
and excluded the low-ratio region at the long-wavelength end (Appendix~\ref{app:preprocessing-details}).
Of the 256 acquired bands, we retained 222 bands spanning 913.10--2305.59\,nm
and linearly interpolated them onto 256 equally spaced points, including both endpoints of the same wavelength range,
to obtain $\tilde{\mathbf{r}}_{u,v}\in\mathbb{R}^{C}$.
Hereafter, $C=256$ denotes the number of interpolated points rather than the number of measured bands.
Pixels were excluded if a reference difference in a retained band was nonpositive or nonfinite,
if a reflectance value was nonfinite, or if the interpolated reflectance spectrum contained a negative value.

\subsubsection{Standard Normal Variate Transformation}
We used the SNV transformation as a common preprocessing step to compare relative spectral shapes~\citep{barnes1989snv}.
This transformation can be applied independently to each pixel without a reference spectrum or tuning parameters,
and removes differences in offsets and positive scaling that are common to all bands.

Let $\mathbf{1}_{C}\in\mathbb{R}^{C}$ be the all-ones vector.
We define the mean $\mu_{u,v}$ and sample standard deviation $s_{u,v}$ across wavelengths as
\begin{equation}
  \begin{aligned}
    \mu_{u,v}
    &= \frac{1}{C}\mathbf{1}_{C}^{\mathsf{T}}
       \tilde{\mathbf{r}}_{u,v}, \\
    s_{u,v}
    &= \sqrt{\frac{1}{C-1}
       \left\|
         \tilde{\mathbf{r}}_{u,v}-\mu_{u,v}\mathbf{1}_{C}
       \right\|_{2}^{2}}
  \end{aligned}
  \label{eq:snv_statistics}
\end{equation}
and calculate the input spectrum after the SNV transformation as
\begin{equation}
  \mathbf{x}_{u,v}
  =
  \frac{
    \tilde{\mathbf{r}}_{u,v}-\mu_{u,v}\mathbf{1}_{C}
  }{s_{u,v}}
  \label{eq:snv}
\end{equation}
Pixels with nonpositive or nonfinite $s_{u,v}$ were excluded from the analysis because the SNV transformation was undefined for them.
After transformation, the mean across wavelengths is zero and the sample variance is one;
apart from numerical error, the input spectra satisfy
\begin{equation}
  \mathbf{1}_{C}^{\mathsf{T}}\mathbf{x}_{u,v}=0,
  \qquad
  \|\mathbf{x}_{u,v}\|_{2}=\sqrt{C-1}
  \label{eq:snv_constraints}
\end{equation}
In Section~\ref{sec:augmentation}, the set of vectors satisfying these two equations is used to design spectral augmentations.

Pixels within the specimen mask that remained after these steps were defined as valid pixels,
and their set is denoted by $\Omega$.
The input spectrum $\mathbf{x}_{u,v}$ at each $(u,v)\in\Omega$ was associated with its original image coordinates for subsequent analyses.

\section{Representation Learning and Unsupervised Segmentation}
\label{sec:method}

\subsection{Overview}
\label{sec:method-overview}
ChemoMAE takes as input the spectrum $\mathbf{x}_{u,v}$ after SNV transformation
at each pixel $(u,v)$ in the set of valid pixels $\Omega$ defined in Section~\ref{sec:preprocessing}.
Hereafter, we omit the pixel indices and write $\mathbf{x}$ unless otherwise specified.

\begin{figure*}[t]
  \centering
  \includegraphics[height=.43\paperheight]{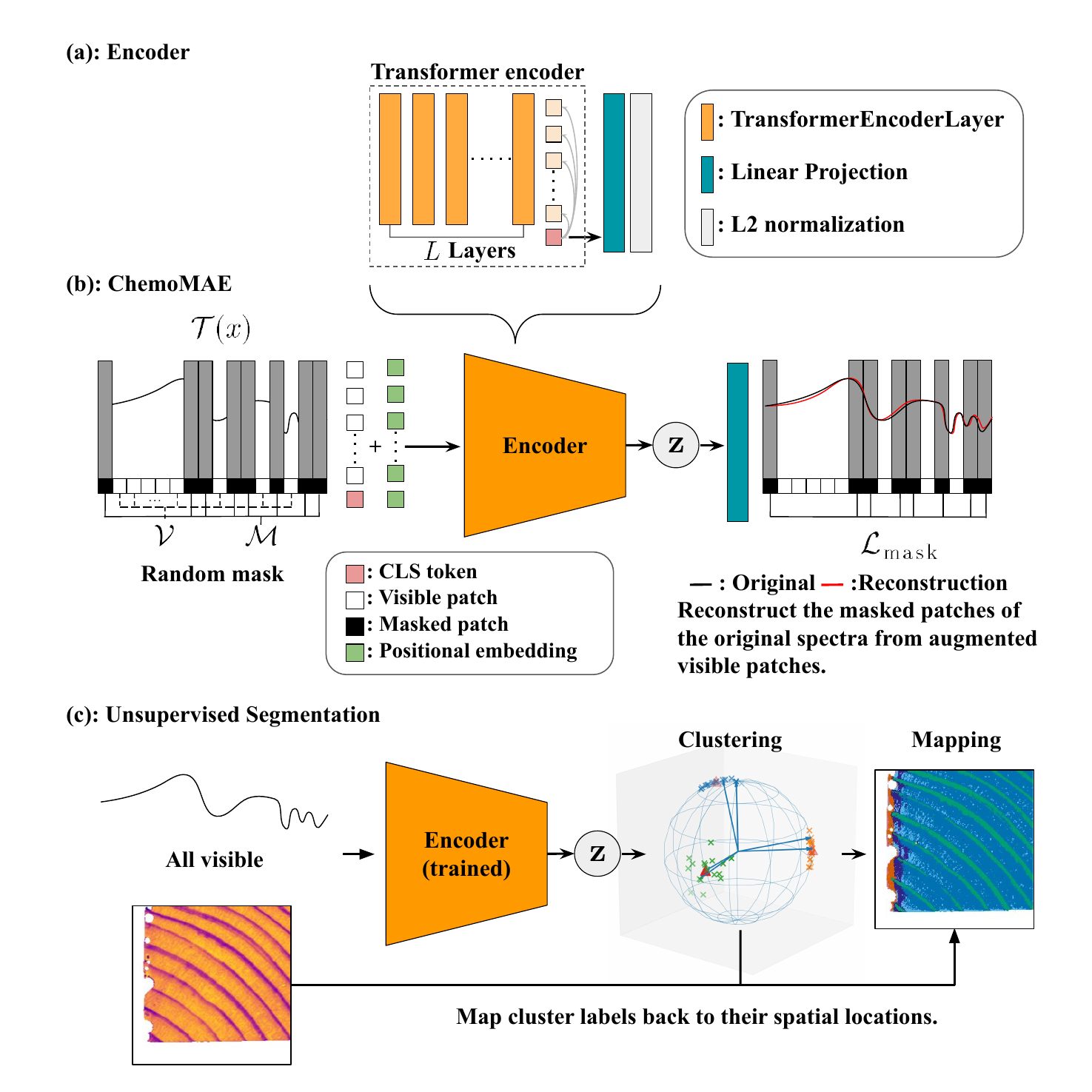}
  \caption{ChemoMAE architecture and analysis workflow.
  (a) Projection and normalization of the CLS-token output.
  (b) Training by masked denoising.
  (c) Unsupervised segmentation based on the learned representations.}
  \label{fig:chemomae-overview}
\end{figure*}

For methods that learn agreement between augmented views,
augmentation design must reflect the information to be preserved~\citep{chen2020simclr}.
When spectral differences related to material state arise from unknown or combined factors,
the desired invariances are difficult to specify in advance.
We therefore adopted reconstruction-based learning, using the observed spectra as reconstruction targets.
The model learns interband dependencies through input reconstruction rather than directly enforcing agreement between latent representations.
Building on denoising autoencoders~\citep{vincent2010stacked} and MAEs~\citep{he2022mae},
we use masked denoising~\citep{wu2023dmae} as the pretext task.

We refer to the input transformations produced by Tangent Gaussian Noise (TGN) and Fractional Shift (FS) as additional perturbations.
In the standard configuration, we apply any selected perturbations to the full spectrum before masking wavelength patches.
The model then predicts the pre-perturbation values of the masked bands from the remaining bands.
The perturbations are intended to reduce sensitivity to spectral fluctuations that could obscure chemical differences between regions.
We clustered the learned representations and mapped the labels back to pixel coordinates to construct spatial maps.
We examine how the spatial coherence of segmentations on held-out specimens changes when training shifts from full-spectrum reconstruction to prediction of masked bands.
We also assess how additional spectral perturbations affect spatial coherence under each reconstruction task (Section~\ref{sec:comparison-design}).

\subsection{ChemoMAE}
\label{sec:chemomae}

Like dimensionality reduction by PCA~\citep{wold1987pca},
ChemoMAE constructs low-dimensional representations for exploring and comparing spectral differences.
The bottleneck maps the final output of the encoder's classification (CLS) token to $d$ dimensions with an affine layer
and normalizes it to give one unit-length representation per spectrum.

The channels of the input spectrum $\mathbf{x}$ are divided into $N$ patches,
$\mathcal{P}_1,\ldots,\mathcal{P}_N$,
each containing the same number of contiguous channels along the wavelength axis.
Let $\mathcal{M}\subset\{1,\ldots,N\}$ denote the set of masked patch indices
and $\mathcal{V}=\{1,\ldots,N\}\setminus\mathcal{M}$ the set of unmasked patch indices.
A specified number of masked patches are sampled uniformly at random without replacement from all $N$ patches,
independently for each input spectrum and anew at every training step.

We denote the spectral augmentation applied to the full input by $\mathcal{T}_{\xi}$.
TGN and FS are defined in Section~\ref{sec:augmentation}.
The variable $\xi$ collectively denotes whether and in what order the transformations are applied,
as well as their internal random variables.
The application settings are described in Section~\ref{sec:comparison-design}.
When no augmentation is applied, $\mathcal{T}_{\xi}$ is the identity transformation.
The augmented unmasked patches are converted into $D$-dimensional tokens by a shared linear map,
and learnable positional embeddings indicating their original wavelength positions are added.
These tokens and a CLS token are passed to the Transformer encoder; masked patches are not provided as input.
The embedding procedure and Transformer encoder together define the mapping $f_{\theta}$ to the final CLS output,
with learnable parameters $\theta$.
The encoding and decoding steps are given by
\begin{align}
    \text{Encoder:}\enspace&
    \mathbf{z}
    =
    \frac{
      \mathbf{W}_{\mathrm{proj}}\mathbf{t}+\mathbf{b}_{\mathrm{proj}}
    }{
      \left\|\mathbf{W}_{\mathrm{proj}}\mathbf{t}+\mathbf{b}_{\mathrm{proj}}\right\|_2
    },\quad
    \mathbf{t}
    =
    f_{\theta}\!\left(\mathcal{T}_{\xi}(\mathbf{x});\mathcal{V}\right),\notag\\[4pt]
    \text{Decoder:}\enspace&
    \widehat{\mathbf{x}}
    =
    \mathbf{W}_{\mathrm{dec}}\mathbf{z}+\mathbf{b}_{\mathrm{dec}},
  \label{eq:chemomae}
\end{align}

where $\mathbf{t}\in\mathbb{R}^{D}$ is the final CLS output,
$\mathbf{z}\in\mathbb{R}^{d}$ is the normalized latent representation,
$\mathbf{W}_{\mathrm{proj}}\in\mathbb{R}^{d\times D}$,
and $\mathbf{b}_{\mathrm{proj}}\in\mathbb{R}^{d}$.
The normalization denominator is assumed to be nonzero.
Input spectra after SNV transformation have zero mean and a constant norm,
so differences in spectral shape can be compared as differences in direction.
We also $\ell_2$-normalized the latent representations onto the unit sphere
so that cosine similarity could serve as a common measure in analyses of the input and latent spaces.
However, no zero-mean constraint is imposed on the latent representations.
The architecture is shown in Figure~\ref{fig:chemomae-overview}(a).

The decoder is a single affine layer with
$\mathbf{W}_{\mathrm{dec}}\in\mathbb{R}^{C\times d}$ and
$\mathbf{b}_{\mathrm{dec}}\in\mathbb{R}^{C}$,
which reconstructs the full spectrum from a single $\mathbf{z}$.
No skip connections from the input or per-patch encoder outputs are used.
The model thus learns a representation that concentrates the information needed for reconstruction in the bottleneck
and makes it accessible to the decoder.

Masked reconstruction exploits multicollinearity in NIR spectra as the model learns to predict masked bands from the remaining bands.
Using the input $\mathbf{x}$ before additional perturbations as the reconstruction target,
we minimize the mean squared error over masked channels (Figure~\ref{fig:chemomae-overview}(b)).
Let $\mathcal{I}_{\mathcal{M}}=\bigcup_{n\in\mathcal{M}}\mathcal{P}_n$ denote the set of masked channels;
the subscript $\mathcal{I}_{\mathcal{M}}$ indicates extraction of the corresponding components.
For $\mathcal{M}\neq\varnothing$, the loss is
\begin{equation}
  \mathcal{L}_{\mathrm{mask}}
  =
  \mathbb{E}_{\mathbf{x},\,\xi,\,\mathcal{M}}
  \left[
    \frac{
      \left\|
        \widehat{\mathbf{x}}_{\mathcal{I}_{\mathcal{M}}}
        -\mathbf{x}_{\mathcal{I}_{\mathcal{M}}}
      \right\|_2^2
    }{
      |\mathcal{I}_{\mathcal{M}}|
    }
  \right]
  \label{eq:masked-loss}
\end{equation}
The expectation is taken over training spectra, augmentation random variables, and masks,
and was approximated by the sample mean within each minibatch.
Although the decoder outputs all channels, errors on unmasked patches are excluded from the loss,
and no patchwise normalization is applied.

\subsection{SNV-Based Spectral Augmentation}
\label{sec:augmentation}

The additional perturbations were designed to vary the normalized spectral shape
without reintroducing the mean and scale variations removed by SNV transformation.

The usefulness of data augmentation for learning is not determined solely
by how closely generated examples resemble actual observations.
Improvements in classification performance have been reported with Mixup~\citep{zhang2018mixup}
and CutMix~\citep{yun2019cutmix}, which artificially combine inputs and their supervisory signals.
Vicinal Risk Minimization (VRM)~\citep{chapelle2000vicinal} assigns vicinity distributions to training examples
and minimizes the expected loss under their mixture.
From this perspective, we regard placing vicinity distributions within the set of vectors
with zero mean and a constant norm as a design choice for the inductive bias of this study.

\citet{fearn2009geometry} explain mean removal and length normalization by SNV transformation
in terms of placing spectra on a spherical surface.
Building on this geometric account, we explicitly define the set containing the inputs
and construct TGN and FS using the tangent space and renormalization.

Both transformations are applied to all $C$ components before masking and generate training examples within the same set.
They are inspired by signal noise~\citep{hamamatsuImageSensors} and wavelength shifts~\citep{busch2000wavelength},
but do not guarantee reproduction of measurement errors or preservation of chemical states.

The set containing inputs that satisfy Eq.~\eqref{eq:snv_constraints} is defined as
\begin{equation}
  \mathcal{X}_{\mathrm{SNV}}
  =
  \left\{
    \mathbf{x}\in\mathbb{R}^{C}
    \;\middle|\;
    \mathbf{1}_{C}^{\mathsf T}\mathbf{x}=0,\;
    \|\mathbf{x}\|_{2}=\sqrt{C-1}
  \right\}
  \label{eq:snv-manifold}
\end{equation}
Writing $\rho=\sqrt{C-1}$, this is a $(C-2)$-dimensional sphere of radius $\rho$
centered at the origin within the zero-mean hyperplane.

Because all vectors on this sphere share the same norm,
the angle relative to the input can be used as a common measure of perturbation strength.
TGN controls this angle directly, whereas FS specifies the displacement along the wavelength axis.

\subsubsection{Tangent Gaussian Noise}
\label{sec:noise}

TGN rotates the input by a specified angle on $\mathcal{X}_{\mathrm{SNV}}$.
Let $\mathbf{n}=\mathbf{x}/\rho$ be the unit vector in the input direction.
A zero-mean unit vector $\mathbf{q}$ orthogonal to $\mathbf{n}$
is selected uniformly from the unit sphere in the tangent space.
We generate $\mathbf{q}$ by projecting and normalizing standard normal random variates
(Appendix~\ref{app:augmentation-geometry}).
The term Gaussian refers to the distribution of these random variates.
TGN is defined as
\begin{equation}
  \begin{aligned}
    \mathcal{T}_{\mathrm{TGN}}(\mathbf{x})
    &=\rho\bigl(
      \cos\alpha\,\mathbf{n}
      +\sin\alpha\,\mathbf{q}
      \bigr),\\
    \alpha&\sim\mathcal{U}(0,5^\circ)
  \end{aligned}
  \label{eq:tgn}
\end{equation}
where $\mathcal{U}$ denotes the continuous uniform distribution over the specified interval.
The transformed spectrum retains zero mean and norm $\rho$, and its angle relative to the input is exactly $\alpha$.
These properties are derived in Appendix~\ref{app:augmentation-geometry}.

\subsubsection{Fractional Shift}
\label{sec:shift}

FS combines a shift on the equally spaced wavelength grid obtained by resampling during preprocessing
with readjustment by SNV transformation.
The displacement $\delta$ is expressed in units of the grid spacing
and is sampled for each spectrum from $\mathcal{U}(-2,2)$.
For channel index $j\in\{1,\ldots,C\}$, output channel $j$ refers to input position $j-\delta$.
Values at noninteger positions are obtained by linear interpolation between adjacent channels;
when the reference position falls outside the input range, the endpoint value is used.
We denote this shift operation, including interpolation and endpoint handling, by $S_\delta$.
Because the mean and norm are generally not preserved after shifting,
we recenter the spectrum by orthogonal projection onto the zero-mean hyperplane,
\begin{equation}
  \mathbf{P}=\mathbf{I}_C-\frac{1}{C}\mathbf{1}_C\mathbf{1}_C^{\mathsf T}
  \label{eq:centering-projection}
\end{equation}
and restore its norm to $\rho$.
Here, $\mathbf{I}_C$ is the $C$-dimensional identity matrix, and the FS transformation is defined as
\begin{equation}
  \mathcal{T}_{\mathrm{FS}}(\mathbf{x})
  =
  \rho
  \frac{\mathbf{P} S_\delta(\mathbf{x})}
       {\|\mathbf{P} S_\delta(\mathbf{x})\|_2}
  \label{eq:fractional-shift}
\end{equation}
If the denominator is nonzero, the transformed spectrum remains on $\mathcal{X}_{\mathrm{SNV}}$.
Even for the same displacement, the angular change relative to the input depends on spectral shape.
Examples of the TGN and FS transformations used to select their strengths are shown in Appendix~\ref{app:augmentation-visualization}.

Because FS is applied before masking, values near the boundaries of the target bands
may be included in unmasked bands through shifting and interpolation.
The reconstruction target remains fixed at the pre-perturbation values at the same wavelength positions.

\subsection{Unsupervised Segmentation}
\label{sec:mapping}

Latent representations for visualization and clustering were obtained with the trained model fixed,
without spectral augmentation and with all patches unmasked.
We quantized these representations into $K$ clusters and compared their spatial distributions.
Cluster labels do not imply an ordering of degradation severity or correspond to predefined chemical states.

For the main analysis, we used Cosine-KMeans, an implementation of spherical $k$-means
that partitions directions on the unit sphere using cosine similarity~\citep{banerjee2005hypersphere}.
In conventional $k$-means, even when the inputs are $\ell_2$-normalized,
cluster centers computed as the arithmetic means of their assigned data points generally do not have unit norm.
Cosine-KMeans $\ell_2$-normalizes the cluster centers after every update
and assigns inputs according to their cosine similarities to the centers.

The resulting cluster labels were mapped back to the original coordinates of each pixel.
Valid pixels were assigned labels $1,\ldots,K$, and background and excluded pixels were assigned 0.
Neither representation learning nor clustering used spatial coordinates or neighborhood relations,
and no spatial regularization or post-assignment label smoothing was applied.
This unsupervised segmentation procedure is illustrated in Figure~\ref{fig:chemomae-overview}(c).

Spectral augmentation, ChemoMAE, and hyperspherical clustering were implemented using
the Python library \texttt{chemomae}, developed and released by the authors.

We used OpenAI Codex to help write research code, including scripts for analyzing the NIR-HSI data.
We reviewed and edited the code and interpreted the results.

\makeatletter
\let\arxivcomparison@addtotoporbot\@addtotoporbot
\let\arxivcomparison@cflt\@cflt
\newskip\arxivcomparison@textfloatsep
\def\arxivcomparison@compactcflt{%
  \textfloatsep=12pt\relax
  \arxivcomparison@cflt
  \textfloatsep=\arxivcomparison@textfloatsep
  \def\@textbottom{\vskip\z@\@plus.0001fil}%
  \global\let\@cflt\arxivcomparison@cflt
}
\def\@addtotoporbot{%
  \ifx\@currbox\arxivcomparison@box
    \ifnum\c@page<8\relax
    \else
      \if@firstcolumn
      \else
        \global\arxivcomparison@textfloatsep=\textfloatsep
        \textfloatsep=12pt\relax
        \arxivcomparison@addtotoporbot
        \textfloatsep=\arxivcomparison@textfloatsep
        \if@insert
          \global\let\@cflt\arxivcomparison@compactcflt
          \global\let\@addtotoporbot\arxivcomparison@addtotoporbot
          \global\let\arxivcomparison@box\@empty
        \fi
      \fi
    \fi
  \else
    \arxivcomparison@addtotoporbot
  \fi
}
\makeatother
\begin{table}[!t]
  \makeatletter
  \global\let\arxivcomparison@box\@currbox
  \makeatother
  \centering
  \caption{Comparison conditions.
  The upper block shows the six main conditions and the lower block the two perturbation ablations with masking. AE denotes autoencoder.}
  \label{tab:comparison-conditions}
  \small
  \setlength{\tabcolsep}{2pt}
  \begin{tabular}{@{}llcll@{}}
    \toprule
    ID & Representation & \shortstack{Mask\\ratio} & \shortstack[l]{Additional\\perturbations} & \shortstack[l]{Loss\\computed on} \\
    \midrule
    B0  & Input spectra & \textemdash & \textemdash & \textemdash \\
    B1  & PCA           & \textemdash & \textemdash & \textemdash \\
    A0  & AE            & 0\%        & \textemdash & All bands \\
    A1  & Denoising AE  & 0\%        & TGN+FS & All bands \\
    M00 & ChemoMAE      & 50\%       & \textemdash & Masked bands \\
    M11 & ChemoMAE      & 50\%       & TGN+FS & Masked bands \\
    \midrule
    M10 & ChemoMAE      & 50\%       & TGN & Masked bands \\
    M01 & ChemoMAE      & 50\%       & FS & Masked bands \\
    \bottomrule
  \end{tabular}
\end{table}

\section{Experimental Design and Evaluation}
\label{sec:experiments}

We used specimen-level cross-validation (CV) to evaluate the spatial coherence of segmentations of unseen specimens,
followed by full-dataset fitting for cross-specimen comparison and NIR-based chemical interpretation.

\subsection{Comparison Conditions and Ablation Design}
\label{sec:comparison-design}

\subsubsection{Main Comparison Conditions}

Table~\ref{tab:comparison-conditions} summarizes six main conditions and two ablations of the additional perturbations;
M11 was the standard ChemoMAE configuration.
To compare representations under common preprocessing, we used SNV-transformed reflectance spectra from individual pixels (Section~\ref{sec:preprocessing}) in every condition, including PCA (B1).
We used second derivatives of pseudo-absorbance only for chemical interpretation after segmentation (Section~\ref{sec:nir-interpretation}).
B0 used the 256-dimensional input directly, whereas the other conditions reduced it to 16 dimensions.

The input spectra in B0 and the PCA scores in B1 were $\ell_2$-normalized before clustering.
No additional wavelength-wise standardization or whitening was applied in the PCA of B1.
All neural network conditions (A0, A1, and M00--M11) shared the encoder, bottleneck, and affine decoder
described in Section~\ref{sec:chemomae}, including $\ell_2$ normalization of the latent representations during training.
All reconstruction targets were the inputs before additional perturbations,
and the loss was the mean squared error over the bands specified in Table~\ref{tab:comparison-conditions}.
The training procedure and loss for the M conditions followed Section~\ref{sec:chemomae} and Eq.~\eqref{eq:masked-loss}.

Each enabled perturbation was applied to each pixel with probability 0.5,
and their order when combined was randomized for each batch.
The strength distribution of each perturbation was shared across conditions;
the operations and the rationale for their strengths are described in
Section~\ref{sec:augmentation} and Appendix~\ref{app:augmentation-visualization}.
The training pixels, number of epochs, and optimization settings were also shared,
and we used the final weights after 800 epochs (Appendix~\ref{app:config-details}).

We compared masked denoising (M11) with direct use of the input spectra (B0), PCA (B1), the conventional and denoising autoencoders (A0 and A1), and masked reconstruction without additional perturbations (M00).
To examine the contribution of the reconstruction task, we compared A0 with M00 without additional perturbations
and A1 with M11 with additional perturbations.
Both comparisons changed input masking and the bands included in the loss,
contrasting reconstruction of all bands with prediction of masked bands.

To examine additional perturbations, we compared A0 with A1 without masking
and used a $2\times2$ TGN--FS design within the M conditions, with a common mask ratio and the loss computed only over masked bands,
to assess the individual contributions of the two perturbations, their combined contribution and their interaction.
Interaction estimates and aggregation are described in Section~\ref{sec:metric-aggregation}.
We did not compare these perturbations with alternatives that fail to preserve the mean and norm.

\subsubsection{Clustering of the Representations}

Representation extraction and unsupervised segmentation followed Section~\ref{sec:mapping},
with Cosine-KMeans used in all conditions (Appendix~\ref{app:clustering-settings}).

We treated the number of clusters $K$ as the granularity at which the spectral distribution was examined.
Before inspecting the evaluation results, we fixed the common set
$\mathcal{K}=\{2,4,6,8,10,12,14\}$ for all conditions, folds, and repetitions.
Considering the combined effects of surface degradation, internal aging, anatomical differences, and moisture histories,
we specified $K_0=8$ in advance as the granularity for the main tables and full-dataset maps.
We did not prescribe the states or factors corresponding to individual clusters.
For each condition, fold, and repetition, we varied $K$ while keeping the representations fixed
to examine how differences between conditions depended on this granularity.

\subsection{Cross-Validation and Evaluation Metrics}
\label{sec:cv-metrics}

\subsubsection{Cross-Validation Design}
\label{sec:cv-design}

All conditions used the same five folds, formed by randomly partitioning the 49 specimen IDs (KYOw) without stratification by exposure history or wood taxon.
We ran cross-validation three times using the same five folds, holding out each specimen once per run.

To balance the contributions of specimens with different numbers of pixels,
we sampled $8192$ pixels uniformly at random without replacement from each training specimen in each fold.
This sample size was fixed in advance based on computational cost and the number of available pixels.
Within each fold, the same sampled pixels were used to fit PCA, train the neural networks,
and fit Cosine-KMeans across all conditions, values of $K$, and repetitions.

For each held-out specimen, we kept the fold-specific representation transformation and cluster centers fixed.
We transformed all valid pixel spectra and assigned the resulting representations to clusters as described in Section~\ref{sec:mapping}.
We refer to this procedure as out-of-fold (OOF) evaluation.

Only the random seeds for neural network training and clustering varied across runs.
In each repetition, we trained the neural networks from scratch and fitted Cosine-KMeans to their representations.
For B0 and B1, the input representations or the PCA transformation fitted within each fold were held fixed,
and only clustering was repeated.

\subsubsection{Evaluation Metrics}
\label{sec:evaluation-metrics}

Local Label Agreement (LLA) was the primary metric for assessing consistency with the expectation
of spatial coherence on unseen specimens.
Under the hypothesis described in Section~\ref{sec:introduction}, representations that capture chemical state differences across specimens
are expected to yield high LLA on unseen specimens.
However, high LLA alone does not establish that these differences have been correctly captured.
Pixel neighborhood relations were used to calculate LLA but were excluded from representation learning and cluster assignment.
Label Flip Rate (LFR) was an auxiliary metric for examining whether more stable assignments
under the specified perturbations were accompanied by higher LLA.

Unless otherwise stated, the condition, $K$, and repetition are fixed in the following definitions.
For specimen $s$, let $\Omega_s$ be the set of valid pixels,
$N_s=|\Omega_s|$ the number of valid pixels,
$y_p$ the label of pixel $p\in\Omega_s$ corresponding to coordinates $(u,v)$ in the original image,
and $n_{s,k}$ the number of pixels assigned to cluster $k$.
Background pixels and pixels excluded by the quality criteria were omitted
from both the evaluation and spatial neighborhoods.

\paragraph{Local Label Agreement}

Here, spatial coherence denotes local agreement among neighboring pixel labels, quantified by LLA.
Neighborhoods were square regions of width $w\in\{3,5,9\}$, excluding the central pixel.
Let $A_{s,w}$ be the fraction of all valid neighboring pixel pairs that have the same label.
This agreement rate corresponds to join-count statistics that count neighboring pairs of the same category~\citep{epperson2003joincounts},
but can be high simply because one cluster occupies a large fraction of the pixels.
We therefore corrected it using the agreement probability $P_s$ under uniformly random assignment of labels,
with valid pixel locations and the number of pixels in each cluster held fixed, and defined LLA as
\begin{equation}
  P_s=\frac{\sum_{k=1}^{K}\binom{n_{s,k}}{2}}{\binom{N_s}{2}},
  \quad
  \mathrm{LLA}_{s,w}=\frac{A_{s,w}-P_s}{1-P_s}
  \label{eq:lla}
\end{equation}
Here, $\binom{n}{2}$ denotes the binomial coefficient, defined as zero for $n<2$.
$P_s$ equals the probability that two distinct valid pixels sampled without replacement have the same label.
Under random assignment, the expected LLA is zero;
positive values indicate agreement above chance, and negative values indicate agreement below chance.

The agreement rate $A_{s,w}$ was calculated by convolution of binary images.
Let $\mathbf{B}_{s,k}$ be the image equal to one at pixels assigned to cluster $k$ and zero elsewhere,
and let $\mathbf{M}_s=\sum_{k=1}^{K}\mathbf{B}_{s,k}$ be the valid-pixel mask.
These images are zero at background pixels, at excluded pixels and outside the image.
With a $w\times w$ kernel $\mathbf{h}_w$ whose central element is zero and all other elements are one,
\begin{equation}
  A_{s,w}
  =
  \frac{
    \sum_{k=1}^{K}\sum_{u,v}
    \bigl[\mathbf{B}_{s,k}\odot(\mathbf{h}_w*\mathbf{B}_{s,k})\bigr]_{u,v}
  }{
    \sum_{u,v}
    \bigl[\mathbf{M}_s\odot(\mathbf{h}_w*\mathbf{M}_s)\bigr]_{u,v}
  }
  \label{eq:lla-observed-agreement}
\end{equation}
where $*$ denotes two-dimensional discrete convolution, $\odot$ denotes elementwise multiplication,
$[\cdot]_{u,v}$ denotes the value at coordinates $(u,v)$,
and $\sum_{u,v}$ denotes summation over all rows and columns of the image.
All valid neighboring pairs receive equal weight; no normalization by the number of neighbors of each pixel is applied.

We report LLA-3, LLA-5, and LLA-9 separately, and use LLA-5 when presenting comparisons for a single neighborhood width.
Cases in which Eq.~\eqref{eq:lla} is undefined, including assignment to a single cluster ($P_s=1$), are excluded from aggregation.

LLA equals one minus the variance-weighted mean of Geary's $C$~\citep{geary1954contiguity}
for the binary indicator variables of the clusters represented within a specimen.
The correspondence between the uncorrected agreement rate and join-count statistics,
the properties of the chance correction, and the derivation of this equivalence
are given in Appendix~\ref{app:lla-spatial-association}.

\paragraph{Label Flip Rate}

LFR is the fraction of pixels whose labels change when spectral perturbations are applied with the model held fixed.
For each perturbation type, all valid pixels of each specimen were evaluated five times using different perturbation realizations,
and the arithmetic mean was taken as the specimen's LFR.
A lower LFR indicates more stable labels under the specified perturbations.

The combined use of TGN and FS was the main evaluation perturbation;
TGN alone and FS alone were also evaluated separately.
The operations, strength distributions, and ordering when combined matched those used during training
(Sections~\ref{sec:augmentation} and~\ref{sec:comparison-design}),
with only the application probability of the enabled operations changed to one.
Perturbations were applied to the input spectra before the representation transformation.
The same perturbation realization for each pixel was shared across all conditions, values of $K$, and repetitions.
The representation transformation and cluster centers were held fixed,
and inference with the AE, denoising AE, and ChemoMAE used all patches without masking.

\paragraph{Cosine Silhouette}

To calculate silhouettes~\citep{rousseeuw1987silhouettes},
we normalized each condition's representations to unit length and used cosine distance,
defined as one minus cosine similarity; we refer to this metric as cosine-silhouette.
We pooled all valid pixels from the held-out specimens in each fold to calculate a silhouette value for each pixel,
then took the arithmetic mean over all valid pixels within each specimen to obtain its score.
Although specimens receive equal weight in the final aggregation,
average distances to each cluster are calculated from all pooled pixels within a fold,
so specimens with more pixels contribute more to these distance averages.
Because distance structures differ between conditions,
we treated cosine-silhouette as an auxiliary diagnostic describing separation within each representation space.
Our clustering quantizes spectral representations at granularity $K$;
it does not require clearly separated groups or maximization of the silhouette.

\paragraph{Cluster Occupancy}

For specimen $s$, cluster occupancy is $\pi_{s,k}=n_{s,k}/N_s$,
and maximum occupancy is $\max_k\pi_{s,k}$.
The number of clusters used, $K_s^{\mathrm{used}}$,
is the number of clusters to which at least one pixel in the specimen is assigned.
These diagnostics were used to detect concentrated assignments or assignment to a single cluster.

\subsubsection{Aggregation and Paired Comparisons}
\label{sec:metric-aggregation}

For each metric, condition, $K$, and repetition, we averaged specimen scores with equal weights,
then averaged the three repetition means to obtain the summary value.
Variability among specimens is reported as the sample standard deviation of the specimen scores averaged over repetitions.
CV summaries used the OOF scores from all five folds.

Differences between conditions were calculated as paired differences for the same specimen, $K$, and repetition.
For each specimen, the paired difference was averaged over the three repetitions.
We summarize these paired differences across specimens by reporting their mean, their sample standard deviation
and the number of specimens with a difference in the direction of improvement.

Undefined values were not imputed.
A condition's summary value used only specimens with defined values in all three repetitions for that condition.
Differences between conditions used the common set of specimens with defined values in all three repetitions of both conditions,
with the number of specimens and the reasons for undefined values reported alongside the results.
The treatment of undefined values and degenerate segmentations is described in Appendix~\ref{app:evaluation-conventions}.

For each specimen, $K$, and training repetition, the TGN--FS interaction was calculated as
$(\mathrm{M11}-\mathrm{M01})-(\mathrm{M10}-\mathrm{M00})$ using the corresponding metric values,
separately for each LLA neighborhood width and LFR perturbation type.
Only specimens with defined values in all four conditions and all three repetitions were included.
These contrasts were aggregated as described above for paired differences.
A positive LLA contrast or a negative LFR contrast indicates an interaction in the direction of improvement
relative to the sum of the individual effects.

\subsection{Full-Dataset Fitting and Spatial Analysis}
\label{sec:full-fit-protocol}

For each condition, we used spectra drawn from all 49 specimens to obtain a common basis for comparing maps across specimens.
We compared spatial maps across the six main conditions (B0, B1, A0, A1, M00, and M11) at $K_0=8$
and used the M11 assignments for the clusterwise NIR analysis in Section~\ref{sec:nir-interpretation}.

We sampled $8192$ valid pixels uniformly at random without replacement from each specimen
and shared the sampled pixels across all conditions.
Using the same settings as in CV, we fitted PCA once and trained each neural network once from scratch.
With the representation transformation and fitting pixels held fixed,
we fitted Cosine-KMeans three times, changing only its initialization seed; the seeds were shared across conditions.
In each repetition, the representation transformation and cluster centers were applied to all valid pixels, including the fitting pixels.
The first repetition, selected in advance, was used for the maps.

We calculated LLA-3, LLA-5, and LLA-9 from all valid pixels to describe and compare
the spatial coherence of segmentations in the specimens included in fitting
(see Section~\ref{sec:evaluation-metrics} for definitions).
Aggregation and paired comparisons followed Section~\ref{sec:metric-aggregation},
including the treatment of undefined values and common specimen sets.

To examine whether spatial structure consistent with prior knowledge emerged in maps derived from spectra alone,
we examined cluster distributions in relation to growth rings on the measured transverse sections and, where the position of a surface exposed during service could be inferred, in relation to that surface.
We compared regions across maps in terms of their extent, boundaries, relative positions and patterns of splitting or merging.
We also examined correspondence with the scanning direction for possible variation associated with the measurement system.
Comparative maps are presented in Appendix~\ref{app:global-map-comparison}.

Display IDs and colors across conditions were aligned to M00 based on cluster means of the input spectra
(Appendix~\ref{app:spectral-summary-settings}).

\subsection{NIR Spectral Analysis and Interpretation}
\label{sec:nir-interpretation}

We related spatial distributions to NIR spectra using the first set of cluster assignments
from full-dataset fitting of M11 at $K=8$ on all 49 specimens.

For each pixel with positive, finite reflectance in all 256 bands, we calculated pseudo-absorbance
$A(\lambda)=-\log_{10}R(\lambda)$ from the resampled reflectance before SNV transformation.
Because differentiation can accentuate high-frequency noise in individual pixel spectra, we first averaged the pseudo-absorbance spectra within each specimen and cluster.
We computed the second derivative of each mean spectrum using the Savitzky--Golay method
with a seven-point window, polynomial order two,
a wavelength spacing of approximately 5.46\,nm, and polynomial fitting at the edges.
Although the horizontal axis is displayed as wavenumber (cm$^{-1}$),
the second derivatives are with respect to wavelength, $d^2A/d\lambda^2$ (nm$^{-2}$).

For the aggregated plot (Figure~\ref{fig:m11-nir-spectra}),
cluster mean spectra were averaged with equal weights across specimens containing contributing pixels.

Based on prior knowledge of wood anatomy and degradation, we expected regions of similar chemical state to span neighboring pixels at the scale of observation.
For specimen-level spectral comparisons, we visually inspected each map and selected clusters whose assigned pixels formed discernible regions (Appendix~\ref{app:specimen-nir-atlas}).
Where the position of a surface exposed during service could be inferred, we also compared regions near that surface with regions in the interior.

We use band numbers 1--18 throughout (Tables~\ref{tab:nir-band-summary} and~\ref{tab:nir-band-assignments}).
We checked the assignments in Table~2 of \citet{schwanninger2011assignments} against original NIR studies~\citep{fackler2010polysaccharide,ali2001cellulosic,moll2022water}.
For the selected specimens and clusters, second-derivative values for these bands
were obtained by piecewise linear interpolation along the wavenumber axis.
For an assignment at a single wavenumber, we used the interpolated value;
for an interval, we used the minimum value over the interval, including its endpoints.
After visual inspection of the second-derivative spectra,
we plotted 13 bands in the bar charts (Figures~\ref{fig:internal-baseline-bands} and~\ref{fig:surface-baseline-bands}), excluding bands 8, 11, 12, 13, and 14 because their second-derivative troughs were indistinct.
For comparison, assignments including combination bands were organized into four groups:
polysaccharide-related C--H, lignin-related C--H, polysaccharide-related O--H, and water-related bands.
Assignments spanning more than one group or having multiple candidates were marked with *.
We defined the band value as the sign-reversed second derivative, $-d^2A/d\lambda^2$,
so that deeper negative troughs gave larger values.
This is a measure of curvature and does not directly represent constituent abundance.
Because the finish of the measured transverse surfaces varied among specimens (Section~\ref{sec:specimens}), we could not separate its possible contribution to differences in second-derivative band values from that of wood chemistry.

The common reference, termed the baseline, was constructed from clusters judged to be relatively sound.
Within each specimen, we averaged band values equally over the selected baseline clusters present, if any.
Baseline and individual-cluster values were then averaged with equal weights across their respective contributing specimens.
Missing values were not imputed in either aggregation.
We examined the direction of paired band-value differences within specimens contributing to both groups.

For each band, the bar charts show the mean across specimens for each cluster or the baseline group divided by the corresponding baseline mean.
The baseline is thus one, and the same baseline was used in both the internal and surface degradation plots.
Paired comparisons used specimen-specific relative band values,
obtained by dividing each specimen's band value by the same baseline mean across specimens.
Second-derivative spectra were used to compare absolute amplitudes between bands.

Within each specimen, we related the mean NIR spectrum of each selected cluster to the location and extent of the regions assigned to it.
We then cross-referenced these observations across specimens to examine whether an interpretation suggested by one specimen was consistent with the spectrum and spatial distribution of the same cluster in others.
In interpreting shared patterns and differences, we drew on published band assignments, wood anatomy, specimen provenance and taxon, and the available evidence on service conditions at the sampled locations in the original timber members.

\begin{arxivwidetoptable}{13}
  \caption{OOF evaluation of the eight comparison conditions at $K=8$ across 49 specimens.
  Upper entries are means; gray lower entries show $\pm$ the between-specimen standard deviation
  after averaging each specimen's scores over three repetitions.
  $\uparrow$ and $\downarrow$ indicate that higher and lower values, respectively, are better.
  Bold denotes the best mean for each metric with a specified improvement direction.}
  \label{tab:oof-k8}
  \par\nointerlineskip\vskip3pt
  \normalsize
  \setlength{\tabcolsep}{3pt}
  \newcommand{\arxivSixOnePair}[2]{\shortstack{$#1$\\[-1pt]\textcolor{black!55}{$\pm#2$}}}
  \begin{tabular*}{\linewidth}{@{\extracolsep{\fill}}lcccccccc@{}}
    \toprule
    & \multicolumn{6}{c}{Main comparison conditions} & \multicolumn{2}{c}{Augmentation ablations} \\
    \cmidrule(lr){2-7}\cmidrule(l){8-9}
    Metric & B0 & B1 & A0 & A1 & M00 & M11 & M10 & M01 \\
    \midrule
    LLA-3 $\uparrow$ & \arxivSixOnePair{0.548}{0.0922} & \arxivSixOnePair{0.537}{0.100} & \arxivSixOnePair{0.437}{0.131} & \arxivSixOnePair{0.394}{0.103} & \arxivSixOnePair{0.598}{0.0839} & \arxivSixOnePair{\mathbf{0.609}}{0.0894} & \arxivSixOnePair{0.603}{0.0894} & \arxivSixOnePair{0.599}{0.0944} \\
    LLA-5 $\uparrow$ & \arxivSixOnePair{0.450}{0.0914} & \arxivSixOnePair{0.458}{0.0987} & \arxivSixOnePair{0.381}{0.121} & \arxivSixOnePair{0.341}{0.0941} & \arxivSixOnePair{0.510}{0.0822} & \arxivSixOnePair{\mathbf{0.519}}{0.0875} & \arxivSixOnePair{0.513}{0.0862} & \arxivSixOnePair{0.509}{0.0916} \\
    LLA-9 $\uparrow$ & \arxivSixOnePair{0.345}{0.0748} & \arxivSixOnePair{0.366}{0.0859} & \arxivSixOnePair{0.310}{0.106} & \arxivSixOnePair{0.273}{0.0822} & \arxivSixOnePair{0.405}{0.0724} & \arxivSixOnePair{\mathbf{0.412}}{0.0779} & \arxivSixOnePair{0.406}{0.0728} & \arxivSixOnePair{0.402}{0.0797} \\
    LFR(TGN+FS) $\downarrow$ & \arxivSixOnePair{0.138}{0.0293} & \arxivSixOnePair{0.201}{0.0791} & \arxivSixOnePair{0.558}{0.116} & \arxivSixOnePair{0.252}{0.0746} & \arxivSixOnePair{0.269}{0.0858} & \arxivSixOnePair{\mathbf{0.0891}}{0.0385} & \arxivSixOnePair{0.214}{0.0812} & \arxivSixOnePair{0.168}{0.0747} \\
    LFR(TGN) $\downarrow$ & \arxivSixOnePair{\mathbf{0.0259}}{0.00595} & \arxivSixOnePair{0.0551}{0.0206} & \arxivSixOnePair{0.239}{0.0639} & \arxivSixOnePair{0.224}{0.0669} & \arxivSixOnePair{0.169}{0.0647} & \arxivSixOnePair{0.0815}{0.0357} & \arxivSixOnePair{0.0848}{0.0385} & \arxivSixOnePair{0.167}{0.0756} \\
    LFR(FS) $\downarrow$ & \arxivSixOnePair{0.135}{0.0292} & \arxivSixOnePair{0.190}{0.0779} & \arxivSixOnePair{0.538}{0.124} & \arxivSixOnePair{0.0809}{0.0270} & \arxivSixOnePair{0.221}{0.0795} & \arxivSixOnePair{\mathbf{0.0234}}{0.0119} & \arxivSixOnePair{0.210}{0.0775} & \arxivSixOnePair{0.0306}{0.0170} \\
    Cosine-silhouette & \arxivSixOnePair{0.368}{0.0714} & \arxivSixOnePair{0.343}{0.0841} & \arxivSixOnePair{0.261}{0.0986} & \arxivSixOnePair{0.243}{0.0855} & \arxivSixOnePair{0.197}{0.101} & \arxivSixOnePair{0.197}{0.104} & \arxivSixOnePair{0.196}{0.107} & \arxivSixOnePair{0.204}{0.107} \\
    $K_s^{\mathrm{used}}$ & \arxivSixOnePair{7.60}{0.663} & \arxivSixOnePair{7.76}{0.945} & \arxivSixOnePair{7.44}{0.740} & \arxivSixOnePair{7.61}{0.558} & \arxivSixOnePair{7.25}{0.815} & \arxivSixOnePair{7.22}{0.899} & \arxivSixOnePair{7.26}{0.893} & \arxivSixOnePair{7.25}{0.889} \\
    $\max_k\pi_{s,k}$ & \arxivSixOnePair{0.494}{0.118} & \arxivSixOnePair{0.483}{0.173} & \arxivSixOnePair{0.514}{0.167} & \arxivSixOnePair{0.487}{0.128} & \arxivSixOnePair{0.557}{0.148} & \arxivSixOnePair{0.562}{0.171} & \arxivSixOnePair{0.552}{0.164} & \arxivSixOnePair{0.563}{0.165} \\
    \bottomrule
  \end{tabular*}
\end{arxivwidetoptable}

\begin{arxivrighttopfigure}{12}
  \centering
  \includegraphics[width=\linewidth]{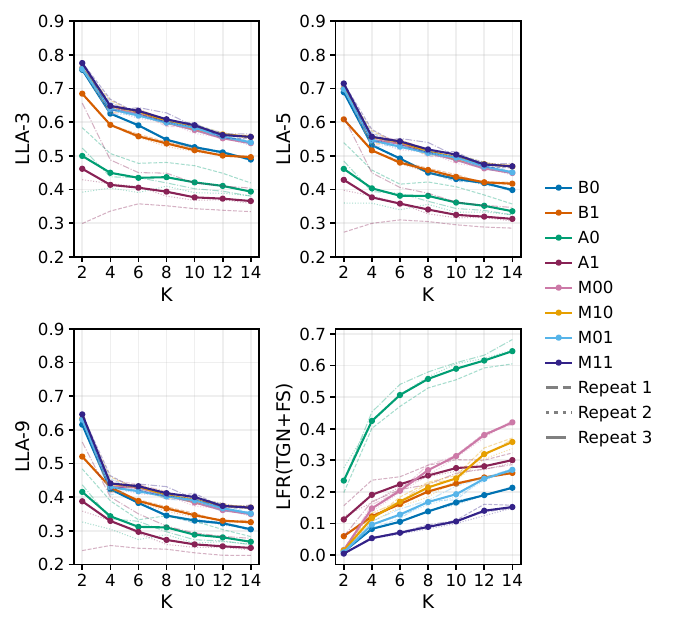}\par
  \caption{OOF metrics as a function of $K$ for the eight conditions in Table~\ref{tab:comparison-conditions}.
  Solid lines with circles show the summary values; thin lines show means across specimens for individual repetitions.
  All 49 specimens were included, except for LLA at $K=2$, where A0, B1, and M11 each had 48 specimens
  after excluding undefined values due to assignment to a single cluster.}
  \label{fig:oof-k-sweep}
\end{arxivrighttopfigure}

\newsavebox{\arxivAblationTableBody}
\begin{arxivrighttoptable}{13}
  \centering
  \caption{Paired differences and interactions across 49 specimens in OOF evaluation at $K=8$.
  Upper entries are mean effects; gray lower entries show $\pm$ the between-specimen standard deviation.
  ``Improved'' gives the number of specimens with a positive effect for LLA-5 or a negative effect for LFR(TGN+FS).}
  \label{tab:reconstruction-effects}
  \sbox{\arxivAblationTableBody}{%
    \small
    \setlength{\tabcolsep}{3pt}%
\begin{tabular}{@{}lcccc@{}}
    \toprule
    Comparison & \multicolumn{2}{c}{LLA-5} & \multicolumn{2}{c}{LFR(TGN+FS)} \\
    \cmidrule(lr){2-3}\cmidrule(l){4-5}
    & Effect & Improved & Effect & Improved \\
    \midrule
    M11$-$B0 & \shortstack{$0.0691$\\\textcolor{black!55}{$\pm0.0658$}} & 41/49 & \shortstack{$-0.0490$\\\textcolor{black!55}{$\pm0.0423$}} & 44/49 \\[2pt]
    M11$-$B1 & \shortstack{$0.0610$\\\textcolor{black!55}{$\pm0.0903$}} & 41/49 & \shortstack{$-0.112$\\\textcolor{black!55}{$\pm0.0757$}} & 45/49 \\
    \midrule
    M00$-$A0 & \shortstack{$0.128$\\\textcolor{black!55}{$\pm0.106$}} & 45/49 & \shortstack{$-0.289$\\\textcolor{black!55}{$\pm0.0966$}} & 49/49 \\[2pt]
    M11$-$A1 & \shortstack{$0.179$\\\textcolor{black!55}{$\pm0.0849$}} & 49/49 & \shortstack{$-0.163$\\\textcolor{black!55}{$\pm0.0677$}} & 49/49 \\
    \midrule
    A1$-$A0 & \shortstack{$-0.0407$\\\textcolor{black!55}{$\pm0.0705$}} & 15/49 & \shortstack{$-0.306$\\\textcolor{black!55}{$\pm0.121$}} & 49/49 \\[2pt]
    M10$-$M00 & \shortstack{$0.00333$\\\textcolor{black!55}{$\pm0.0241$}} & 30/49 & \shortstack{$-0.0548$\\\textcolor{black!55}{$\pm0.0426$}} & 46/49 \\[2pt]
    M01$-$M00 & \shortstack{$-0.000515$\\\textcolor{black!55}{$\pm0.0429$}} & 25/49 & \shortstack{$-0.101$\\\textcolor{black!55}{$\pm0.0475$}} & 49/49 \\[2pt]
    M11$-$M00 & \shortstack{$0.00970$\\\textcolor{black!55}{$\pm0.0261$}} & 31/49 & \shortstack{$-0.180$\\\textcolor{black!55}{$\pm0.0578$}} & 49/49 \\
    \midrule
    TGN$\times$FS & \shortstack{$0.00689$\\\textcolor{black!55}{$\pm0.0488$}} & 28/49 & \shortstack{$-0.0239$\\\textcolor{black!55}{$\pm0.0510$}} & 36/49 \\
    \bottomrule
  \end{tabular}%
  }
  \ifdim\wd\arxivAblationTableBody>\linewidth
    \resizebox{\linewidth}{!}{\usebox{\arxivAblationTableBody}}%
  \else
    \usebox{\arxivAblationTableBody}%
  \fi
\end{arxivrighttoptable}

\begin{arxivwidetopfigure}{14}
  \resizebox{!}{\dimexpr\height+3\baselineskip\relax}{\includegraphics[width=0.8\linewidth,keepaspectratio]{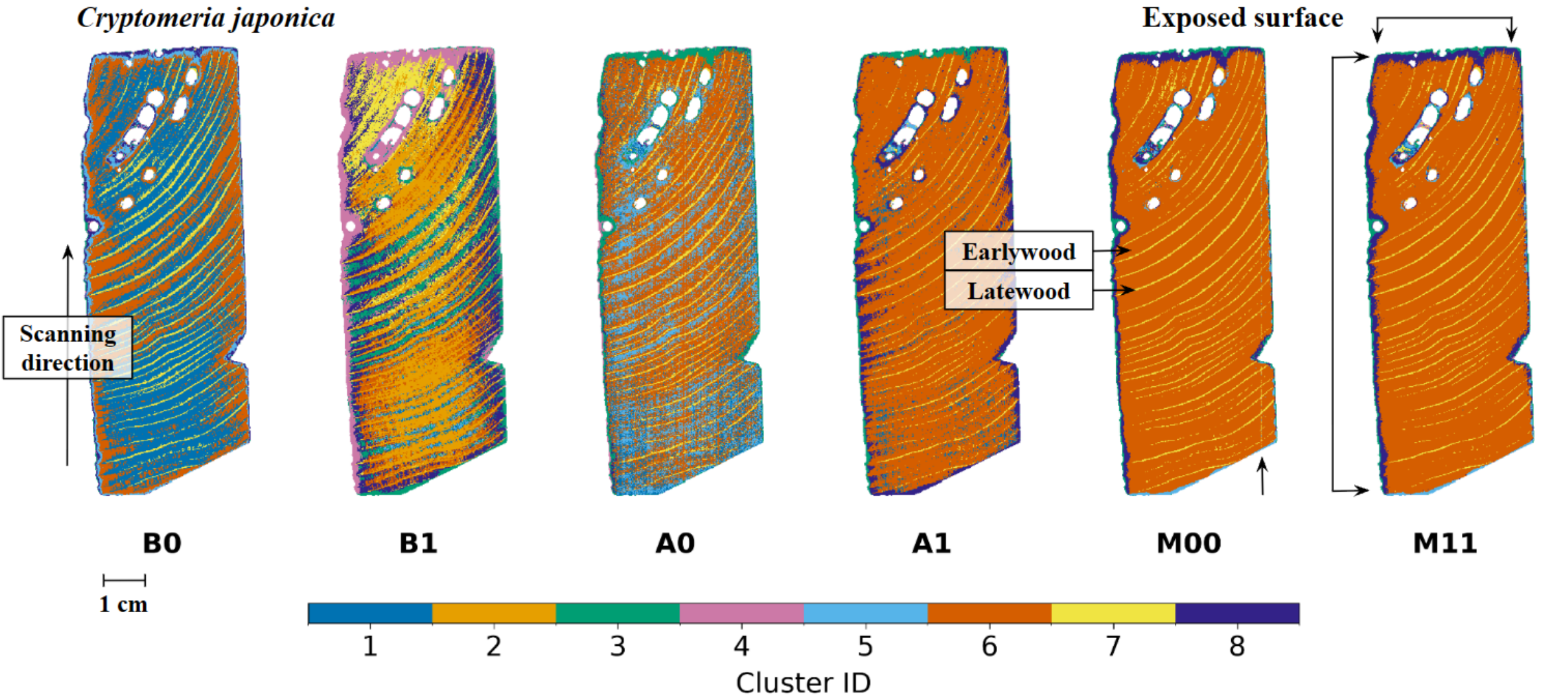}}\par
  \caption{Spatial maps of the measured transverse surface of KYOw16744 after full-dataset fitting under the six main conditions ($K=8$).
  The first clustering result is shown for each condition.
  Colors and display IDs are aligned to M00; white denotes background and excluded pixels.
  The arrows labeled ``Exposed surface'' mark perimeter segments corresponding to the original timber member's
  radial or tangential side surfaces that were externally exposed during service.
  The arrow at the lower right of M00 indicates a cluster assignment pattern aligned with the scanning direction.}
  \label{fig:global-16744-maps}
\end{arxivwidetopfigure}

\newsavebox{\arxivGlobalTableBody}
\begin{arxivrighttoptable}{15}
  \centering
  \caption{Spatial coherence under the six main conditions across three repetitions of Cosine-KMeans
  on fixed representations from full-dataset fitting ($K=8$, 49 specimens).
  Upper entries are means; gray lower entries show $\pm$ the between-specimen standard deviation
  after averaging each specimen's scores over three repetitions.
  Bold denotes the highest mean for each neighborhood width.}
  \label{tab:global-k8}
  \sbox{\arxivGlobalTableBody}{%
    \small
    \setlength{\tabcolsep}{2pt}%
    \begin{tabular}{@{}lcccccc@{}}
    \toprule
    Metric & B0 & B1 & A0 & A1 & M00 & M11 \\
    \midrule
    LLA-3 $\uparrow$ & \shortstack{$0.545$\\\textcolor{black!55}{$\pm0.0945$}} & \shortstack{$0.533$\\\textcolor{black!55}{$\pm0.125$}} & \shortstack{$0.348$\\\textcolor{black!55}{$\pm0.115$}} & \shortstack{$0.609$\\\textcolor{black!55}{$\pm0.0853$}} & \shortstack{$0.608$\\\textcolor{black!55}{$\pm0.106$}} & \shortstack{$\mathbf{0.616}$\\\textcolor{black!55}{$\pm0.102$}} \\[2pt]
    LLA-5 $\uparrow$ & \shortstack{$0.448$\\\textcolor{black!55}{$\pm0.0903$}} & \shortstack{$0.454$\\\textcolor{black!55}{$\pm0.120$}} & \shortstack{$0.299$\\\textcolor{black!55}{$\pm0.105$}} & \shortstack{$\mathbf{0.536}$\\\textcolor{black!55}{$\pm0.0903$}} & \shortstack{$0.518$\\\textcolor{black!55}{$\pm0.110$}} & \shortstack{$0.526$\\\textcolor{black!55}{$\pm0.107$}} \\[2pt]
    LLA-9 $\uparrow$ & \shortstack{$0.342$\\\textcolor{black!55}{$\pm0.0718$}} & \shortstack{$0.361$\\\textcolor{black!55}{$\pm0.106$}} & \shortstack{$0.239$\\\textcolor{black!55}{$\pm0.0883$}} & \shortstack{$\mathbf{0.440}$\\\textcolor{black!55}{$\pm0.0945$}} & \shortstack{$0.411$\\\textcolor{black!55}{$\pm0.100$}} & \shortstack{$0.422$\\\textcolor{black!55}{$\pm0.102$}} \\[2pt]
    \bottomrule
  \end{tabular}%
  }
  \ifdim\wd\arxivGlobalTableBody>\linewidth
    \resizebox{\linewidth}{!}{\usebox{\arxivGlobalTableBody}}%
  \else
    \usebox{\arxivGlobalTableBody}%
  \fi
\end{arxivrighttoptable}

\begin{arxivwidetopfigure}{17}
  \includegraphics[width=\linewidth,keepaspectratio]{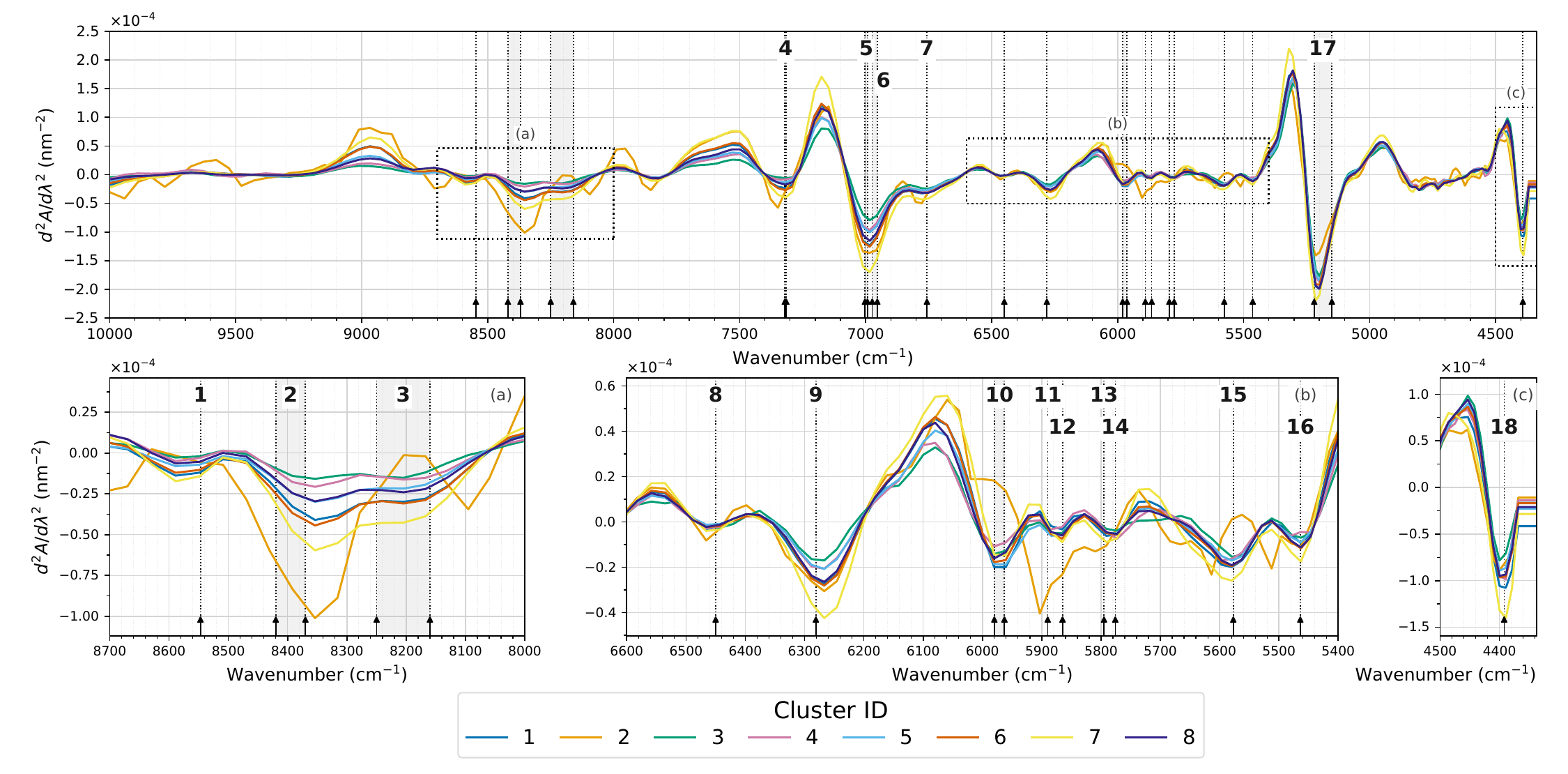}\par
  \caption{Clusterwise NIR spectra from full-dataset fitting of M11 ($K=8$).
  The upper panel shows second derivatives of pseudo-absorbance; the lower panels enlarge regions (a)--(c).
  Colors denote cluster IDs. Band numbers 1--18 correspond to Table~\ref{tab:nir-band-summary}.
  Arrows and vertical dotted lines mark the literature wavenumbers used for comparison.
  Gray bands indicate wavenumber ranges reported in the literature.}
  \label{fig:m11-nir-spectra}
\end{arxivwidetopfigure}

\section{Results and Discussion}
\label{sec:results}

\subsection{Spatial Coherence on Unseen Specimens}
\label{sec:results-k8}
\label{sec:results-k-sweep}

At the prespecified $K=8$, the standard masked denoising configuration (M11) gave the highest mean OOF LLA
at all three neighborhood widths (Table~\ref{tab:oof-k8}).
For most held-out specimens, M11 also gave higher LLA-5 than direct use of input spectra (B0) or PCA (B1)
(Table~\ref{tab:reconstruction-effects}).
B0 and B1 yielded higher LLA than the two unmasked AEs (A0 and A1),
making them strong reference conditions for unseen specimens.
For B0 and B1, representations were held fixed within each fold and only clustering was repeated;
for the neural network conditions, representations were learned anew in each repetition,
so the evaluation also included variation due to representation learning.

The advantage over B0 and B1 was not limited to a particular number of clusters (Figure~\ref{fig:oof-k-sweep}).
LLA generally decreased with increasing $K$, but the M conditions, including M11,
exceeded B0 and B1 at every $K$ and neighborhood width examined.
Differences among the M conditions were small:
M10 slightly exceeded M11 at all three neighborhood widths for $K=12$ and in LLA-9 for $K=14$.
These results show that representations learned by masked reconstruction remained useful
as $K$, the granularity of analysis defined in Section~\ref{sec:comparison-design}, was varied.

The higher LLA of the M conditions in segmentation based on spectra alone
is consistent with the expectation in Section~\ref{sec:introduction}
that spatial coherence would emerge in unseen specimens if features capturing state differences were learned across specimens.
These results support the learned representations' usefulness in terms of the expected spatial coherence on unseen specimens.

The value of evaluating this expected property is also illustrated by the contrast with cosine-silhouette.
The M conditions showed higher mean LLA but lower cosine-silhouette values than B0 and B1
(Table~\ref{tab:oof-k8}).
Cosine-silhouette and LLA capture distinct properties: cluster separation within each representation space
and local spatial coherence within each segmentation, respectively.
Choosing a metric of cluster separation in representation space solely because clustering is used would overlook this expected spatial property.
This contrast highlights the value of hypothesis-driven evaluation design:
selecting metrics according to properties expected from scientific hypotheses rather than the analytical method alone.

\subsection{Effects of Reconstruction Tasks and Spectral Augmentation}
\label{sec:results-ablation}

Replacing reconstruction of all bands with prediction of masked bands improved
the spatial coherence of segmentations of unseen specimens.
In OOF evaluation at $K=8$, mean LLA-5 increased and LFR(TGN+FS) decreased
in both M00$-$A0 and M11$-$A1, which changed input masking and the bands included in the loss
while keeping the model architecture the same (Table~\ref{tab:reconstruction-effects}).
Improvements were observed across a wide range of specimens,
and the mean LLA-5 difference was positive in all three repetitions for both comparisons
(Table~\ref{tab:oof-k8-repeat-effects}).

These improvements suggest that masked reconstruction is well suited to NIR spectra,
which contain related chemical information across multiple bands.
NIR spectra contain overtone and combination bands associated with the same chemical bonds
across multiple wavelength regions, and band overlap and multicollinearity are characteristic features of these spectra.
Thus, information reflecting state differences is not confined to a single band
and can occur redundantly across different bands.
Predicting masked bands from the remaining bands exploits this redundancy
and encourages learning of variation shared across bands and interband dependencies.
Under the constraint of reconstruction by an affine decoder from a single low-dimensional representation
(Section~\ref{sec:chemomae}), predictive cues linking multiple bands must be concentrated in that representation.
We suggest that this contributed to learning features shared across specimens
without relying on particular bands alone.

More stable assignments under the specified perturbations did not necessarily coincide
with greater spatial coherence of segmentations of unseen specimens.
In A1$-$A0, LFR(TGN+FS), averaged over three repetitions, decreased in all 49 specimens,
whereas mean LLA-5 decreased.
LFR assesses responses to perturbations of the same types and strength distributions as those used during training,
whereas LLA assesses label agreement among actual spatial neighbors,
whose neighborhood relations are excluded from both representation learning and clustering.

In M11, which combined additional perturbations with masked reconstruction,
the reduction in LFR(TGN+FS) relative to M00 was accompanied by a small increase in mean LLA-5.
For inputs sampled from the vicinity distributions designed in Section~\ref{sec:augmentation},
the model must predict the pre-perturbation values of masked bands even when the unmasked bands have been perturbed.
This combination may have encouraged the representation to capture interband relationships
that were less sensitive to these transformations, contributing to the additional gain in spatial coherence.
Because M00 also had lower LFR(TGN+FS) than A0 despite having no additional perturbations,
the reconstruction task itself contributed to more stable assignments.
The role of denoising training with additional perturbations should therefore be considered
in combination with the reconstruction task.

Adding TGN or FS individually mainly stabilized assignments under the corresponding specified perturbation,
and changes in mean LLA-5 were small.
The mean interaction effect when the two were combined was in the direction of improvement
for LLA and LFR(TGN+FS), but in the opposite direction for LFR(FS).
Moreover, the sign of the LLA-5 difference changed across repetitions
in all four comparisons of additional perturbations,
so these effects were less consistent than the improvements due to changing the reconstruction task
(Table~\ref{tab:oof-k8-repeat-effects}).

We did not search over the mask ratio or the augmentation strengths of TGN and FS to maximize LLA.
The supplementary comparison of mask ratios is presented in Appendix~\ref{app:mask-sweep}
as a sensitivity analysis relative to the main condition of 50\% and was not used to select the main condition.

\subsection{Spatial Patterns after Full-Dataset Fitting}
\label{sec:results-global-maps}

We compared spatial maps obtained by full-dataset fitting under the six main conditions, focusing on M11.
The fitting included spectra from all 49 specimens examined in this analysis.
As an illustrative example, we present KYOw16744, a \textit{Cryptomeria japonica} specimen
from a bearing block (masu) in the main hall of Komyoji Temple
(Figure~\ref{fig:global-16744-maps}; Table~\ref{tab:specimens}).

M00 and M11 clearly delineated regions following the growth rings and continuous regions along the periphery.
Growth-ring structure comprising earlywood and latewood is visible on the measured transverse surface
(Figure~\ref{fig:specimen-surfaces}).
Parts of the perimeter of the measured transverse section correspond to surfaces of the original timber member that were exposed during service.
Given the photodegradation of wood surfaces by ultraviolet radiation~\citep{agresti2013photodegraded},
the peripheral regions corresponding to exposed surfaces are expected to differ in state from the interior.
Spatial structure consistent with this prior knowledge emerged without using spatial information
in representation learning or clustering and without label smoothing.
This correspondence supports the use of the learned spectral representations to explore state differences.

For B0 and B1, which provided strong reference conditions in OOF evaluation,
correspondence with wood anatomy was less readily discernible.
In B0, regions varying over broader spatial extents were superimposed on those following the growth rings.
In B1, the combination of labels representing the growth-ring structure varied by location.
These assignments may have strongly reflected variations in surface properties and other factors,
in addition to differences in chemical state.
Such variations can also yield high LLA if they produce spatially continuous regions;
the usefulness of the maps for interpreting the distribution of state differences
therefore needs to be considered together with qualitative evaluation based on prior knowledge.

A0 exhibited numerous fine-scale variations in assignments along the vertical scanning direction,
superimposed on the growth-ring structure.
The prominence of these variations in A0, which shared the network architecture of M00 and M11,
shows that introducing a neural network alone is insufficient
and that the design of the learning task used to organize representations is important.

A1 exhibited fewer fine-scale variations in assignments along the scanning direction than A0.
Although these variations were barely noticeable in M00,
a straight-line pattern crossing the growth-ring structure appeared in the cluster assignments near the right edge of the specimen.
This pattern was not clearly visible at the same location in M11.
Their alignment with the scanning direction suggests an influence of the measurement system.
We hypothesize that denoising training with additional perturbations reduced assignment sensitivity
to the local variations underlying the straight-line pattern while the model learned interband relationships,
contributing to both the weaker pattern in M11 and the small increase in mean OOF LLA relative to M00
(Section~\ref{sec:results-ablation}).

In KYOw16744, regions had similar extents and arrangements in the A1, M00, and M11 maps.
Other specimens showed similar patterns (Appendix~\ref{app:global-map-comparison}).
Across all 49 specimens, mean LLA-5 was 0.536 for A1 and 0.526 for M11 (Table~\ref{tab:global-k8}).
A1 had the highest LLA-5 and LLA-9, whereas M11 had the highest LLA-3.
Mean LLA-5 was higher in A1 than in A0 and higher in M11 than in M00.

The difference in spatial coherence between A1 and M11 became clear in OOF evaluation.
Mean LLA-5 for A1 exceeded that for A0 in full-dataset fitting but was lower in OOF evaluation
(Table~\ref{tab:oof-k8}).
The full-dataset maps showed how A1 and M11 organized spectral variation within the specimens used to train the encoders and fit the cluster centers.
When we used each fold's trained encoders and fitted cluster centers to segment its held-out specimens,
mean LLA-5 over the three repetitions was higher under M11 than under A1 for all 49 specimens
(Table~\ref{tab:reconstruction-effects}).
This contrast highlights the need to evaluate generalization explicitly, even in unsupervised segmentation.

When we repeated representation learning and clustering for M11 on the same data, characteristic cluster distributions persisted
and many boundaries recurred at similar locations despite changes in individual pixel assignments
(Appendix~\ref{app:spatial-map-reproducibility}).
Taken together, the spatial coherence of segmentations on unseen specimens (Section~\ref{sec:results-k8}),
the agreement of the full-dataset maps with prior knowledge, and the stability of spatial patterns across repetitions
support our use of M11 as the common basis for cross-referencing spectra and spatial distributions across specimens.
In the next section, we draw on these comparisons, published NIR band assignments, and specimen information
to interpret chemical differences among the mapped regions.

\subsection{Chemical Interpretation and Degradation Models}
\label{sec:results-spectra}
\label{sec:degradation-discussion}

\begin{arxivrighttoptable}{\number\value{page}}
  \centering
  \caption{NIR band assignments for Figure~\ref{fig:m11-nir-spectra}, based on the review by \citet{schwanninger2011assignments}.
  Appendix~\ref{app:nir-band-assignments} gives the detailed assignments and their primary sources.
  Daggers (\textdagger) mark the 13 bands used in the bar charts (Figures~\ref{fig:internal-baseline-bands} and~%
  \ref{fig:surface-baseline-bands}). A plus sign denotes a combination of vibrations.}
  \label{tab:nir-band-summary}
  \small
  \setlength{\tabcolsep}{2.5pt}
  \renewcommand{\arraystretch}{1.30}
  \newcommand{\bandrow}[3]{#1 & #2 & \raggedright #3\tabularnewline}
  \begin{tabular*}{\linewidth}{@{\extracolsep{\fill}}llp{.61\linewidth}@{}}
    \toprule
    Band & \shortstack[l]{Wavenumber\\(cm$^{-1}$)} & Principal assignment \\
    \midrule
    \bandrow{1\textsuperscript{\textdagger}}{8547}{Lignin, unsaturated C--H (HC=CH) asymmetric stretching}
    \nirassignmentrowrule
    \bandrow{2\textsuperscript{\textdagger}}{8420--8370}{Lignin (possibly also cellulose), C--H stretching}
    \nirassignmentrowrule
    \bandrow{3\textsuperscript{\textdagger}}{8250--8160}{Cellulose, C--H stretching}
    \nirassignmentrowrule
    \bandrow{4\textsuperscript{\textdagger}}{7321--7315}{Cellulose, C--H stretching +~C--H bending}
    \nirassignmentrowrule
    \bandrow{5\textsuperscript{\textdagger}}{7003--6993}{Amorphous polysaccharides (overlap with water), O--H stretching}
    \nirassignmentrowrule
    \bandrow{6\textsuperscript{\textdagger}}{6974--6954}{Water and amorphous polysaccharides, overlapping O--H stretching bands (symmetric +~asymmetric in water)}
    \nirassignmentrowrule
    \bandrow{7\textsuperscript{\textdagger}}{6757}{Semicrystalline cellulose, O--H stretching}
    \nirassignmentrowrule
    \bandrow{8}{6450}{Crystalline cellulose, hydrogen-bonded O--H stretching}
    \nirassignmentrowrule
    \bandrow{9\textsuperscript{\textdagger}}{6281}{Crystalline cellulose, strongly hydrogen-bonded O--H stretching}
    \nirassignmentrowrule
    \bandrow{10\textsuperscript{\textdagger}}{5980--5963}{Lignin (possibly also aromatic extractives), aromatic C--H stretching}
    \nirassignmentrowrule
    \bandrow{11}{5890}{Lignin, C--H stretching}
    \nirassignmentrowrule
    \bandrow{12}{5865}{Hemicellulose, C--H stretching}
    \nirassignmentrowrule
    \bandrow{13}{5795}{Lignin, C--H stretching}
    \nirassignmentrowrule
    \bandrow{14}{5776}{Cellulose, C--H stretching}
    \nirassignmentrowrule
    \bandrow{15\textsuperscript{\textdagger}}{5577}{Cellulose and lignin, C--H stretching}
    \nirassignmentrowrule
    \bandrow{16\textsuperscript{\textdagger}}{5464}{Semicrystalline and crystalline cellulose, O--H stretching +~C--O stretching}
    \nirassignmentrowrule
    \bandrow{17\textsuperscript{\textdagger}}{5220--5150}{Bound water in air-dry cell walls, O--H asymmetric stretching +~H--O--H bending}
    \nirassignmentrowrule
    \bandrow{18\textsuperscript{\textdagger}}{4392}{Cellulose, (O--H stretching +~C--C stretching) and/or (C--H stretching +~C--H bending)}
    \bottomrule
  \end{tabular*}
\end{arxivrighttoptable}

\begin{arxivwidetopfigure}{18}
  \centering
  \includegraphics[width=0.85\textwidth]{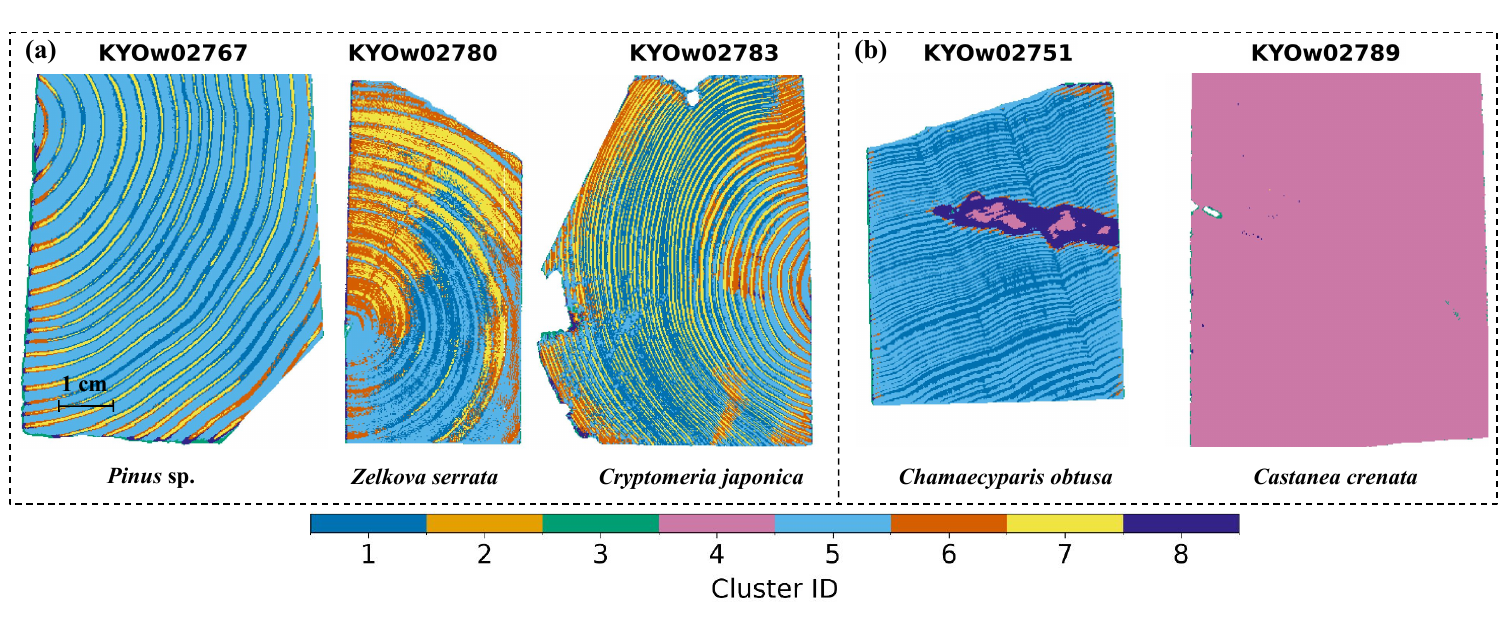}
  \caption{Spatial distributions of clusters in the interiors of historical wood specimens.
  (a) Interior regions of \textit{Pinus} sp., \textit{Zelkova serrata}, and \textit{Cryptomeria japonica} selected to examine aging.
  (b) Distribution of cluster~4 in an eaves support (left) and a column-base splice (right), examined for possible effects of moisture.
  Figure~\ref{fig:chemical-atlas} shows the spatial maps for all 49 specimens and the second-derivative spectra of selected clusters.}
  \label{fig:internal-degradation}
\end{arxivwidetopfigure}

\begin{arxivrighttopfigure}{18}
  \centering
  \includegraphics[width=0.85\columnwidth]{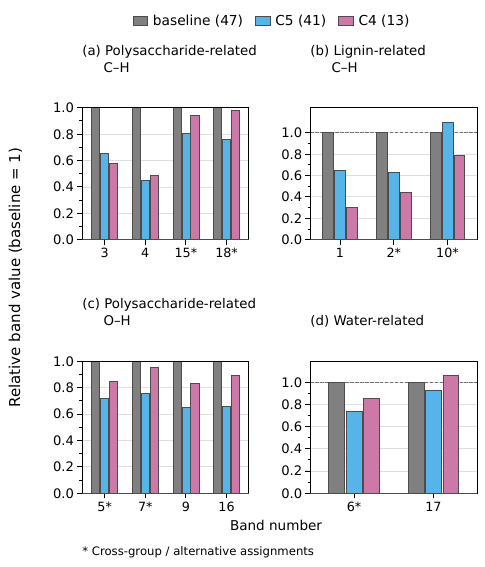}
  \caption{Relative band values for the baseline and interior clusters~5 and~4 at the 13 selected bands (baseline $=1$).
  Band numbers correspond to Table~\ref{tab:nir-band-summary}. Asterisks mark assignments that span more than one group or have multiple candidates.
  Numbers in parentheses in the legend are specimen counts.}
  \label{fig:internal-baseline-bands}
\end{arxivrighttopfigure}

\begin{arxivwidetopfigure}{21}
\centering\includegraphics[width=0.85\textwidth]{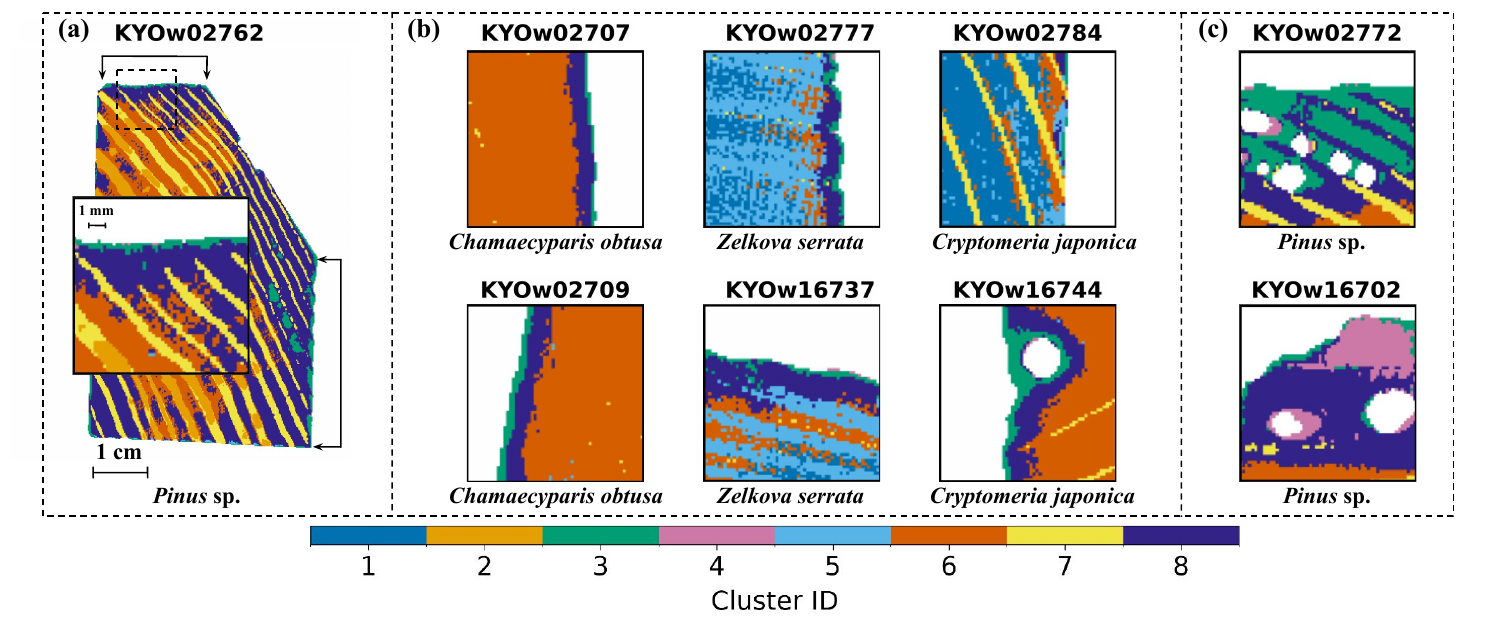}
\caption{Examples of cluster distributions suspected to reflect weathering, selected from specimens differing in source, timber member type, and wood species.
(a) Bearing block KYOw02762. The measured surface lies 10~mm inward from exposed end grain inferred to have faced outdoors. Arrows indicate edges corresponding to exposed surfaces, and the inset enlarges the dashed box.
(b, c) Enlarged regions of the measured transverse surfaces near member sides where weathering is suspected, showing six specimens in (b) and two specimens with insect holes in (c).
The inset in (a) and panels (b) and (c) share the same magnification. The 1~mm scale bar in the inset applies to all these enlarged views.}
\label{fig:weather-degradation}
\end{arxivwidetopfigure}

\begin{arxivrighttopfigure}{21}
\centering\includegraphics[width=0.85\columnwidth]{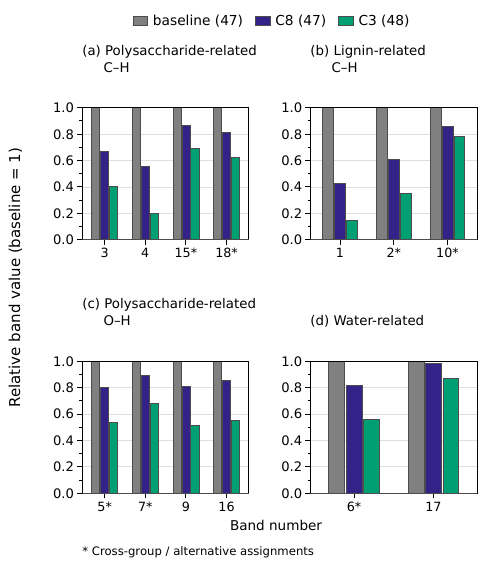}
\caption{Relative band values for the interior baseline and surface clusters~3 and~8 at the 13 selected bands (baseline $=1$).
Band numbers correspond to Table~\ref{tab:nir-band-summary}. Asterisks mark assignments that span more than one group or have multiple candidates.
Numbers in parentheses in the legend are specimen counts.}
\label{fig:surface-baseline-bands}
\end{arxivrighttopfigure}

\begin{arxivbottomwidetable}{22}
\centering
\caption{Interpretations of the M11 clusters. Band numbers correspond to Tables~\ref{tab:nir-band-summary} and~\ref{tab:nir-band-assignments}. Each row specifies the basis for comparison.}
\label{tab:cluster-interpretations}
\begingroup
  \linespread{1}\fontsize{9}{10.8}\selectfont
  \setlength{\tabcolsep}{4pt}
  \renewcommand{\arraystretch}{1.05}
  \begin{tabular*}{\linewidth}{@{\extracolsep{\fill}}lp{.215\linewidth}p{.67\linewidth}@{}}
    \toprule
    Cluster & Interpretation & Features \\
    \midrule
    1 & \raggedright Earlywood little affected by aging, or aged latewood or hardwood &
    \raggedright Cellulose O--H features (7, 9) are similar to those of cluster~6, whereas aromatic C--H (10) is more distinct. The distribution includes softwood earlywood, some latewood, and hardwood interiors (Figures~\ref{fig:internal-degradation}(a) and~\ref{fig:chemical-atlas}). \tabularnewline
    \nirassignmentrowrule
    2 & \raggedright Likely contribution from pine resin &
    \raggedright All six specimens selected for comparison of cluster~2 spectra were \textit{Pinus} sp.~(Figure~\ref{fig:chemical-atlas}). Compared with cluster~1, the aggregated cluster~2 spectrum (Figure~\ref{fig:m11-nir-spectra}) shows a stronger C--H feature at band~2 (possible CH$_3$ contribution), no discernible aromatic C--H trough at band~10, and a weaker bound-water feature at band~17. \tabularnewline
    \nirassignmentrowrule
    3 & \raggedright Surface degradation &
    \raggedright The distribution follows the outermost layer along edges inferred to have been exposed (Figures~\ref{fig:weather-degradation} and~\ref{fig:chemical-atlas}). Relative band values calculated from means across specimens are much lower than the baseline for lignin-related C--H (1, 2) and the cellulose C--H combination band (4). Possible processes include photochemical alteration, erosion and leaching, and changes in the supramolecular structure of the remaining polysaccharides. \tabularnewline
    \nirassignmentrowrule
    4 & \raggedright Region interpreted as affected by aging and moisture &
    \raggedright The distribution, position in service, and timber observations suggest past wetting (Figures~\ref{fig:internal-degradation}(b) and~\ref{fig:chemical-atlas}). Relative band values calculated from means across specimens exceed those of cluster~5 for polysaccharide-related O--H (5, 7, 9, 16) and bound water (17), while the polysaccharide-related C--H bands vary in their responses. Leaching and changes in supramolecular structure during wetting and redrying may account for these differences. \tabularnewline
    \nirassignmentrowrule
    5 & \raggedright Aged wood, or softwood compression wood &
    \raggedright The distribution includes softwood and hardwood interiors (Figures~\ref{fig:internal-degradation}(a) and~\ref{fig:chemical-atlas}). Relative band values calculated from means across specimens are lower than the baseline for polysaccharide-related C--H and O--H and lignin-related C--H (1, 2), and higher for aromatic C--H (10). Where aging is a candidate explanation, alteration of wood constituents may change the supramolecular structure and increase heterogeneity in hydrogen-bonding environments. \tabularnewline
    \nirassignmentrowrule
    6 & \raggedright Earlywood little affected by aging, or aged latewood or hardwood &
    \raggedright Cellulose O--H features (7, 9) are similar to those of cluster~1, whereas aromatic C--H (10) is weaker. The distribution includes softwood earlywood, some latewood, and hardwood interiors (Figures~\ref{fig:internal-degradation}(a) and~\ref{fig:chemical-atlas}). \tabularnewline
    \nirassignmentrowrule
    7 & \raggedright Latewood or hardwood little affected by aging &
    \raggedright Cellulose-related features (3, 4, 7, 9, 16, 18) are more distinct than in clusters~1 and~6. The distribution includes softwood latewood and hardwood interiors (Figures~\ref{fig:internal-degradation}(a) and~\ref{fig:chemical-atlas}). \tabularnewline
    \nirassignmentrowrule
    8 & \raggedright Transition region associated with surface degradation, where it lies inward of cluster~3 &
    \raggedright Relative band values calculated from means across specimens lie between the baseline and cluster~3 for the 13 selected bands, with bound water (17) close to the baseline. Leaching, wetting, and redrying may be accompanied by the supply and retention of acidic products from weathering in the outermost layer, transported by water, and by secondary reactions upon rewetting (Figures~\ref{fig:weather-degradation} and~\ref{fig:chemical-atlas}). \tabularnewline
    \bottomrule
  \end{tabular*}
\endgroup

\end{arxivbottomwidetable}

We used the published band assignments summarized in Table~\ref{tab:nir-band-summary}
to interpret the aggregated second-derivative spectra of the M11 clusters (Figure~\ref{fig:m11-nir-spectra}).
We compared these spectral features with the cluster distributions and spectra in individual specimens
(Figure~\ref{fig:chemical-atlas}), taking into account information on the original timber members.
Similarities and differences within and across specimens informed our interpretation of chemical states in the wood interior and at the surface.
Table~\ref{tab:cluster-interpretations} summarizes the interpretation of each cluster and the evidence supporting it.
All six specimens selected for the comparison of cluster~2 spectra were \textit{Pinus} sp., and the spectral features suggest a contribution from pine resin, so we exclude this cluster from the degradation models.

Band values reflect both the contributions of wood constituents and their molecular environments. They do not directly measure constituent abundance.
We refer to cellulose and hemicellulose collectively as polysaccharides, and to constituents containing aromatic rings, such as lignin and phenolic extractives, as aromatic constituents.

In interpreting the cellulose O--H bands, we consider both the contribution of cellulose and the hydrogen-bonding partners of its hydroxyl groups in the cell wall.
Cellulose chains form microfibrils, and multiple microfibrils form bundles within the cell wall.
The hydrogen bonds that the hydroxyl groups form can vary with the local arrangement of the chains,
the relative positions of adjacent microfibrils within a bundle, and their positions relative to surrounding polymers.
We refer to these arrangements collectively as the \emph{supramolecular structure of the cell wall}.
Because the specimens were measured in the air-dry state, we interpret band~17 as a feature of bound water retained in the cell walls.

The spatial distribution of chemical states may also reflect the pathways by which water entered the timber during service and transported wood constituents.
The orientation of wood cells makes water penetration direction-dependent.
The longitudinal (L) direction follows the grain, along the long axes of tracheids and vessels.
The radial (R) direction extends outward from the pith, and the tangential (T) direction is tangent to the growth rings.
Free water can flow along cell lumina and through pits between cells. It generally penetrates wood more readily in the L direction than in the R or T direction~\citep{glass2010moisture}.
On the measured transverse surfaces, L is perpendicular to the plane of observation, and R and T lie within it.

\phantomsection
\paragraph{Internal Degradation}
\label{sec:internal-degradation}

Cluster~7 occurred in softwood latewood and in the interior of hardwood specimens. Its spectra showed a pronounced second overtone of cellulose C--H stretching (band~3), first overtones of O--H stretching (7 and~9), and combination bands (4, 16, and~18) associated with cellulose
(Figures~\ref{fig:m11-nir-spectra}, \ref{fig:internal-degradation}(a), and~\ref{fig:chemical-atlas}).
These features were weaker in clusters~1 and~6, which occurred in softwood earlywood, in parts of the latewood, and in hardwood interiors.
Based on these distributions and the wood anatomy, we interpreted cluster~7 as latewood or hardwood relatively little affected by aging.
Clusters~1 and~6 could represent earlywood relatively little affected by aging, or latewood or hardwood affected by aging. The evidence did not establish which of the two clusters was more affected.

Cluster~5 also occurred in softwood earlywood and hardwood interiors (Figures~\ref{fig:internal-degradation}(a) and~\ref{fig:chemical-atlas}).
In the aggregated second-derivative spectra, the polysaccharide-related C--H bands (3, 4, 15, and~18) and O--H bands (5, 7, 9, and~16) were generally weaker in cluster~5 than in clusters~1, 6, and~7, whereas the aromatic C--H band (10) remained relatively distinct (Figure~\ref{fig:m11-nir-spectra}).
These spectral features and their occurrence in the interior suggest that cluster~5 may represent wood affected by aging.
In softwoods, compression wood is rich in lignin and differs in polysaccharide composition~\citep{chen2007compression}. It could produce similar features (Figure~\ref{fig:chemical-atlas}), so pre-existing anatomical differences are another possible explanation.

Cluster~4 occupied a localized interior region in the eaves support KYOw02751 and much of the measured surface in the column-base splice KYOw02789 (Figure~\ref{fig:internal-degradation}(b)).
The end grain of the eaves support likely faced outward beneath the roof. Rainwater entering through the end grain could account for the localized distribution.
The column-base splice rested directly on a foundation stone, and the back of the specimen showed severe deterioration.
The use of Japanese chestnut (\textit{Castanea crenata}), a species with naturally decay-resistant heartwood~\citep{matsuoka1970durability}, is also consistent with the damp conditions expected at this location.
These distributions, the positions of the members during service, and observations of the timbers led us to interpret cluster~4 as a state shaped by both aging and moisture exposure.

We chose clusters~1, 6, and~7 for the baseline because their cellulose-related features were more distinct than those of cluster~5 and their distributions spanned earlywood, latewood, and hardwood interiors.
We treated these clusters as candidates for relatively sound interior wood. The baseline provides a relative reference, and the wood it represents may itself have aged.
We compared cluster~5 with this baseline to examine the effects of aging, and clusters~5 and~4 with each other \mbox{to examine the possible role of water (Figure~\ref{fig:internal-baseline-bands}).}

In all 41 specimens in which cluster~5 could be compared with the baseline, the specimen-specific relative band value for O--H in amorphous polysaccharides (band~5) was lower in cluster~5 than in the baseline for the same specimen.
For the bands associated with semicrystalline and crystalline cellulose, cluster~5 had lower values at both O--H stretching bands (7 and~9) in all 41 specimens and at the O--H and C--O stretching combination band (16) in 40 of 41.
Values for the cellulose-related C--H stretching band (3) and combination band (4) were also lower in cluster~5 in all 41 specimens.
Relative band values calculated from the means across specimens fell below the baseline at every polysaccharide-related C--H band (3, 4, 15, and 18) and O--H band (5, 7, 9, and 16) (Figure~\ref{fig:internal-baseline-bands}).
The corresponding values for lignin-related C--H bands~1 and~2 also fell below the baseline. The value for aromatic C--H (band~10) exceeded the baseline, although the direction of the difference varied among specimens.

In second-derivative NIR transmission spectra, \citet{tsuchikawa2005ageing} observed a shallower trough at 7000~cm$^{-1}$ (within band~5) in wood from a historical \textit{Chamaecyparis obtusa} pillar than in modern wood of the same species.
They attributed the weaker O--H band to a decrease in amorphous cellulose.
In NIR reflectance spectra, \citet{yonenobu2003antique} found no clear difference between historical and modern \textit{C.~obtusa} at the O--H bands associated with semicrystalline and crystalline cellulose (bands~7 and~9 here).
A compositional analysis of \textit{C.~obtusa} and \textit{Zelkova serrata} from historical buildings in Japan found lower holocellulose contents in older wood~\citep{kohara1955oldtimbers}, consistent with the generally weaker polysaccharide-related C--H and O--H features in cluster~5.
Selective loss of amorphous polysaccharides alone is difficult to reconcile with the lower band values for cluster~5 at bands~7, 9, and~16 in nearly all of our within-specimen comparisons of interior regions.
Because the width of an absorption band also affects the depth of its second-derivative trough, we considered the hydrogen-bonding environment of O--H groups in the remaining cellulose.

Structural analyses of spruce wood suggest that cellulose chain packing and hydrogen bonding are more heterogeneous near the surface of an individual microfibril than in its core.
From their solid-state NMR results, the authors inferred that disordered cellulose chains were, on average, closer to regions containing hemicelluloses and lignin than well-ordered chains were~\citep{fernandes2011microfibrils}.
As hemicelluloses and lignin change during aging, their sites of contact with cellulose and other polymers may shift even when these constituents remain in the cell wall. Some of the intermolecular interactions that maintained their original arrangement, including hydrogen bonds, may then be lost.
We propose that this relaxes constraints both on cellulose chains near the microfibril surface and on the arrangement of adjacent microfibrils within a bundle.
Changes in the relative positions or orientations of neighboring chains within a microfibril could alter interchain hydrogen-bond distances and angles.
At the interface between adjacent microfibrils, changes in their spacing or in the orientation of their contacting faces could likewise alter the geometry in which an O--H group on one face can form a hydrogen bond with an oxygen atom on the other.
Greater variation in bond strength at both sites could broaden the O--H absorption bands. In cluster~5, this broadening, together with a smaller polysaccharide contribution, could account for the shallower second-derivative troughs at bands~7, 9, and~16.

In the ten specimens in which clusters~4 and~5 could be compared directly, specimen-specific relative band values were higher in cluster~4 than in cluster~5 in eight specimens at band~7, eight at band~9, and seven at band~16. All three bands involve O--H stretching in semicrystalline or crystalline cellulose.
The value for bound water in the air-dry specimens (band~17) was higher in cluster~4 in nine of the ten specimens. Five specimens had higher values in cluster~4 at all four bands.
Relative band values calculated from the means across specimens were higher in cluster~4 than in cluster~5 at all polysaccharide-related O--H bands (5, 7, 9, and 16). At band~17, cluster~4 exceeded both cluster~5 and the baseline (Figure~\ref{fig:internal-baseline-bands}).
Among the polysaccharide-related C--H bands, cluster~4 had a lower relative band value at band~3, a similar value at band~4, and higher values at bands~15 and~18. Lignin C--H may also contribute to band~15, and band~18 may include a combination involving O--H.

The interior distribution of cluster~4 and the positions of the timbers in service suggest that water entering the wood may have flowed through cell lumina as free water, carrying soluble constituents produced during aging.
In beech that had been waterlogged for an extended period, solid-state NMR showed a smaller carbohydrate contribution relative to lignin than in modern beech. The authors attributed part of this difference to leaching of solubilized sugars~\citep{bardet2002archaeological}.
If leaching also changes cell-wall composition in structural timbers subject to intermittent water ingress, the O--H features of the remaining cellulose may become relatively more prominent.
Among the bands assigned to cellulose, those involving O--H stretching in semicrystalline or crystalline regions (7, 9, and 16) were stronger in cluster~4 than in cluster~5. By contrast, the C--H feature at band~3 was weaker in cluster~4, and the two clusters differed little at band~4.
To explain these differences among bands, we considered both changes in composition and the effect of the hydrogen-bonding environments of the remaining cellulose chains on absorption band shapes.

Historical poplar timber had a lower X-ray diffraction crystallinity index and a larger apparent crystallite diameter than modern wood of the same genus. The authors attributed the difference in crystallite diameter to local rearrangement of cellulose chains and also reported a higher equilibrium moisture content in the historical wood~\citep{guo2018ageing}.
For cluster~4, we assume that the relaxation of constraints during aging increased the variability in the relative positions and orientations of neighboring cellulose chains and in the orientations and spacing of microfibrils within a bundle.
In this state, repeated swelling and shrinkage of the cell wall during wetting and redrying could allow cellulose chains and microfibrils to move more readily. On redrying, some chains may rearrange, and hydrogen bonds may re-form between chains that move closer together.
Where adjacent microfibrils move closer together and become more regularly oriented and spaced, their hydroxyl-bearing faces may come into contact, allowing hydrogen bonds to re-form across the interface.
If redrying thus restores local order to arrangements disrupted during aging, hydrogen-bond distances and angles may be more uniform than in cluster~5, both between cellulose chains and between microfibrils.
A narrower range of hydrogen-bond strengths would also narrow the range of O--H stretching frequencies, concentrating absorption at each of bands~7, 9, and~16 over a narrower wavelength range and potentially producing deeper second-derivative troughs than in cluster~5.

In a deuterium-exchange NIR study of historical \textit{C.~obtusa} wood, the water band was weaker than in modern wood, yet a larger fraction of the remaining O--H groups in amorphous and semicrystalline regions exchanged with D$_2$O at saturation.
The authors suggested that changes in the relative arrangement of cellulose microfibrils could increase the fraction of remaining O--H groups accessible to water even as the total number of water sorption sites declined~\citep{tsuchikawa2005ageing}.
In wood assigned to cluster~4, relaxation of the constraints between microfibrils could open gaps through which water reaches more of the remaining O--H groups inside a bundle. Leaching of soluble constituents could expose additional O--H groups.
Bound water retained at such interfaces in the air-dry state could help explain why the value for band~17 was higher in cluster~4 than in cluster~5.

On this basis, we propose the following model of internal degradation, in which age-related changes in cell-wall constituents provide a common starting point.
During aging, slow hydrolysis and oxidation in cell walls containing sorbed moisture may alter the hemicelluloses and lignin surrounding individual cellulose microfibrils and their bundles.
We hypothesize that acetic acid released by hemicellulose deacetylation~\citep{hudson2020historic} can promote further hydrolysis of polysaccharide backbones if the acid remains in the wood. Some ether linkages in condensed lignin or lignin--carbohydrate complexes may also undergo hydrolysis~\citep{guo2018ageing}.
Even if the altered constituents remain in the cell wall, shifts in where the polymers contact one another and which groups form hydrogen bonds may disrupt some of the intermolecular interactions that maintain their arrangement. This would relax the constraints on cellulose chains near the microfibril surface. The surrounding polymers may also provide less support for a bundle, allowing adjacent microfibrils within it to change their relative positions and orientations more readily.
Greater variability in the relative positions and orientations of neighboring chains within the same microfibril could increase the variation in interchain hydrogen-bond distances and angles. Less regular orientations and spacing of the microfibrils within a bundle could likewise increase the variation in hydrogen-bond distances and angles across their interfaces.
The resulting broadening of NIR O--H absorption bands, together with a smaller contribution from polysaccharides, could produce shallower second-derivative O--H troughs associated with semicrystalline and crystalline regions.

Where water enters the wood through the end grain along the L direction or through cracks and joints, as may occur during rainfall, soluble constituents produced during aging can dissolve out of the cell wall and be carried away by free water flowing through the cell lumina. This leaching changes the composition of the material that remains in the cell wall.
Where aging has relaxed the constraints on cellulose, swelling and shrinkage of the cell wall during wetting and redrying can change both the relative positions and orientations of neighboring chains and the spacing and orientations of microfibrils within a bundle.
During redrying, some chains may rearrange and hydrogen bonds may re-form between neighboring chains. Where constraints have relaxed, cell-wall shrinkage may also draw adjacent microfibrils closer together and tighten the bundle locally.
If this partially restores the order disrupted during aging and brings opposing hydroxyl-bearing faces into contact, hydrogen bonds may also re-form across the interface.
A narrower range of distances and angles among the re-formed hydrogen bonds would reduce the variation in bond strength. This could narrow each absorption band to which O--H stretching in semicrystalline or crystalline regions contributes, producing more distinct troughs in the second-derivative NIR spectra.
At the same time, relaxation of the constraints between microfibrils may open gaps to water, increasing the fraction of remaining O--H groups accessible even inside a bundle. Leaching of nearby soluble constituents may expose further O--H groups to water.
These more accessible interfaces can coexist with locally ordered regions and may retain bound water in the air-dry state.

\paragraph{Surface Degradation}
\label{sec:surface-degradation}
Even in specimens from different sources and timber member types, clusters~3 and~8 formed contiguous regions within a few millimeters of edges inferred to correspond to surfaces exposed during service
(Figures~\ref{fig:weather-degradation}(b) and~\ref{fig:chemical-atlas}).
For cluster~3 in the outermost layer, the relative band values calculated from the means across specimens for lignin-related C--H bands~1 and~2 were substantially below the interior baseline (Figure~\ref{fig:surface-baseline-bands}).
Lignin oxidation has been suggested in the surface layers of naturally weathered wood~\citep{sudiyani1999weathering}.
The occurrence of cluster~3 along exposed surfaces and the weaker bands~1 and~2, consistent with lignin alteration during natural weathering, led us to interpret it as a region strongly affected by weathering.

In cluster~3, the relative band value for the cellulose C--H combination band~4 was also substantially below the interior baseline (Figure~\ref{fig:surface-baseline-bands}).
In naturally weathered radiata pine veneers, the solution viscosity of holocellulose isolated from the entire veneer thickness decreased. The authors interpreted this as a decrease in the degree of polymerization of cellulose~\citep{evans1996weathering}.
We consider that both leaching of soluble fractions produced by polysaccharide depolymerization and changes in the supramolecular structure of the polysaccharides remaining in the cell wall contributed to the weaker band~4 in cluster~3.

In some specimens, cluster~8 occurred immediately inward of cluster~3 (Figure~\ref{fig:weather-degradation}(b)).
The relative band values for cluster~8 calculated from the means across specimens were below the baseline and above cluster~3 at all 13 selected bands (Figure~\ref{fig:surface-baseline-bands}).
In both clusters~3 and~8, the relative band values within the polysaccharide-related C--H group were higher at bands~15 and~18 than at bands~3 and~4.
In cluster~8, the polysaccharide-related O--H bands (5, 7, 9, and~16) had relative band values below the baseline, whereas band~17, attributed to bound water in the air-dry specimens, was close to the baseline.
Band~6, where water and polysaccharide absorptions overlap, weakened similarly to the adjacent polysaccharide-related band~5.

The baseline and both clusters could be compared within the same specimen in 46 specimens.
The number of specimens in which the relative band value decreased in the order baseline, cluster~8, and cluster~3 ranged from 41 to 46 for cellulose-related C--H (3) and polysaccharide-related O--H (5, 7, and~9), and was 30 for the O--H and C--O combination band (16).
The corresponding counts for aromatic C--H (10) and bound water (17) were 25 and~24. In some specimens, cluster~8 exceeded the baseline for that specimen at these bands.
The tendency for polysaccharide-related features to lie between those of cluster~3 and the interior baseline, together with the location immediately inward of cluster~3, led us to interpret cluster~8 in this spatial arrangement as a transition region in chemical state associated with weathering of the outermost layer.

In Figure~\ref{fig:weather-degradation}(b), clusters~3 and~8 occurred within a few millimeters of the specimen edges (Figure~\ref{fig:chemical-atlas}).
Where a radial or tangential face had been exposed, this width corresponds to the depth inward from the side of the timber member.
Kataoka et al. exposed radial faces of \textit{Cryptomeria japonica} and \textit{Chamaecyparis obtusa} earlywood to xenon light for 50--1500~h. They found a logarithmic relationship between irradiation time and the depth at which a change in the FT-IR carbonyl band at 1730~cm$^{-1}$ could be detected~\citep{kataoka2005density}.
They estimated that the energy received during 1500~h of irradiation was approximately equivalent to 10--11 months of natural exposure on a horizontal surface in Tokyo.
Using this conversion to extrapolate the regressions in their Figure~4 from the equivalent of 100 years to 400 years gives depths increasing from approximately 1.4 to 1.6~mm for \textit{C.~japonica} earlywood and from 0.9 to 1.1~mm for \textit{C.~obtusa} earlywood.
Because the actual light exposure and erosion by wind and rain at the sides of the timber members are unknown, we used these estimates only to compare the order of magnitude of the depths.
The occurrence of a weathered outermost layer and an adjacent transition region within a few millimeters of the side is compatible with this estimate.

In Figure~\ref{fig:weather-degradation}(a), however, cluster~8 occupied a broad area of the measured surface located 10~mm inward from an end-grain surface inferred to have been exposed.
In Figure~\ref{fig:weather-degradation}(c), clusters~3 and~8 occurred around insect holes in KYOw02772, and cluster~8 extended into an interior region containing insect holes in KYOw16702.
The depth of photodegradation from an exposed side face alone is unlikely to explain the occurrence of cluster~8 on the plane 10~mm inward from exposed end grain or its inward extent in regions containing insect holes.
We therefore considered water ingress through the end grain or insect galleries as a factor common to these two cases.

Where end grain is exposed, water can enter open cell lumina and penetrate deeper along the L direction.
Where insect damage occurs near a surface, galleries connected to the exterior of the original timber member can provide routes for water ingress.
The entrance need not lie on the measured transverse surface. A gallery extending out of the observation plane may connect to the exterior elsewhere.
Water entering through end grain or insect galleries may thus have helped shape the chemical state of the wood in regions assigned to cluster~8 at these locations.

If water enters a transition region where the direct effects of light and raindrops are weak, we might expect aging and moisture exposure to produce a chemical state similar to that inferred for regions assigned to cluster~4.
Figure~\ref{fig:weather-degradation}, however, includes examples in which cluster~8 lies inward of cluster~3.
Among the 12 specimens in which clusters~4 and~8 could be compared within the same specimen, the specimen-level relative band value for aromatic C--H (10) was lower in cluster~8 in nine specimens. In contrast, the values for lignin-related C--H bands~1 and~2 were lower in cluster~4 in ten specimens each.
For the polysaccharide-related O--H group and bound water in the air-dry specimens (17), which cluster had the higher value varied among specimens.

We propose that leaching and transport of soluble constituents by free water, together with wetting and redrying of the cell wall, helped shape the chemical states represented by clusters~4 and~8.
In regions assigned to cluster~4, free water flowing through cell lumina can leach soluble constituents produced during aging, changing the sites of contact between the polymers remaining in the cell wall.
We proposed that these changes, combined with local rearrangement during wetting and redrying, alter the supramolecular structure of the remaining constituents.
Where regions assigned to cluster~8 adjoin the weathered outer layer (cluster~3), water entering the wood may also carry acidic products from that layer into these regions.
Spectra measured before and after water extraction of photodegraded wood showed that photodegradation products containing carboxyl groups could be removed by water~\citep{tolvaj1995photodegradation}.
In artificial weathering with light and water spray, changes in lignin were also observed on the lumen side of the outermost tracheid walls. The authors discussed erosion of the thickened centers of bordered pit membranes (tori). They attributed the lignin changes in part to leaching of degradation products by spray water entering the lumina through microcracks originating at the pit apertures~\citep{kanbayashi2018weathering}.
If some products remain in the cell wall after drying interrupts water flow, they may promote secondary reactions, including hydrolysis of cell-wall constituents, upon rewetting.
Secondary reactions superimposed on these shared processes may help explain why the chemical state of the transition region differs from that of the wood assigned to cluster~4.

On this basis, we propose the following model of surface degradation.
In the outermost layer, light alters lignin and other constituents, and wind and rain erode the weakened tissue. Rainwater leaches soluble lignin-derived products and soluble fractions of depolymerized polysaccharides. We propose that these losses, together with changes in the supramolecular structure of the remaining polysaccharides, weaken both lignin-related and polysaccharide-related bands.
Some soluble acidic products formed by weathering of the outermost layer leave the timber with rainwater, while others are transported inward by water entering the wood. A transition region in chemical state associated with weathering of the outermost layer forms where the transported acidic products remain in the cell wall and promote secondary reactions upon rewetting.

The transition region shares the basic processes described in the \hyperref[sec:internal-degradation]{internal degradation model}, including age-related alteration of polysaccharides and lignin, leaching of soluble constituents by free water, and local rearrangement during wetting and redrying of the cell wall.
An additional process supplies acidic products formed by weathering of the outermost layer through water entering the wood. Products retained in the cell wall may promote secondary reactions upon rewetting, including hydrolysis of glycosidic linkages in polysaccharides and some ether linkages involving lignin.
Because photodegradation from direct irradiation and erosion by wind and rain are less pronounced in the transition region, polysaccharide-related spectral features are expected to be better preserved there than in the outermost layer.
Changes in the supramolecular structure of the remaining constituents can make hydrogen-bonding environments more heterogeneous, broadening polysaccharide-related O--H absorption bands in the NIR and producing shallower second-derivative troughs.
As in the internal degradation model, regions with increased water accessibility to the remaining O--H groups may retain bound water in the air-dry state.

The depth of the transition region depends on the anisotropy of water penetration and the pathways from exposed surfaces into the timber.
Water entering inward from a radial face travels mainly in the T direction, while water entering from a tangential face travels mainly in the R direction. Both directions lie within the measured transverse plane and cross the long axes of the cells.
The lumina of tracheids and vessels extend along the grain, and water moves more readily along these lumina in the L direction than in the R or T direction. Thus, when water entering from a side face in the R or T direction reaches cell lumina through pits or microcracks enlarged by wetting and redrying, some of it may be diverted into the L direction, perpendicular to the measured transverse plane.
We propose that this diversion limits how far the products supplied from the side face reach into the timber.
Where end grain is exposed, water enters in the L direction and can transport products deeper than when it enters in the R or T direction. If an insect gallery extending out of the measured transverse plane connects to the exterior elsewhere on the timber member, products can also be supplied along that route to the vicinity of insect holes that lie away from the edges in the observed plane.
Regardless of the entry route, uptake and retention of some acidic products in the cell walls reduce the amount supplied farther inward.
When drying interrupts water movement, product transport also ceases, and some of the products retained in the cell wall can remain locally.
Upon rewetting, these residual products promote secondary reactions such as hydrolysis.
We propose that the water entry routes and diversion of flow, combined with retention of products and interruption of transport by drying, alter the extent of secondary reactions and give rise to transition regions of different depths.

\section{Conclusions}
\label{sec:conclusion}

Where a chemical state occurs, how far it extends, and whether it recurs in specimens with different histories can offer clues to how it formed.
We examined this question in historical timber, where centuries of environmental exposure are superimposed on chemical differences among anatomical regions.
To compare these distributions without defining chemical-state categories in advance, we developed ChemoMAE, a self-supervised method that exploits dependencies among near-infrared (NIR) bands through masked denoising.
We applied ChemoMAE to pixel spectra obtained by near-infrared hyperspectral imaging (NIR-HSI) and used the learned representations for unsupervised segmentation.
Across the cluster counts examined, ChemoMAE consistently yielded higher mean Local Label Agreement (LLA) on held-out specimens than direct use of the input spectra, PCA, and the conventional autoencoder baselines.
Additional perturbations designed to preserve the zero mean and fixed norm of spectra after standard normal variate (SNV) transformation contributed little to this LLA advantage.
They may nevertheless help suppress fine-scale artifacts in spatial maps, including linear patterns associated with the scanning direction.

We used prior knowledge of wood anatomy and degradation to specify a property expected of a useful spectral representation.
If regions of similar chemical state span neighboring pixels and the spectral features that distinguish those regions are shared across specimens, clustering spectra alone should yield spatially coherent regions in an unseen specimen.
We therefore withheld spatial information from both representation learning and clustering and used it to test this prediction.
We formulated LLA to measure neighboring-label agreement after correcting for chance at fixed cluster sizes and showed that it equals one minus the variance-weighted mean of Geary's \(C\) across binary cluster indicators.
Specimen-level cross-validation revealed that the unmasked denoising autoencoder achieved spatial coherence comparable to ChemoMAE on the specimens used for fitting, yet generalized less well to held-out specimens.
The comparison with PCA also depended on the evaluation criterion.
PCA gave a higher cosine-silhouette score, while ChemoMAE gave higher mean LLA on held-out specimens.
For chemical interpretation using hyperspectral images, the emergence of spatial structure consistent with material knowledge in unseen specimens thus provides one basis for evaluating a learned representation.

We learned the representation and fitted common cluster centers anew across 49 specimens representing seven wood taxa from several historical buildings in Japan.
The resulting atlas links each specimen's provenance, timber member type, and wood taxon to its spatial map and regional second-derivative NIR spectra.
By cross-referencing specimens, we identified similarities and differences in the spatial distributions of clusters and interpreted them in light of their spectra, the members' service histories, and prior studies.
These comparisons led us to propose two degradation models.
In the internal degradation model, aging makes hydrogen-bonding environments around cellulose microfibrils more heterogeneous.
Where water enters, leaching changes the constituents that remain, and repeated wetting and drying may promote local rearrangement.
In the surface degradation model, water carries soluble acidic products from the weathered outer layer into the wood.
Products retained in cell walls may promote secondary reactions upon rewetting, giving rise to a transition region beneath the surface.
Together, the atlas and the models offer a guide to where to look and what to test in other historical timbers.
The atlas can help researchers locate regions potentially affected by weathering, choose depths for chemical sampling, or select sites from which to cut specimens for mechanical testing.
The models suggest what to examine at those sites.
For example, anatomically comparable interior regions with and without indications of past water ingress could be sampled to examine whether differences in apparent cellulose crystallinity measured by X-ray diffraction are accompanied by differences in hydroxyl accessibility measured by deuterium exchange.
Sampling across the boundary of a mapped transition region beneath a weathered surface could test whether water extracts from that region have a lower pH than those from the adjacent interior.

In this exploratory AI for Science framework, self-supervised learning gives researchers a common basis for comparing observations whose chemical meaning is initially uncertain.
An interpretation suggested by a region's spectrum can be examined against where similar features recur in other specimens, how far they extend, and which regions they adjoin.
Shared patterns can support an interpretation, while differences can prompt researchers to ask what additional process might explain a particular distribution.
The open-source \texttt{chemomae} library, experimental code, and configurations make the learning and evaluation procedures available for studies of other materials.
In historical wood, we demonstrated how researchers can evaluate patterns learned from unlabeled spectra against expectations drawn from material knowledge, then connect observations across diverse specimens with prior findings to formulate testable hypotheses about how material states arise.

\section*{CRediT authorship contribution statement}

\textbf{Ryuji Yamaguchi}: Conceptualization, Data curation, Formal analysis, Investigation, Methodology,
Software, Validation, Visualization, Writing -- original draft.
\textbf{Miyuki Matsuo}: Data curation, Funding acquisition, Investigation, Project administration,
Supervision, Writing -- review \& editing.

\section*{Data availability}

Public release of the NIR-HSI data requires permission from the holding institution.
The data are available from the corresponding authors upon request, subject to that permission.

The \texttt{chemomae} Python package is available from \href{https://pypi.org/project/chemomae/}{PyPI} under the Apache License 2.0.
The \href{https://github.com/Mantis-Ryuji/ChemoMAE}{ChemoMAE repository} provides the source code, documentation, and usage examples.
The \href{https://github.com/Mantis-Ryuji/WoodDegradationMap}{WoodDegradationMap repository} contains the experimental code and configuration files. The code is distributed under the MIT License.

\section*{Declaration of competing interest}

The authors declare that they have no known competing financial interests or personal relationships that could have appeared to influence the work reported in this paper.

\section*{Acknowledgments}

This work was supported by the Database for the Humanosphere (Xylarium) collaborative program at Kyoto University’s Research Institute for Sustainable Humanosphere, FY2025 MUSUBIME, and JSPS KAKENHI Grant Number 26K04403. 
We are grateful to Dr. Suyako Tazuru and Mr. Hajime Sorimachi for preparing the wood specimens.
We thank Dr. Masashi Nakamura for helpful discussions and Dr. Tetsuya Inagaki and Dr. Satoru Tsuchikawa of Nagoya University
for lending us the Compovision NIR-HSI system.

\section*{Declaration of generative AI and AI-assisted technologies in the manuscript preparation process}
\label{sec:ai-use}

We used OpenAI Codex to search the literature, translate and revise the manuscript, and edit the \LaTeX{} source.
Section~\ref{sec:mapping} describes how we used Codex to help write research code.
We verified and edited its output and take full responsibility for the research findings and the content of this manuscript.

\bibliography{refs}

\appendix
\numberwithin{equation}{section}
\numberwithin{figure}{section}
\numberwithin{table}{section}

\onecolumn
\section{Specimen Metadata and Overview}
\label{app:specimen-overviews}

\subsection{Specimen Metadata}
\label{app:specimen-metadata}

\begin{table}[!ht]
\centering
\caption{Metadata for the 49 specimens. Japanese names for the timber member types are given in parentheses.}
\label{tab:specimens}
\fontsize{10}{11}\selectfont
\setlength{\tabcolsep}{2.5pt}
\renewcommand{\arraystretch}{0.98}
\begin{tabular}{@{}lllll@{}}
\toprule
Specimen ID & Site & Building & Timber member type & Wood taxon \\
\midrule
\texttt{KYOw02702} & Horyuji & Five-story pagoda & Continuous bracket arm (t\=oshi-hijiki) & \textit{Chamaecyparis obtusa} \\
\texttt{KYOw02707} & Gokurakuin & Main hall & Lintel-level tie beam (uchinori-nuki) & \textit{Chamaecyparis obtusa} \\
\texttt{KYOw02708} & Gokurakuin & Main hall & Column (hashira) & \textit{Chamaecyparis obtusa} \\
\texttt{KYOw02709} & Gokurakuin & Main hall & Column (hashira) & \textit{Chamaecyparis obtusa} \\
\texttt{KYOw02715} & Byodoin & Phoenix Hall & Roof framing timber (koyazai) & \textit{Chamaecyparis obtusa} \\
\texttt{KYOw02716} & Hokkeji & Main hall & Bracket arm (hijiki) & \textit{Chamaecyparis obtusa} \\
\texttt{KYOw02717} & Daihoonji & Main hall & Column (hashira) & \textit{Chamaecyparis obtusa} \\
\texttt{KYOw02719} & Rengeoin & Main hall & Longitudinal beam (keta) & \textit{Chamaecyparis obtusa} \\
\texttt{KYOw02720} & Horyuji & Five-story pagoda & Rafter (taruki) & \textit{Chamaecyparis obtusa} \\
\texttt{KYOw02751} & Konchi-in & Hasso-seki tearoom & Eaves support (kayaoi) & \textit{Chamaecyparis obtusa} \\
\texttt{KYOw02752} & Kenninji & Imperial Envoy Gate & Rafter (taruki) & \textit{Chamaecyparis obtusa} \\
\texttt{KYOw02754} & Myohoin & \=Oshoin & Edging timber (fuchigi) & \textit{Chamaecyparis obtusa} \\
\texttt{KYOw02756} & Tomyoji & Main hall & Head tie beam (kashiranuki) & \textit{Pinus} sp. \\
\texttt{KYOw02758} & Hokkeji & Main hall & Roof strut (koyazuka) & \textit{Pinus} sp. \\
\texttt{KYOw02759} & Daihoonji & Main hall & Penetrating tie beam (nuki) & \textit{Pinus} sp. \\
\texttt{KYOw02760} & Kyo-o-gokokuji & Great South Gate & Bearing block (masu) & \textit{Pinus} sp. \\
\texttt{KYOw02762} & Kyo-o-gokokuji & Lecture hall & Bearing block (masu) & \textit{Pinus} sp. \\
\texttt{KYOw02763} & Enman-in & Shinden & Floor joist (neda) & \textit{Pinus} sp. \\
\texttt{KYOw02764} & Enman-in & Shinden & Roof framing timber (koyazai) & \textit{Pinus} sp. \\
\texttt{KYOw02766} & Hokkeji & Main hall & Large bearing block (daito) & \textit{Pinus} sp. \\
\texttt{KYOw02767} & Hokkeji & Main hall & Roof purlin (moya) & \textit{Pinus} sp. \\
\texttt{KYOw02768} & Konchi-in & Hasso-seki tearoom & Longitudinal beam (keta) & \textit{Pinus} sp. \\
\texttt{KYOw02769} & Daihoonji & Main hall & Roof framing timber (koyazai) & \textit{Pinus} sp. \\
\texttt{KYOw02770} & Myohoin & \=Oshoin & Penetrating tie beam (nuki) & \textit{Pinus} sp. \\
\texttt{KYOw02771} & Ninnaji & Five-story pagoda & Eaves cantilever (hanegi) & \textit{Pinus} sp. \\
\texttt{KYOw02772} & Ninnaji & Five-story pagoda & Cantilever fulcrum purlin (doigeta) & \textit{Pinus} sp. \\
\texttt{KYOw02773} & Nishi Hongwanji & Kuro Shoin & Rafter (taruki) & \textit{Pinus} sp. \\
\texttt{KYOw02774} & Myoshinji & Small Hojo & Eaves cantilever (hanegi) & \textit{Pinus} sp. \\
\texttt{KYOw02775} & Kajuji & Shoin & Roof framing timber (koyazai) & \textit{Pinus} sp. \\
\texttt{KYOw02776} & Gokurakuin & Main hall & Bearing block (masu) & \textit{Zelkova serrata} \\
\texttt{KYOw02777} & Nijo Castle & Kara-mon Gate & Door framing member (tobira-gamachi) & \textit{Zelkova serrata} \\
\texttt{KYOw02780} & Kannonji & Main hall & Column (hashira) & \textit{Zelkova serrata} \\
\texttt{KYOw02783} & Konchi-in & Hasso-seki tearoom & Column (hashira) & \textit{Cryptomeria japonica} \\
\texttt{KYOw02784} & Nishi Hongwanji & Kuro Shoin & Column (hashira) & \textit{Cryptomeria japonica} \\
\texttt{KYOw02787} & Kannonji & Main hall & Porch tie beam (k\=ohai-nuki) & \textit{Tsuga sieboldii} \\
\texttt{KYOw02788} & Myoshinji & Small Hojo & Eaves cantilever (hanegi) & \textit{Castanea crenata} \\
\texttt{KYOw02789} & Nijo Castle & Rice storehouse & Column-base splice (hashira-netsugi) & \textit{Castanea crenata} \\
\texttt{KYOw02790} & Komyoji & Main hall & Gable pendant (gegyo) & \textit{Castanea crenata} \\
\texttt{KYOw16662} & Horyuji & Five-story pagoda & Rafter (taruki) & \textit{Chamaecyparis obtusa} \\
\texttt{KYOw16666} & Rengeoin & Main hall & Beam (hari) & \textit{Chamaecyparis obtusa} \\
\texttt{KYOw16700} & Hokkeji & Main hall & Roof framing timber (koyazai) & \textit{Pinus} sp. \\
\texttt{KYOw16702} & Hokkeji & Main hall & Large bearing block (daito) & \textit{Pinus} sp. \\
\texttt{KYOw16711} & Enman-in & Shinden & Floor joist (neda) & \textit{Pinus} sp. \\
\texttt{KYOw16714} & Konchi-in & Hasso-seki tearoom & Longitudinal beam (keta) & \textit{Pinus} sp. \\
\texttt{KYOw16716} & Myohoin & \=Oshoin & Hidden rafter (nodaruki) & \textit{Pinus} sp. \\
\texttt{KYOw16719} & Ninnaji & Five-story pagoda & Rafter (taruki) & \textit{Pinus} sp. \\
\texttt{KYOw16737} & Kyo-o-gokokuji & Lecture hall & Bearing block (masu) & \textit{Zelkova serrata} \\
\texttt{KYOw16744} & Komyoji & Main hall & Bearing block (masu) & \textit{Cryptomeria japonica} \\
\texttt{KYOw16750} & Nishi Hongwanji & Kuro Shoin & Roof tie beam (koyabari) & \textit{Abies} sp. \\
\bottomrule
\end{tabular}
\end{table}

\newsavebox{\appendixABCHeading}
\newsavebox{\appendixABCCapA}
\newsavebox{\appendixABCCapB}
\newlength{\appendixABCImageHeight}
\makeatletter
\newcommand{\appendixABCCaption}[1]{\def\@captype{figure}\caption{#1}}
\makeatother
\clearpage
\sbox{\appendixABCHeading}{\begin{minipage}{\textwidth}
\subsection{Specimen Images}
\end{minipage}}
\sbox{\appendixABCCapA}{\begin{minipage}{\textwidth}
\appendixABCCaption{Pseudo-RGB images of all 49 specimens, arranged by specimen ID from left to right and top to bottom. Each ID appears below its image.}
\label{fig:all-specimens-pseudo-rgb}
\end{minipage}}
\setlength{\appendixABCImageHeight}{\dimexpr\textheight
  -\ht\appendixABCHeading-\dp\appendixABCHeading
  -\ht\appendixABCCapA-\dp\appendixABCCapA-3pt\relax}
\vbox{%
  \hbox{\usebox{\appendixABCHeading}}\nointerlineskip\vskip2pt
  \hbox to\textwidth{\hfil
    \includegraphics[width=\textwidth,height=\appendixABCImageHeight,keepaspectratio]{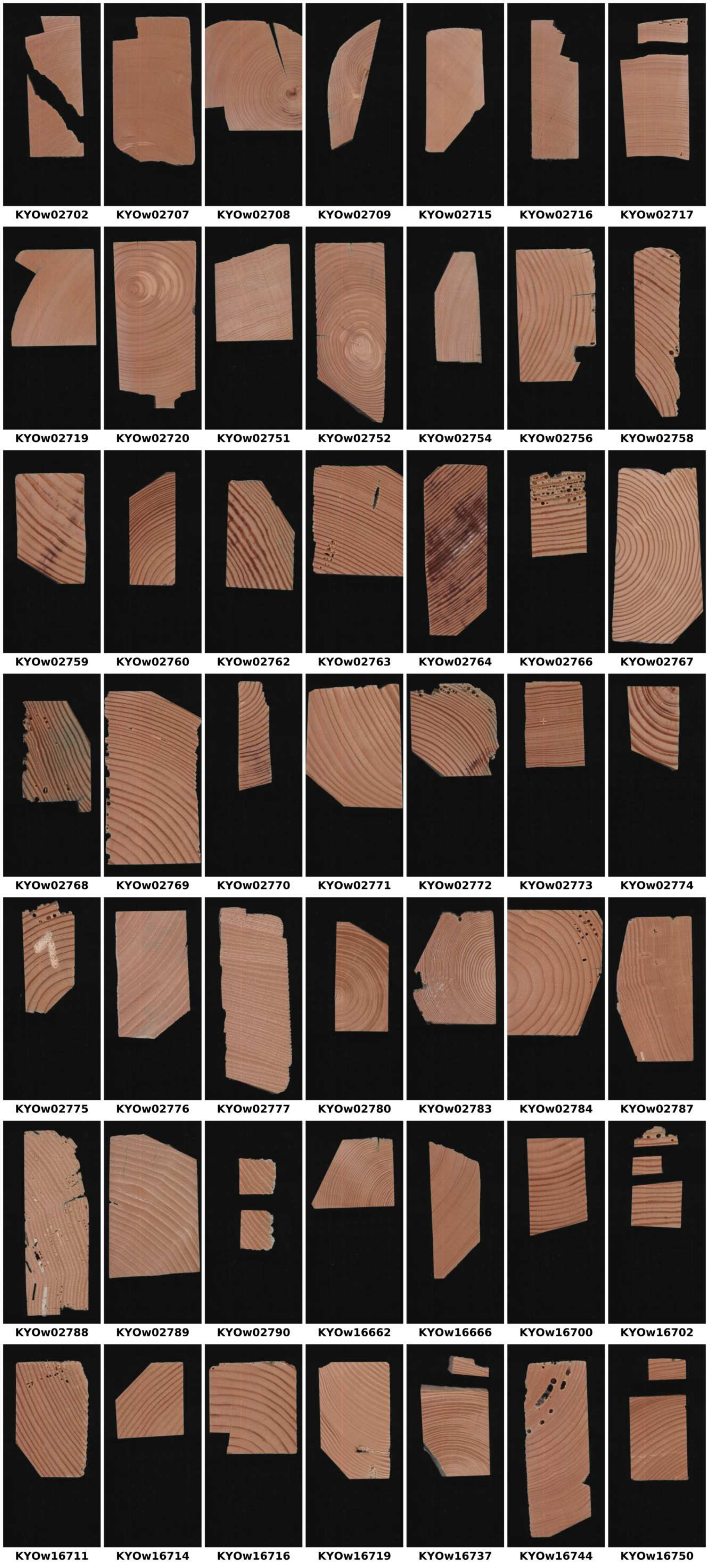}\hfil}%
  \nointerlineskip\hbox{\usebox{\appendixABCCapA}}}

\clearpage
\sbox{\appendixABCHeading}{\begin{minipage}{\textwidth}
\section{Spectral Augmentation after SNV Transformation}
\label{app:snv-augmentation}

This appendix describes how we selected the strengths of TGN and FS and shows how both transformations preserve the mean and norm of each input spectrum.

\subsection{Spectral Deformation and Strength Selection}
\label{app:augmentation-visualization}

We compared fixed TGN rotation angles of $2.5^\circ$, $5^\circ$, and $7.5^\circ$ (Figure~\ref{fig:tgn-examples}) and absolute FS displacements of 1, 2, and 3 channels (Figure~\ref{fig:fs-examples}).
After visually inspecting the resulting deformations, we set the maximum TGN rotation angle to $5^\circ$ and the FS displacement range to $[-2,2]$ channels for training.
\end{minipage}}
\sbox{\appendixABCCapA}{\begin{minipage}{\textwidth}
\appendixABCCaption{Spectral deformation at different TGN rotation angles. The right column shows the augmented spectrum minus the original spectrum.}
\label{fig:tgn-examples}
\end{minipage}}
\sbox{\appendixABCCapB}{\begin{minipage}{\textwidth}
\appendixABCCaption{Spectral deformation at different FS displacements. The right column shows the augmented spectrum minus the original spectrum.}
\label{fig:fs-examples}
\end{minipage}}
\setlength{\appendixABCImageHeight}{\dimexpr(\textheight
  -\ht\appendixABCHeading-\dp\appendixABCHeading
  -\ht\appendixABCCapA-\dp\appendixABCCapA
  -\ht\appendixABCCapB-\dp\appendixABCCapB-7pt)/2\relax}
\vbox{%
  \hbox{\usebox{\appendixABCHeading}}\nointerlineskip\vskip2pt
  \hbox to\textwidth{\hfil\includegraphics[width=\textwidth,height=\appendixABCImageHeight,keepaspectratio]{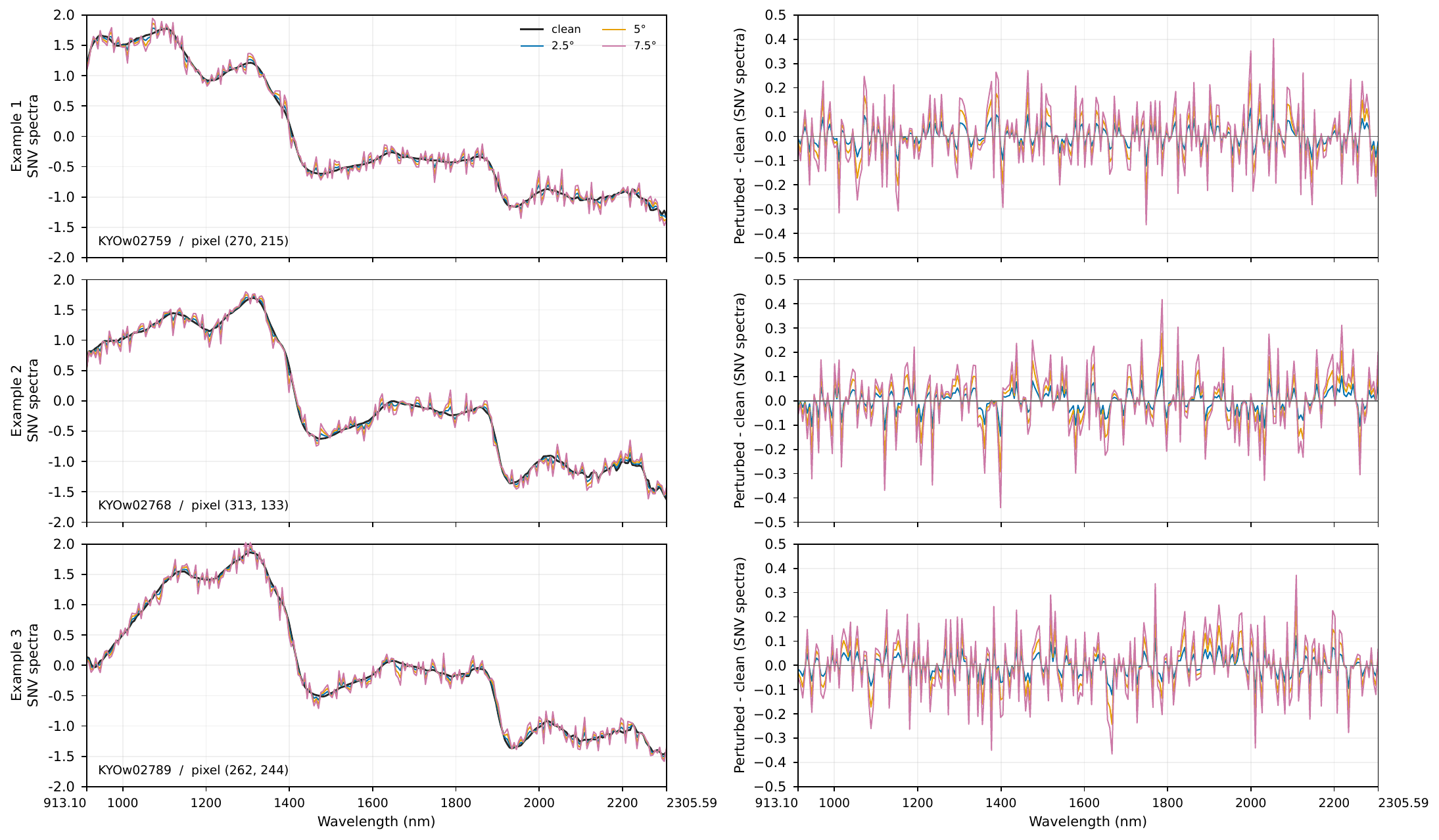}\hfil}%
  \nointerlineskip\hbox{\usebox{\appendixABCCapA}}\nointerlineskip\vskip4pt
  \hbox to\textwidth{\hfil\includegraphics[width=\textwidth,height=\appendixABCImageHeight,keepaspectratio]{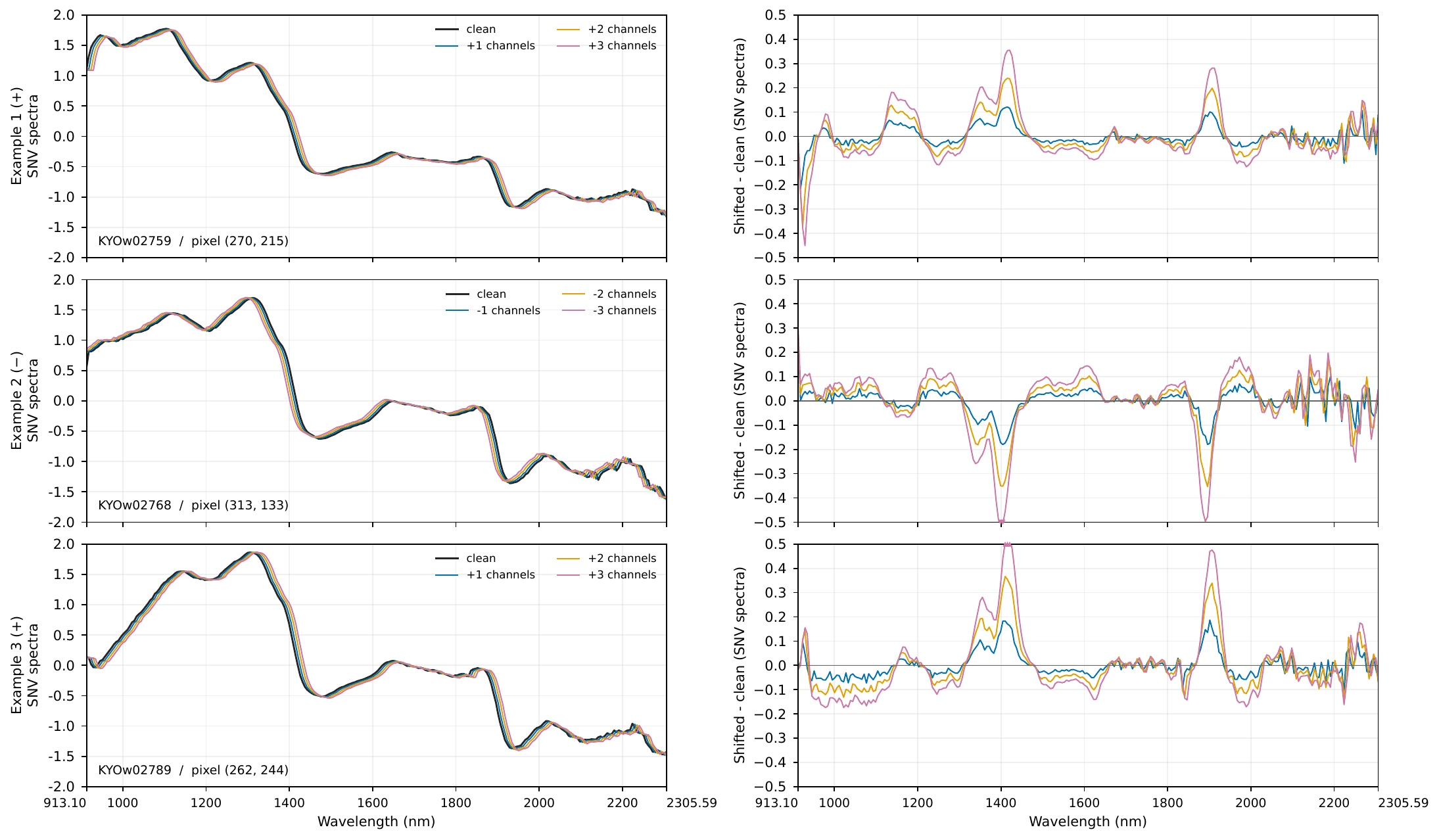}\hfil}%
  \nointerlineskip\hbox{\usebox{\appendixABCCapB}}}

\clearpage
\subsection{Geometry and Constraint Preservation}
\label{app:augmentation-geometry}

For the set in Eq.~\eqref{eq:snv-manifold}, we derive the tangent direction used by TGN and show that TGN and FS preserve the mean and norm in exact arithmetic.
Let $\rho=\sqrt{C-1}$ and $\mathbf{n}=\mathbf{x}/\rho$.
The projection matrix $\mathbf{P}$ in Eq.~\eqref{eq:centering-projection} satisfies $\mathbf{P}^{\mathsf T}=\mathbf{P}$, $\mathbf{P}^2=\mathbf{P}$, and $\mathbf{1}_C^{\mathsf T}\mathbf{P}=\mathbf{0}^{\mathsf T}$.
Since $\mathbf{n}$ has zero mean, $\mathbf{P}\mathbf{n}=\mathbf{n}$.

\par\medskip
\noindent\begin{minipage}[t]{\dimexpr(\textwidth-\columnsep)/2\relax}\vspace{0pt}
\subsubsection{Tangent Gaussian Noise}

We generate a unit tangent vector by removing the mean direction and the radial direction of the input from a standard normal random vector and normalizing the result:
\begin{equation}
  \begin{aligned}
    \boldsymbol{\epsilon}
    &\sim\mathcal{N}(\mathbf{0},\mathbf{I}_C),\\
    \widetilde{\mathbf{q}}
    &=\mathbf{P}\boldsymbol{\epsilon}
      -\mathbf{n}
       \bigl(\mathbf{n}^{\mathsf T}\mathbf{P}\boldsymbol{\epsilon}\bigr),\\
    \mathbf{q}
    &=\frac{\widetilde{\mathbf{q}}}
            {\|\widetilde{\mathbf{q}}\|_2}
  \end{aligned}
  \label{eq:tgn-direction}
\end{equation}
Here, $\mathcal{N}(\mathbf{0},\mathbf{I}_C)$ is the $C$-dimensional standard normal distribution.
The matrix $\mathbf{P}-\mathbf{n}\mathbf{n}^{\mathsf T}$ is the orthogonal projection onto the tangent space of zero-mean vectors orthogonal to $\mathbf{n}$.
If $\|\widetilde{\mathbf{q}}\|_2>0$, then $\mathbf{1}_C^{\mathsf T}\mathbf{q}=0$, $\mathbf{n}^{\mathsf T}\mathbf{q}=0$, and $\|\mathbf{q}\|_2=1$.
For a fixed input, orthogonal projection preserves the component of $\boldsymbol{\epsilon}$ along each tangent basis vector, so the coordinates of $\widetilde{\mathbf{q}}$ in any orthonormal basis of the tangent space are independent standard normal variables.
The projected distribution is therefore rotationally invariant within the tangent space, and its normalized direction $\mathbf{q}$ is uniform with respect to surface area on the unit sphere in that space.

The transformation in Eq.~\eqref{eq:tgn} therefore satisfies
\begin{equation}
  \begin{aligned}
    \mathbf{1}_C^{\mathsf T}\mathcal{T}_{\mathrm{TGN}}(\mathbf{x})&=0,\\
    \|\mathcal{T}_{\mathrm{TGN}}(\mathbf{x})\|_2^2
      &=\rho^2(\cos^2\alpha+\sin^2\alpha)=\rho^2,\\
    \frac{\mathbf{x}^{\mathsf T}\mathcal{T}_{\mathrm{TGN}}(\mathbf{x})}{\rho^2}
      &=\cos\alpha
  \end{aligned}
\end{equation}
The output belongs to the same set, and its angle relative to the input is $\alpha$ for $0\le\alpha\le\pi$.
The term \emph{Gaussian} describes the distribution used to generate the tangent direction. It does not describe the distribution of the rotation angle or of the difference between the transformed and original spectra.

\end{minipage}\hfill
\begin{minipage}[t]{\dimexpr(\textwidth-\columnsep)/2\relax}\vspace{0pt}
\subsubsection{Fractional Shift}

For FS, let $\mathbf{y}=\mathbf{P}S_\delta(\mathbf{x})$. Then $\mathbf{1}_C^{\mathsf T}\mathbf{y}=0$.
If $\|\mathbf{y}\|_2>0$, normalization in Eq.~\eqref{eq:fractional-shift} preserves the zero mean and norm $\rho$.
Since $\mathbf{x}^{\mathsf T}\mathbf{y}/(\rho\|\mathbf{y}\|_2)$ depends on spectral shape, the angle relative to the input can vary even for a fixed displacement $\delta$.
\end{minipage}\par

\twocolumn

\newsavebox{\appendixCIntro}
\newsavebox{\appendixCMiddle}
\newsavebox{\appendixCEnd}
\newsavebox{\appendixCTableA}
\newsavebox{\appendixCTableB}
\makeatletter
\newcommand{\appendixCTableCaption}[1]{\def\@captype{table}\caption{#1}}
\newcommand{\appendixCTableBox}[2]{%
  \sbox{#1}{#2}%
  \ifdim\wd#1>\linewidth
    \sbox{#1}{\resizebox{\linewidth}{!}{\usebox{#1}}}%
  \fi}
\newcommand{\appendixCFitColumn}[5]{%
  \par\begingroup
  \sbox{\appendixCIntro}{\begin{minipage}{\linewidth}#1\end{minipage}}%
  \appendixCTableBox{\appendixCTableA}{#2}%
  \sbox{\appendixCMiddle}{\begin{minipage}{\linewidth}#3\end{minipage}}%
  \appendixCTableBox{\appendixCTableB}{#4}%
  \sbox{\appendixCEnd}{\begin{minipage}{\linewidth}#5\end{minipage}}%
  \@tempdima=\dimexpr\textheight-\ht\appendixCIntro-\dp\appendixCIntro
    -\ht\appendixCMiddle-\dp\appendixCMiddle-\ht\appendixCEnd-\dp\appendixCEnd-11pt\relax
  \@tempdimb=\dimexpr\ht\appendixCTableA+\dp\appendixCTableA+\ht\appendixCTableB+\dp\appendixCTableB\relax
  \def\appendixCScale{1}%
  \ifdim\@tempdimb>\@tempdima\relax
    \Gscale@div\appendixCScale{\@tempdima}{\@tempdimb}%
  \fi
  \vbox{%
    \hbox{\usebox{\appendixCIntro}}\nointerlineskip
    \hbox to\linewidth{\hfil\scalebox{\appendixCScale}{\usebox{\appendixCTableA}}\hfil}\vskip4pt\nointerlineskip
    \hbox{\usebox{\appendixCMiddle}}\nointerlineskip
    \hbox to\linewidth{\hfil\scalebox{\appendixCScale}{\usebox{\appendixCTableB}}\hfil}\vskip4pt\nointerlineskip
    \hbox{\usebox{\appendixCEnd}}}%
  \endgroup}
\makeatother
\section{Data Processing and Experimental Settings}
\label{app:config-details}

This appendix gives further details of preprocessing, model architecture, training, clustering, and evaluation.

\subsection{Wavelength Selection and Quality Control}
\label{app:preprocessing-details}

In bands where the difference between the white and dark references is small, errors in the measurements or references have a greater effect on reflectance because this difference is the denominator in the reflectance calculation.
To select the wavelength range, we computed the ratio of the mean to the standard deviation across columns $v$ of the reference difference $\Delta_{v,c}=w_{v,c}-d_{v,c}$ at band $c$.
We defined this proxy for the signal-to-noise ratio (SNR) as
\begin{equation}
  \mathrm{SNR}_{\mathrm{proxy}}(c)
  =
  \frac{\operatorname{Mean}_{v}(\Delta_{v,c})}
       {\operatorname{Std}_{v}(\Delta_{v,c})+10^{-12}}
  \label{eq:reference_snr_proxy}
\end{equation}
where $\operatorname{Std}_{v}$ is the standard deviation without a degrees-of-freedom correction and $10^{-12}$ guards against division by zero.
Variation across columns can reflect both spatial nonuniformity and measurement noise, so this proxy differs from an SNR based on temporal variation in repeated measurements.

Starting at the long-wavelength end, we excluded consecutive bands for which the proxy was nonfinite or at most 10.
We chose the threshold of 10 empirically by inspecting the reference signal as a function of wavelength, to exclude the low values following the sharp drop near 2300\,nm.
The last retained band was at 2305.59\,nm and the first excluded band was at 2311.85\,nm. Figure~\ref{fig:wavelength-cutoff} marks their midpoint, 2308.72\,nm, as the cutoff.

\par\smallskip\noindent\begin{minipage}{\linewidth}
\centering
\includegraphics[width=\linewidth]{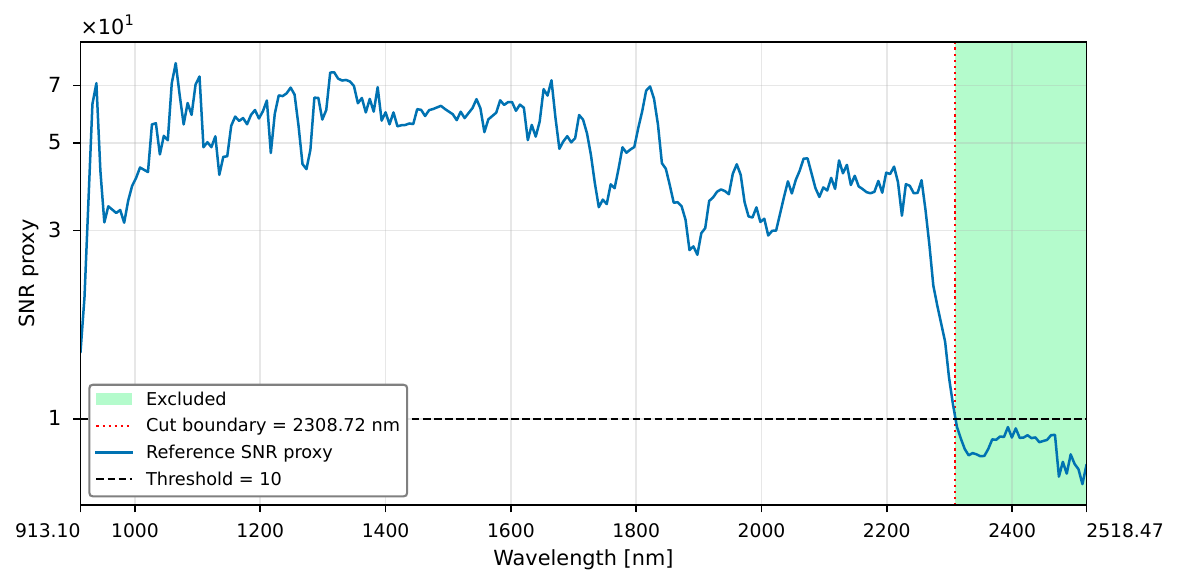}
\appendixABCCaption{SNR proxy based on the difference between the white and dark references, and the retained wavelength range.
The solid blue line shows the proxy in Eq.~\eqref{eq:reference_snr_proxy}, the dashed black line the threshold of 10, and the dotted red line the midpoint between the last retained and first excluded bands (2308.72\,nm).}
\label{fig:wavelength-cutoff}
\end{minipage}\par\smallskip

We resampled the spectra and evaluated pixel quality as described in Section~\ref{sec:preprocessing}.
Reflectance was calculated without adding a small constant to the denominator or clipping the result to $[0,1]$.
We also excluded pixels with any nonfinite reflectance values after interpolation.

\newpage
\appendixCFitColumn{%
\subsection{Model Settings}

Table~\ref{tab:model-settings} lists the model settings shared by A0, A1, M00--M11, and the supplementary mask ratio conditions M11-25 and M11-75.
\appendixCTableCaption{Model settings common to the AE, denoising AE, and ChemoMAE.}
\label{tab:model-settings}%
}{%
\footnotesize
\begin{tabular}{@{}lc@{}}
\toprule
Parameter & Value \\
\midrule
Input channels & 256 \\
Patches & 16 \\
Channels per patch & 16 \\
Encoder embedding dimension & 256 \\
Encoder layers & 8 \\
Attention heads & 8 \\
Feed-forward hidden dimension & 1024 \\
Dropout rate & 0 \\
Latent dimension & 16 \\
\bottomrule
\end{tabular}%
}{%
We initialized the CLS token and positional embeddings from a truncated normal distribution with the standard deviation parameter set to 0.02.
For linear layers, the weight initialization used a uniform distribution whose range depended on the input dimension.
For the query, key, and value weights in attention, we used Xavier uniform initialization.
LayerNorm scales were initialized to one and biases to zero.
All encoder layers were initialized with identical parameter values and then optimized jointly, with separate parameters for each layer.

Table~\ref{tab:condition-settings} lists the settings that differed by condition.
M11-25 and M11-75 are the supplementary conditions for the mask ratio comparison in Appendix~\ref{app:mask-sweep}.
In that comparison, M11(-50) denotes the main M11 condition.
The supplementary comparison under full-dataset fitting used M11(-50) and M11-75 (Appendix~\ref{app:global-mask-rate}).
The loss was the mean squared error over the specified target channels, without further normalization within patches.
\appendixCTableCaption{Number of masked patches out of 16, loss targets, and probabilities of applying the additional perturbations in each condition.}
\label{tab:condition-settings}%
}{%
\footnotesize
\setlength{\tabcolsep}{3pt}
\begin{tabular}{@{}lcccc@{}}
\toprule
Condition & \shortstack{Masked\\patches} & Loss target & TGN & FS \\
\midrule
A0     & 0  & All channels & 0   & 0   \\
A1     & 0  & All channels & 0.5 & 0.5 \\
M00    & 8  & Masked channels & 0   & 0   \\
M11    & 8  & Masked channels & 0.5 & 0.5 \\
\midrule
M10    & 8  & Masked channels & 0.5 & 0   \\
M01    & 8  & Masked channels & 0   & 0.5 \\
\midrule
M11-25 & 4  & Masked channels & 0.5 & 0.5 \\
M11-75 & 12 & Masked channels & 0.5 & 0.5 \\
\bottomrule
\end{tabular}%
}{%
}
\newpage
\appendixCFitColumn{%
\subsection{Training Settings}

We used the settings in Table~\ref{tab:training-settings} for all neural network conditions in both CV and full-dataset fitting.
\appendixCTableCaption{Training settings common to the AE, denoising AE, and ChemoMAE.}
\label{tab:training-settings}%
}{%
\footnotesize
\begin{tabular}{@{}ll@{}}
\toprule
Parameter & Value \\
\midrule
Optimizer & AdamW \\
$(\beta_1,\beta_2)$ & $(0.9,0.95)$ \\
Weight decay & $0.05^{\dagger}$ \\
Batch size & 1024 \\
Training epochs & 800 \\
Maximum learning rate & $6.0\times10^{-4}$ \\
Minimum learning rate & 0 \\
Warmup & 40 epochs \\
Learning rate after warmup & Cosine decay \\
\bottomrule
\end{tabular}%
}{%
\smallskip
\begin{minipage}{.97\linewidth}
\footnotesize
$^{\dagger}$We excluded biases and LayerNorm parameters from weight decay. Weight decay was applied to all other parameters, including the CLS token and learnable positional embeddings.
\end{minipage}

Following the training procedure for MAE~\citep{he2022mae}, we scaled the learning rate linearly with batch size.
With an effective batch size of 1024 and a base learning rate of $1.5\times10^{-4}$ for a batch size of 256, the maximum learning rate was
\begin{equation}
  \eta_{\max}
  =1.5\times10^{-4}\frac{1024}{256}
  =6.0\times10^{-4}
\end{equation}
During warmup, we increased the learning rate linearly from zero to $\eta_{\max}$.

We shuffled the training pixels at each epoch and omitted any final incomplete batch from neural network training for that epoch.
We used FP16 automatic mixed precision with dynamic gradient scaling and kept the model parameters in FP32.

\subsection{Cosine-KMeans Settings}
\label{app:clustering-settings}

We used Cosine-KMeans as described in Section~\ref{sec:mapping}, with $\ell_2$-normalized input representations and FP32 arithmetic.
For $k$-means++-style initialization, we drew the first center uniformly from the representations of the sampled pixels and sampled each subsequent center with probability proportional to its cosine dissimilarity to the nearest existing center.
Each fit used a single initialization.

We allowed up to 500 iterations.
We stopped when the mean cosine dissimilarity between pixels and their assigned centers changed by less than $10^{-7}$ in absolute terms or less than $10^{-4}$ in relative terms.
We calculated the relative change by dividing the absolute change by the absolute value of the previous mean plus $10^{-12}$.
If a cluster was empty, we replaced its center with the representation of a sampled pixel, selecting pixels in descending order of cosine dissimilarity to their nearest center.

For CV, we fixed the initialization seed in advance for each combination of fold, repetition, and $K$ and used the same seed across conditions for that combination.%
}{%
}{%
}
\newpage
For full-dataset fitting, we held the representations for each condition and the common set of fitting pixels fixed.
We ran Cosine-KMeans three times at $K=8$, using the same sequence of initialization seeds across the six main conditions and the supplementary condition M11-75.
All three repetitions contributed to the quantitative summaries. The maps show the first clustering result, as specified in advance.

\subsection{PCA Settings}

We used the \texttt{covariance\_eigh} solver in \texttt{scikit-learn} for PCA, retaining 16 principal components without additional wavelength-wise standardization or whitening.

\subsection{Evaluation Conventions}
\label{app:evaluation-conventions}

We used the common specimen sets defined in Section~\ref{sec:metric-aggregation}, without imputing undefined values.
We required at least one specimen to compute an aggregate and at least two to compute the between-specimen sample standard deviation.
LLA was undefined for fewer than two valid pixels, for an empty set of valid neighboring pairs, or when assignment to a single cluster gave $P_s=1$.

For the cosine-silhouette, we set the contribution of a pixel to zero if its cluster contained only that pixel or if both its mean within-cluster distance and its smallest mean distance to another cluster were zero.
The metric was undefined when all valid pixels from the held-out specimens in a fold belonged to one cluster or when each of those pixels belonged to a different cluster.

\subsection{Cluster Label Alignment}
\label{app:spectral-summary-settings}

To align display IDs and colors across conditions, we used the input spectra of pixels that were valid in all conditions.
For each condition, we calculated the mean 256-dimensional spectrum for each specimen--cluster combination, then averaged these spectra with equal weights across specimens containing pixels assigned to that cluster.
We excluded empty combinations from the average without substituting zeros.
The resulting spectrum retains zero mean across wavelengths, but its norm can vary.

Using M00 as the reference, we applied the Hungarian algorithm to find the one-to-one correspondence that maximized the sum of cosine similarities between matched cluster mean spectra.
Each condition used the same correspondence to M00 for all specimens.
We recorded the similarities of matched spectra and the overlap of their labels on the same valid pixels. Only spectral similarity entered the matching objective.
Matching aligns display IDs and colors. A shared color does not imply chemical identity.
We calculated all metrics from the original partitions independently of the display alignment.

\clearpage
\section{LLA and Spatial Association Measures}
\label{app:lla-spatial-association}

LLA, defined in Eq.~\eqref{eq:lla}, equals one minus the variance-weighted mean of Geary's $C$ for the binary indicator variables of the individual clusters.
To show this, we first express LLA as a join-count agreement rate corrected for chance under random assignment of labels with the number of pixels in each cluster held fixed.
We then express the correction in terms of a ratio of disagreement rates and relate these rates to squared differences between neighboring indicators and to the variances of the indicators.

Throughout this derivation, we fix the comparison condition, number of clusters $K\in\mathcal{K}$, repetition, specimen $s$, and neighborhood width $w$.
Here, $\mathcal{K}$ is the set of candidate cluster counts defined in Section~\ref{sec:comparison-design}, and $k\in\{1,\ldots,K\}$ indexes the clusters.

We use the definitions in Section~\ref{sec:cv-metrics} for the set of valid pixels $\Omega_s$, their number $N_s$, the number $n_{s,k}$ and fraction $\pi_{s,k}$ of pixels assigned to cluster $k$, and the chance agreement probability $P_s$.
The binary image for cluster $k$ has value $\mathbf{B}_{s,k}(p)=\mathbf{1}[y_p=k]$ at pixel $p$, where $\mathbf{1}[\cdot]$ is one when its argument is true and zero otherwise.
We assume that $N_s\ge2$, at least one valid neighboring pair exists, and $P_s<1$.

Using the neighborhood kernel $\mathbf{h}_w$ defined in the main text, we set the weight for pixels $p,q\in\Omega_s$ to $\omega_{pq}^{(w)}=\mathbf{h}_w(p-q)$.
All sums over $p,q$ are restricted to $\Omega_s$, excluding background pixels, excluded pixels, and positions outside the image.
The weights are binary and symmetric, with $\omega_{pp}^{(w)}=0$.
Each valid neighboring pair receives equal weight, without normalization by the number of neighbors of each pixel.
Pixels with no valid neighbors still contribute to $N_s$ and $n_{s,k}$.

\subsection{Join Counts and Chance Correction}

The uncorrected agreement rate corresponds to join-count statistics that count neighboring pairs of the same category~\citep{epperson2003joincounts}.
Let $S_{s,w}$ denote the total number of ordered pairs of valid neighboring pixels and $J_{s,k,w}$ the number of unordered neighboring pairs within cluster $k$.
The uncorrected agreement rate $A_{s,w}$ defined in the main text is then given by
\begin{equation}
  \begin{aligned}
    S_{s,w}
    &=\sum_{p,q}\omega_{pq}^{(w)},\\
    J_{s,k,w}
    &=\frac{1}{2}\sum_{p,q}\omega_{pq}^{(w)}\mathbf{B}_{s,k}(p)\mathbf{B}_{s,k}(q),\\
    A_{s,w}
    &=\frac{2\sum_{k=1}^{K}J_{s,k,w}}{S_{s,w}}
  \end{aligned}
  \label{eq:lla-join-counts}
\end{equation}
The factor $1/2$ in $J_{s,k,w}$ accounts for counting each pair in both directions, $(p,q)$ and $(q,p)$.
Writing $(u,v)$ for the coordinates of $p$, the sums of the elementwise products over the image are
\begin{equation}
  \begin{aligned}
    \sum_{u,v}
    \bigl[\mathbf{B}_{s,k}\odot(\mathbf{h}_w*\mathbf{B}_{s,k})\bigr]_{u,v}
    &=2J_{s,k,w},\\
    \sum_{u,v}
    \bigl[\mathbf{M}_s\odot(\mathbf{h}_w*\mathbf{M}_s)\bigr]_{u,v}
    &=S_{s,w}
  \end{aligned}
  \label{eq:lla-pixel-sums}
\end{equation}
Both $\mathbf{B}_{s,k}$ and $\mathbf{M}_s$ are zero at background and excluded pixels, which therefore contribute zero.
Thus, the convolution, elementwise multiplication, and summation in the main text give the fraction $A_{s,w}$ of all valid neighboring pairs accounted for by same-cluster joins.

The agreement rate can be high even in the absence of spatial clustering if a single cluster occupies a large fraction of the pixels.
We therefore take as a reference a uniformly random assignment of labels with valid pixel locations and the number of pixels in each cluster held fixed.
Any two distinct valid pixels then have the same label with probability $P_s$.
The neighborhood relations are fixed, and this probability is the same for every valid neighboring pair.
With $\mathbb{E}$ denoting expectation over random assignments, we have
\begin{equation}
  \mathbb{E}[A_{s,w}]
  =P_s=\frac{\sum_{k=1}^{K}\binom{n_{s,k}}{2}}{\binom{N_s}{2}}
  \label{eq:lla-random-agreement}
\end{equation}
Thus, $P_s$ is the expected neighboring-pair agreement rate determined solely by the cluster composition of the specimen.

The observed agreement exceeds its expectation by $A_{s,w}-P_s$, whereas perfect agreement ($A_{s,w}=1$) exceeds it by $1-P_s$.
LLA divides the former difference by the latter.
Equivalently, in terms of disagreement rates,
\begin{equation}
  \begin{aligned}
    \mathrm{LLA}_{s,w}
    &=\frac{A_{s,w}-P_s}{1-P_s}\\
    &=1-\frac{1-A_{s,w}}{1-P_s}
  \end{aligned}
  \label{eq:lla-chance-correction}
\end{equation}
Since $P_s$ is constant under random assignment, $\mathbb{E}[\mathrm{LLA}_{s,w}]=0$.
Perfect agreement ($A_{s,w}=1$) gives an LLA of one.
If $A_{s,w}<P_s$, LLA is negative, indicating less agreement between neighbors than expected under random assignment.

The ratio $(1-A_{s,w})/(1-P_s)$ in Eq.~\eqref{eq:lla-chance-correction} expresses the observed neighboring-pair disagreement rate as a fraction of its expectation under random assignment.
LLA can therefore also be interpreted as the relative reduction in disagreement from this reference level.

\newpage
\subsection{Relation to Geary's C}

Geary's $C$ measures spatial autocorrelation by normalizing squared differences between neighbors by the overall variation~\citep{geary1954contiguity}.
We apply it to the binary indicator variable for each cluster and relate the sums of squared neighboring differences and of indicator variances to the disagreement rates above.
The calculation treats cluster labels as nominal categories.

Let $\mathcal{C}_s=\{k\in\{1,\ldots,K\}:n_{s,k}>0\}$ index the clusters represented in specimen $s$.
Its cardinality equals the number of clusters used, $K_s^{\mathrm{used}}$, defined in Section~\ref{sec:cv-metrics}.
Because $P_s<1$, we have $K_s^{\mathrm{used}}\ge2$ and $0<n_{s,k}<N_s$ for every $k\in\mathcal{C}_s$.

For each $k\in\{1,\ldots,K\}$, let $V_{s,k}$ denote the sum of squared deviations of $\mathbf{B}_{s,k}$ from its mean over valid pixels and $Q_{s,k,w}$ the weighted sum of squared differences between neighbors.
For $k\in\mathcal{C}_s$, denote Geary's $C$ by $C_{s,k,w}$.
These quantities are given by
\begin{equation}
  \begin{aligned}
    V_{s,k}
    &=\sum_p\bigl[\mathbf{B}_{s,k}(p)-\pi_{s,k}\bigr]^2\\
    &=N_s\pi_{s,k}(1-\pi_{s,k}),\\
    Q_{s,k,w}
    &=\sum_{p,q}\omega_{pq}^{(w)}\bigl[\mathbf{B}_{s,k}(p)-\mathbf{B}_{s,k}(q)\bigr]^2,\\
    C_{s,k,w}
    &=\frac{N_s-1}{2S_{s,w}}\frac{Q_{s,k,w}}{V_{s,k}}
  \end{aligned}
  \label{eq:lla-geary-indicators}
\end{equation}
Since $0<\pi_{s,k}<1$, we have $V_{s,k}>0$, so $C_{s,k,w}$ is defined.
The valid pixels and neighborhood weights are the same as for LLA.

First, consider the relationship between squared neighboring differences and label disagreement.
Each valid pixel belongs to one cluster in $\mathcal{C}_s$.
If two pixels have the same label, all indicator differences vanish.
If their labels differ, the squared difference is one for each of the two indicators corresponding to their respective clusters.
Hence,
\begin{equation}
  \sum_{k\in\mathcal{C}_s}\bigl[\mathbf{B}_{s,k}(p)-\mathbf{B}_{s,k}(q)\bigr]^2
  =2\cdot\mathbf{1}[y_p\ne y_q]
\end{equation}
Multiplying both sides by $\omega_{pq}^{(w)}$ and summing over valid pixel pairs gives
\begin{equation}
  \sum_{k\in\mathcal{C}_s}Q_{s,k,w}
  =2S_{s,w}(1-A_{s,w})
  \label{eq:lla-neighbor-differences}
\end{equation}

Next, consider the relationship between the sum of indicator variances and the expected disagreement rate under random assignment.
Writing $P_s$ in Eq.~\eqref{eq:lla-random-agreement} in terms of cluster occupancies and using $V_{s,k}$ from Eq.~\eqref{eq:lla-geary-indicators} gives
\begin{equation}
  \begin{aligned}
    1-P_s
    &=\frac{N_s}{N_s-1}\sum_{k=1}^{K}\pi_{s,k}(1-\pi_{s,k}),\\
    \sum_{k\in\mathcal{C}_s}V_{s,k}
    &=N_s\sum_{k=1}^{K}\pi_{s,k}(1-\pi_{s,k})\\
    &=(N_s-1)(1-P_s)
  \end{aligned}
  \label{eq:lla-total-variation}
\end{equation}
Empty clusters have zero occupancy, so summing over $\mathcal{C}_s$ or over $\{1,\ldots,K\}$ gives the same result.

Each $V_{s,k}$ equals $N_s$ times the variance of its binary indicator over valid pixels, computed with divisor $N_s$.
Since $N_s$ is common to all clusters within a specimen, weighting by $V_{s,k}$ gives the variance-weighted mean.
Denoting this mean by $\overline{C}_{s,w}$, we obtain
\begin{equation}
  \begin{aligned}
    \overline{C}_{s,w}
    &=\frac{\sum_{k\in\mathcal{C}_s}V_{s,k}C_{s,k,w}}
            {\sum_{k\in\mathcal{C}_s}V_{s,k}}\\
    &=\frac{N_s-1}{2S_{s,w}}
      \frac{\sum_{k\in\mathcal{C}_s}Q_{s,k,w}}
           {\sum_{k\in\mathcal{C}_s}V_{s,k}}\\
    &=\frac{1-A_{s,w}}{1-P_s}
  \end{aligned}
  \label{eq:lla-weighted-geary}
\end{equation}
The last equality follows by substituting Eq.~\eqref{eq:lla-neighbor-differences} in the numerator and Eq.~\eqref{eq:lla-total-variation} in the denominator.
Comparison with Eq.~\eqref{eq:lla-chance-correction} gives the exact identity
\begin{equation}
  \mathrm{LLA}_{s,w}=1-\overline{C}_{s,w}
  \label{eq:lla-geary-equivalence}
\end{equation}

When $K_s^{\mathrm{used}}=2$, the two indicators are complements and have identical values of $C_{s,k,w}$.
Thus $\mathrm{LLA}_{s,w}=1-C_{s,k,w}$ for either cluster.

A cluster with no pixels in the specimen has $V_{s,k}=0$.
Its $C_{s,k,w}$ is undefined, and the cluster is excluded from the weighted mean.
If $K_s^{\mathrm{used}}=1$, all $V_{s,k}$ are zero and $P_s=1$, so LLA is undefined, as specified in the main text.

\onecolumn
\makeatletter
\global\@colroom\@colht
\global\vsize\@colroom
\makeatother
\begingroup
\newsavebox{\appEProse}
\newsavebox{\appEFigCaption}
\newsavebox{\appEFigImage}
\newsavebox{\appETabOne}
\newsavebox{\appETabTwo}
\newsavebox{\appETabCapOne}
\newsavebox{\appETabCapTwo}
\newsavebox{\appETop}
\newsavebox{\appELeft}
\newlength{\appEFree}
\newlength{\appEColWidth}
\setlength{\appEColWidth}{\dimexpr(\textwidth-\columnsep)/2\relax}
\makeatletter
\long\def\appECaption#1#2{\def\@captype{#1}#2\par}
\makeatother
\newcommand{\appEFitWidth}[1]{%
  \ifdim\wd#1>\linewidth
    \sbox{#1}{\resizebox{\linewidth}{!}{\usebox{#1}}}%
  \fi}
\long\def\appETable#1#2{%
  {\appECaption{table}{#1}}%
  \sbox{\appETabOne}{#2}\appEFitWidth{\appETabOne}%
  \nointerlineskip\hbox to\linewidth{\hfil\usebox{\appETabOne}\hfil}}
\long\def\appEPlate#1#2#3{%
  \setbox\appEProse=\vbox{#1\par}%
  \setbox\appEFigCaption=\vbox{\appECaption{figure}{#3}}%
  \setlength{\appEFree}{\dimexpr\textheight
    -\ht\appEProse-\dp\appEProse-\ht\appEFigCaption-\dp\appEFigCaption-2pt\relax}%
  \noindent\vbox{%
    \unvbox\appEProse\nointerlineskip
    \hbox to\linewidth{\hfil
      \includegraphics[width=\linewidth,height=\appEFree,keepaspectratio]{#2}\hfil}%
    \nointerlineskip\unvbox\appEFigCaption}\par}

\noindent\begin{minipage}{\linewidth}
\setlength{\multicolsep}{0pt}
\begin{multicols}{2}
\raggedcolumns

\section{Supplementary Comparisons and Sensitivity Analysis}
\label{app:supplementary-evaluation}

To supplement the evaluation of M11 and its components in Sections~\ref{sec:results-k8}--\ref{sec:results-global-maps}, we examined variation in condition effects across repetitions, spatial maps from full-dataset fitting, and sensitivity to the mask ratio.
Appendix~\ref{app:reproducibility} examines the reproducibility of the partitions and their internal boundary configurations.

\subsection{Variability in Condition Effects across Repetitions}
\label{app:repeat-variability}

We examined whether the effects of the reconstruction task and additional perturbations had the same direction when we repeated the entire procedure from representation learning through clustering.
Table~\ref{tab:oof-k8-repeat-effects} gives the mean paired differences in LLA-5 and LFR(TGN+FS) for each repetition in two comparisons of reconstruction tasks and four comparisons of additional perturbations.
In each repetition, we calculated the difference between conditions within each specimen and averaged the differences with equal weight across the 49 specimens.
The improvement counts in Table~\ref{tab:reconstruction-effects} are based on each specimen's mean over three repetitions, whereas the differences reported here are means across specimens for each repetition.

Changing the reconstruction task to prediction of masked bands (M00$-$A0 and M11$-$A1) increased mean LLA-5 and decreased LFR(TGN+FS) in all three repetitions.
In all four comparisons of additional perturbations, LFR(TGN+FS) decreased in every repetition, but the direction of the change in LLA-5 varied across repetitions.
The small gain for M11 over M00 in Section~\ref{sec:results-ablation} is an effect averaged over three repetitions.

\end{multicols}

\smallskip

\appETable{\caption{Mean paired differences by repetition in OOF evaluation at $K=8$.
  The first two rows compare reconstruction tasks, and the last four compare additional perturbations.
  Positive differences in LLA-5 and negative differences in LFR(TGN+FS) indicate improvement.}
  \label{tab:oof-k8-repeat-effects}}{%
\small
  \setlength{\tabcolsep}{7pt}
\begin{tabular}{@{}lcccccc@{}}
    \toprule
    Comparison & \multicolumn{3}{c}{LLA-5} & \multicolumn{3}{c}{LFR(TGN+FS)} \\
    \cmidrule(lr){2-4}\cmidrule(l){5-7}
    & Rep. 1 & Rep. 2 & Rep. 3 & Rep. 1 & Rep. 2 & Rep. 3 \\
    \midrule
    M00$-$A0 & $0.0751$ & $0.154$ & $0.155$ & $-0.261$ & $-0.311$ & $-0.294$ \\
    M11$-$A1 & $0.205$ & $0.180$ & $0.151$ & $-0.196$ & $-0.165$ & $-0.128$ \\
    \midrule
    A1$-$A0 & $-0.118$ & $-0.0283$ & $0.0240$ & $-0.245$ & $-0.316$ & $-0.357$ \\
    M10$-$M00 & $0.00534$ & $0.0112$ & $-0.00652$ & $-0.0373$ & $-0.0690$ & $-0.0582$ \\
    M01$-$M00 & $0.0177$ & $-0.000772$ & $-0.0184$ & $-0.0962$ & $-0.0936$ & $-0.113$ \\
    M11$-$M00 & $0.0121$ & $-0.00247$ & $0.0195$ & $-0.179$ & $-0.170$ & $-0.190$ \\
    \bottomrule
  \end{tabular}
}

\end{minipage}\par

\clearpage
\appEPlate{%
\subsection{Spatial Maps after Full-Dataset Fitting}
\label{app:full-dataset-analysis}
\label{app:global-map-comparison}

Figure~\ref{fig:global-condition-maps} compares conditions for the seven specimens selected for visualization, and Figure~\ref{fig:m11-all-maps} shows M11 maps for all 49 specimens.
Figure~\ref{fig:global-16744-maps} in the main text presents KYOw16744 from the former comparison.
Each map shows the first clustering result, selected in advance, whereas the quantitative tables report means over three repetitions.
We aligned display IDs and colors to M00 (Appendix~\ref{app:spectral-summary-settings}) and used the same specimen order and field of view across conditions.
The extent of fine-scale intermixing and the delineation of broad regions varied across conditions.
The maps allow the reader to compare the example in Section~\ref{sec:results-global-maps} with the distributions observed in other specimens.
}{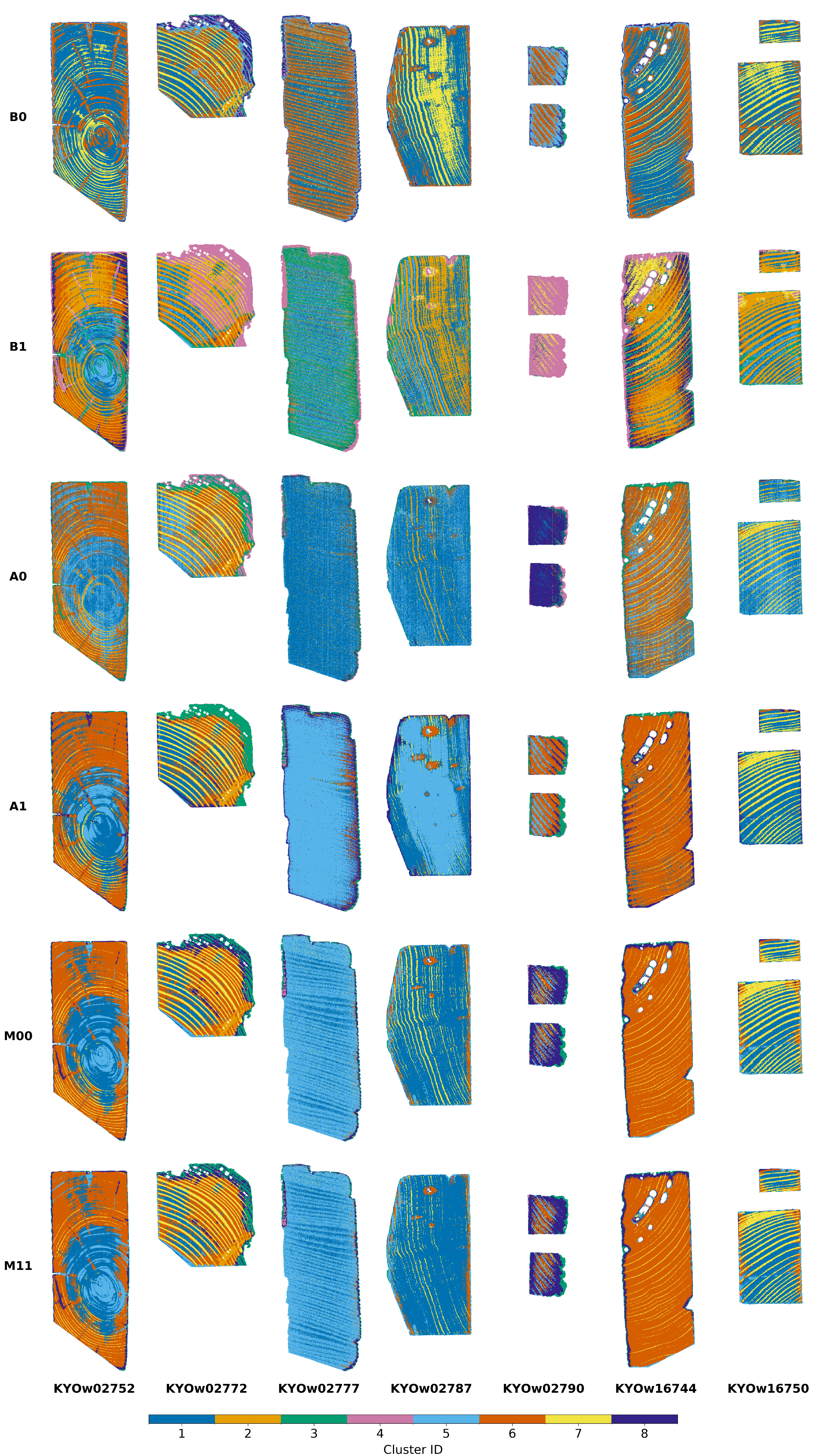}{%
\caption{Spatial maps of the seven selected specimens under the six main conditions after full-dataset fitting ($K=8$).
  White denotes background and excluded pixels.}
  \label{fig:global-condition-maps}
}

\clearpage
\appEPlate{%

}{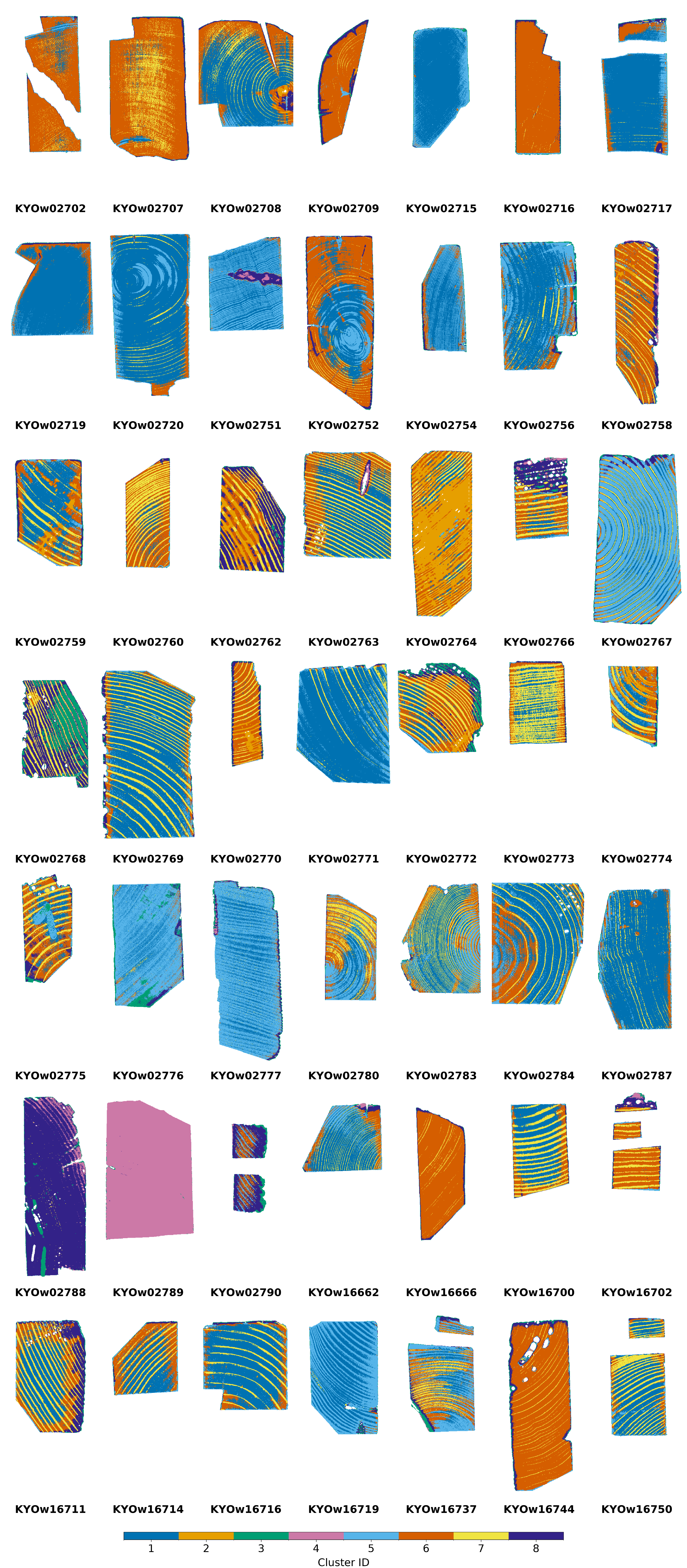}{%
\caption{Spatial maps of all 49 specimens from full-dataset fitting of M11 ($K=8$).
  Specimens are arranged in ascending order of specimen ID from left to right and top to bottom.
  Display IDs and colors match those in Figure~\ref{fig:global-16744-maps}.}
  \label{fig:m11-all-maps}
}

\clearpage
\setbox\appEProse=\vbox{\hsize=\appEColWidth\linewidth=\hsize\columnwidth=\hsize

\subsection{Mask Ratio Sensitivity}
\label{app:mask-sweep}

We examined how the mask ratio affected M11 in both OOF evaluation and full-dataset fitting, which differed in the specimens included in fitting and the steps repeated.
The main mask ratio of 50\% was set independently of these analyses.

\subsubsection{Cross-Validation}

We compared mask ratios of 25\%, 50\%, and 75\%, denoted by M11-25, M11(-50), and M11-75, respectively.
Each model was trained for 800 epochs using the same specimen splits, sampled pixels, seeds for the three repetitions, and settings other than the mask ratio as in the main comparison.
We reused the main results for M11(-50) and aggregated the metrics as in Section~\ref{sec:metric-aggregation}.
Only for M11(-50) at $K=2$ was a specimen excluded from the LLA calculation because all its pixels were assigned to one cluster.

For $K=4,6,8,10,12,14$, mean LLA increased with the mask ratio at all three neighborhood widths (Figure~\ref{fig:mask-oof-k-sweep}).
We examine the changes in the maps accompanying this increase in spatial coherence under full-dataset fitting below.

All three LFR measures at $K=8$ were lowest for M11(-50) (Table~\ref{tab:mask-oof-k8}).
Thus, the higher mean LLA for M11-75 was not accompanied by more stable assignments.
LLA-5 increased from M11(-50) to M11-75 in 32 of 49 specimens, but the mean paired difference was positive in the first two repetitions and negative in the third (Table~\ref{tab:mask-oof-effects}).
In this comparison, spatial coherence and assignment stability under the specified perturbations responded differently to a change in mask ratio.
Table~\ref{tab:mask-oof-ari} reports agreement between partitions across repetitions.

\par}

\setbox\appETabCapOne=\vbox{\appECaption{table}{%
\caption{OOF metrics in the supplementary comparison of mask ratios ($K=8$, 49 specimens).
  Gray values indicate $\pm$ the between-specimen standard deviation.}
  \label{tab:mask-oof-k8}
}}

\sbox{\appETabOne}{%
\small
  \setlength{\tabcolsep}{8pt}
\begin{tabular}{@{}lccc@{}}
    \toprule
    Metric & M11-25 & M11(-50) & M11-75 \\
    \midrule
    LLA-3 $\uparrow$ & $0.595\color{black!55}\pm0.0884$ & $0.609\color{black!55}\pm0.0894$ & $0.632\color{black!55}\pm0.0753$ \\
    LLA-5 $\uparrow$ & $0.504\color{black!55}\pm0.0876$ & $0.519\color{black!55}\pm0.0875$ & $0.545\color{black!55}\pm0.0760$ \\
    LLA-9 $\uparrow$ & $0.397\color{black!55}\pm0.0770$ & $0.412\color{black!55}\pm0.0779$ & $0.440\color{black!55}\pm0.0783$ \\
    LFR(TGN+FS) $\downarrow$ & $0.114\color{black!55}\pm0.0417$ & $0.0891\color{black!55}\pm0.0385$ & $0.0972\color{black!55}\pm0.0409$ \\
    LFR(TGN) $\downarrow$ & $0.104\color{black!55}\pm0.0398$ & $0.0815\color{black!55}\pm0.0357$ & $0.0874\color{black!55}\pm0.0368$ \\
    LFR(FS) $\downarrow$ & $0.0442\color{black!55}\pm0.0199$ & $0.0234\color{black!55}\pm0.0119$ & $0.0271\color{black!55}\pm0.0116$ \\
    Cosine-silhouette & $0.192\color{black!55}\pm0.0877$ & $0.197\color{black!55}\pm0.104$ & $0.197\color{black!55}\pm0.109$ \\
    $K_s^{\mathrm{used}}$ & $7.32\color{black!55}\pm0.930$ & $7.22\color{black!55}\pm0.899$ & $7.29\color{black!55}\pm0.894$ \\
    $\max_k\pi_{s,k}$ & $0.552\color{black!55}\pm0.144$ & $0.562\color{black!55}\pm0.171$ & $0.563\color{black!55}\pm0.157$ \\
    \bottomrule
  \end{tabular}
}

\appEFitWidth{\appETabOne}

\setbox\appETabCapTwo=\vbox{\appECaption{table}{%
\caption{Paired OOF differences within specimens relative to M11(-50) ($K=8$, 49 specimens).
  Gray values indicate $\pm$ the between-specimen standard deviation.
  The improvement count is the number of specimens whose mean over three repetitions changed in the direction of the arrow.
  The last three columns give the mean paired differences for individual repetitions.}
  \label{tab:mask-oof-effects}
}}

\sbox{\appETabTwo}{%
\small
  \setlength{\tabcolsep}{4pt}
\begin{tabular}{@{}llccrrr@{}}
    \toprule
    Comparison & Metric & Mean diff.$\pm$SD & Improved & Rep. 1 & Rep. 2 & Rep. 3 \\
    \midrule
    M11-25$-$M11(-50) & LLA-5 $\uparrow$ & $-0.0149\color{black!55}\pm0.0512$ & 18/49 & $-0.000356$ & $0.0118$ & $-0.0563$ \\
    M11-25$-$M11(-50) & LFR(TGN+FS) $\downarrow$ & $0.0250\color{black!55}\pm0.0261$ & 9/49 & $0.0140$ & $0.0442$ & $0.0169$ \\
    \midrule
    M11-75$-$M11(-50) & LLA-5 $\uparrow$ & $0.0260\color{black!55}\pm0.0473$ & 32/49 & $0.0336$ & $0.0462$ & $-0.00176$ \\
    M11-75$-$M11(-50) & LFR(TGN+FS) $\downarrow$ & $0.00819\color{black!55}\pm0.0239$ & 15/49 & $0.00442$ & $0.0105$ & $0.00970$ \\
    \bottomrule
  \end{tabular}
}

\appEFitWidth{\appETabTwo}

\sbox{\appEFigImage}{\includegraphics[width=\appEColWidth]{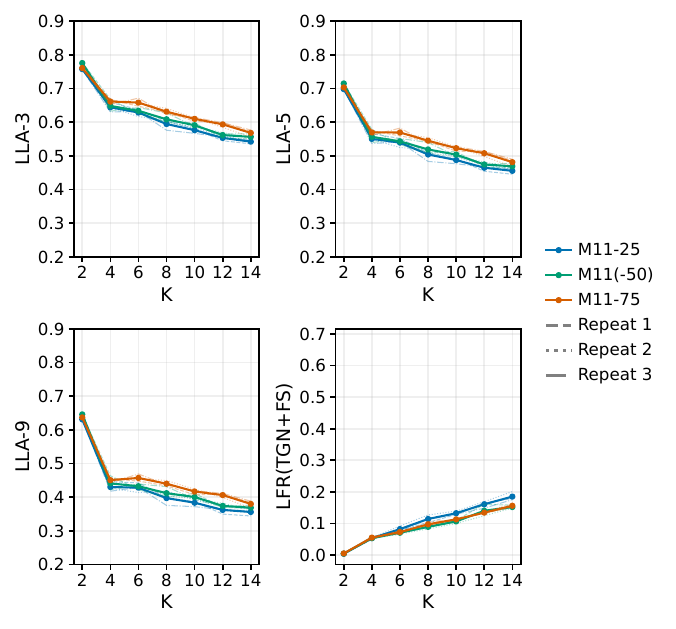}}
\setbox\appEFigCaption=\vbox{%
  \hsize=\appEColWidth\linewidth=\hsize\appECaption{figure}{%

\caption{OOF metrics as a function of $K$ at the three mask ratios.
  Thick lines connecting points indicate means over three repetitions. Thin line styles distinguish individual repetitions, each of which included representation learning.
  All results include 49 specimens, except LLA for M11(-50) at $K=2$, which includes 48.}
  \label{fig:mask-oof-k-sweep}

}}
\setlength{\appEFree}{\dimexpr\textheight
  -(\ht\appEProse+\dp\appEProse+\ht\appEFigImage+\dp\appEFigImage
    +\ht\appEFigCaption+\dp\appEFigCaption)/2
  -\ht\appETabCapOne-\dp\appETabCapOne-\ht\appETabCapTwo-\dp\appETabCapTwo
  -2\baselineskip-10pt\relax}
\ExplSyntaxOn
\edef\appETableScale{\fp_eval:n{min(1,
  \dim_to_fp:n{\appEFree} /
  \dim_to_fp:n{\ht\appETabOne+\dp\appETabOne+\ht\appETabTwo+\dp\appETabTwo})}}
\ExplSyntaxOff
\setbox\appETop=\vbox{%
  \unvbox\appETabCapOne\nointerlineskip
  \hbox to\linewidth{\hfil\scalebox{\appETableScale}{\usebox{\appETabOne}}\hfil}%
  \nointerlineskip\unvbox\appETabCapTwo\nointerlineskip
  \hbox to\linewidth{\hfil\scalebox{\appETableScale}{\usebox{\appETabTwo}}\hfil}%
  \vskip6pt}
\setlength{\appEFree}{\dimexpr\textheight-\ht\appETop-\dp\appETop-2pt\relax}
\splittopskip=\topskip \splitmaxdepth=0pt
\setbox\appELeft=\vsplit\appEProse to\appEFree
\setbox\appELeft=\vbox{\unvbox\appELeft}%
\setbox\appEProse=\vbox{\unvbox\appEProse}%
\setlength{\appEFree}{\dimexpr\appEFree-\ht\appEProse-\dp\appEProse
  -\ht\appEFigCaption-\dp\appEFigCaption-6pt\relax}
\noindent\vbox{%
  \unvbox\appETop\nointerlineskip
  \hbox to\linewidth{%
    \vtop{\hsize=\appEColWidth\hrule width\hsize height0pt\unvbox\appELeft}\hfil
    \vtop{\hsize=\appEColWidth\hrule width\hsize height0pt
      \hbox to\hsize{\hfil
        \ifdim\dimexpr\ht\appEFigImage+\dp\appEFigImage\relax>\appEFree
          \resizebox*{!}{\appEFree}{\usebox{\appEFigImage}}%
        \else\usebox{\appEFigImage}\fi\hfil}%
      \nointerlineskip\unvbox\appEFigCaption\vskip6pt\unvbox\appEProse}}}\par

\clearpage
\appEPlate{%
\subsubsection{Full-Dataset Fitting}
\label{app:global-mask-rate}

Table~\ref{tab:global-mask-rate} compares M11(-50) and M11-75 after full-dataset fitting.
We trained M11-75 once on the same $401{,}408$ pixels with the same training settings and prespecified training seed as in Section~\ref{sec:full-fit-protocol}.
For M11(-50), we reused the learned representations and evaluation results.
With each representation held fixed, we fitted Cosine-KMeans at $K=8$ using the same three seeds.
For this supplementary analysis, we calculated the six metrics in Table~\ref{tab:global-mask-rate} on the full set of valid pixels and used the aggregation and display ID alignment conventions of Section~\ref{sec:full-fit-protocol}.
For cosine-silhouette, we calculated distances across all 49 specimens and averaged the pixelwise scores within each specimen.
Figure~\ref{fig:global-mask-rate-maps} shows the first clustering result for the seven specimens used in the main comparison.

M11-75 had higher mean LLA at all three neighborhood widths and higher mean maximum occupancy than M11(-50), with a similar mean number of clusters used.
The paired difference in LLA-5 averaged 0.0227 and was positive for 33 of 49 specimens.
For KYOw02790, M11-75 showed fewer of the banded regions seen with M11(-50), assigning larger areas to the same cluster.
In this specimen, higher LLA coincided with the merging of regions. Excessive merging could reduce sensitivity to differences in material state.
For this specimen, mean LLA-5 over three repetitions rose from 0.505 to 0.687 under full-dataset fitting but fell from 0.491 to 0.474 in OOF evaluation.
Thus, changing the mask ratio had different effects when the specimen was included in fitting and when it was evaluated as an unseen specimen.

\appETable{\caption{Mask ratio comparison after full-dataset fitting ($K=8$, 49 specimens).
  Gray values indicate $\pm$ the between-specimen standard deviation.
  The last row gives paired differences within specimens, calculated as M11-75 minus M11(-50).}
  \label{tab:global-mask-rate}}{%
\small
  \setlength{\tabcolsep}{4pt}
\begin{tabular}{@{}lcccccc@{}}
    \toprule
    Condition & LLA-3 $\uparrow$ & LLA-5 $\uparrow$ & LLA-9 $\uparrow$ & \shortstack{Cosine-\\silhouette} & $K_s^{\mathrm{used}}$ & $\max_k\pi_{s,k}$ \\
    \midrule
    M11(-50) & $0.616\color{black!55}\pm0.102$ & $0.526\color{black!55}\pm0.107$ & $0.422\color{black!55}\pm0.102$ & $0.217\color{black!55}\pm0.109$ & $7.12\color{black!55}\pm0.894$ & $0.594\color{black!55}\pm0.181$ \\
    M11-75 & $0.637\color{black!55}\pm0.119$ & $0.549\color{black!55}\pm0.122$ & $0.439\color{black!55}\pm0.116$ & $0.230\color{black!55}\pm0.117$ & $7.09\color{black!55}\pm0.945$ & $0.624\color{black!55}\pm0.191$ \\
    \midrule
    Difference & $0.0213\color{black!55}\pm0.0479$ & $0.0227\color{black!55}\pm0.0541$ & $0.0170\color{black!55}\pm0.0539$ & $0.0133\color{black!55}\pm0.0445$ & $-0.0272\color{black!55}\pm0.471$ & $0.0300\color{black!55}\pm0.0770$ \\
    \bottomrule
  \end{tabular}
}
}{Figure_E.4.pdf}{%
\caption{Spatial maps of the seven selected specimens from full-dataset fitting of M11(-50) and M11-75 ($K=8$).
  Each map shows the first clustering result. White denotes background and excluded pixels.}
  \label{fig:global-mask-rate-maps}
}

\endgroup

\newsavebox{\appFGBodyBox}
\newsavebox{\appFGTableBox}
\newlength{\appFGTableSpace}
\makeatletter
\newcommand{\appFGPage}[2]{%
  \par\begingroup
  \setlength{\multicolsep}{0pt}%
  \sbox{\appFGBodyBox}{\begin{minipage}{\linewidth}
    \begin{multicols}{2}
      #1
    \end{multicols}
  \end{minipage}}%
  \sbox{\appFGTableBox}{\begin{minipage}{\linewidth}%
    \def\@captype{table}%
    #2%
  \end{minipage}}%
  \setlength{\appFGTableSpace}{\dimexpr
    \textheight-\ht\appFGBodyBox-\dp\appFGBodyBox-10pt\relax}%
  \noindent\begin{minipage}{\linewidth}
    \centering
    \ifdim\dimexpr\ht\appFGTableBox+\dp\appFGTableBox\relax>\appFGTableSpace
      \resizebox*{!}{\appFGTableSpace}{\usebox{\appFGTableBox}}%
    \else
      \usebox{\appFGTableBox}%
    \fi
    \par\vspace{6pt}\nointerlineskip
    \usebox{\appFGBodyBox}%
  \end{minipage}\par
  \endgroup}
\makeatother
\twocolumn
\begingroup
\setlength{\textfloatsep}{8pt plus 1pt minus 1pt}
\brokenpenalty=10000
\section{Partition and Spatial Map Reproducibility}
\label{app:reproducibility}

\begin{table}[t]
  \centering
  \caption{OOF partition agreement for the main comparison, ablations of the additional perturbations, and the supplementary mask ratio comparison.}
  \label{tab:oof-k8-ari}\label{tab:mask-oof-ari}
  \setlength{\tabcolsep}{8pt}
\begin{tabular}{@{}lc@{}}
    \toprule
    Condition & ARI $\uparrow$ \\
    \midrule
    B0 & $\mathbf{0.746}$\textcolor{black!55}{$\pm0.183$} \\
    B1 & $0.710$\textcolor{black!55}{$\pm0.197$} \\
    A0 & $0.308$\textcolor{black!55}{$\pm0.0933$} \\
    A1 & $0.297$\textcolor{black!55}{$\pm0.106$} \\
    M00 & $0.440$\textcolor{black!55}{$\pm0.117$} \\
    M11(-50) & $0.504$\textcolor{black!55}{$\pm0.177$} \\
    \midrule
    M10 & $0.508$\textcolor{black!55}{$\pm0.166$} \\
    M01 & $0.531$\textcolor{black!55}{$\pm0.153$} \\
    \midrule
    M11-25 & $0.398$\textcolor{black!55}{$\pm0.137$} \\
    M11-75 & $0.456$\textcolor{black!55}{$\pm0.128$} \\
    \bottomrule
  \end{tabular}
\end{table}
\suppressfloats[t]

We interpret regions with particular NIR spectral features in light of their spatial distributions.
To assess whether mapped regions and their internal boundaries recurred across repetitions, we compared how pixels were grouped and where boundaries fell in each specimen.
Across conditions, we related partition agreement to LLA. For M11, we also compared the first map used for interpretation with those from the other repetitions.

\subsection{Partition Reproducibility across Conditions}
\label{app:partition-reproducibility}

We measured partition agreement with the adjusted Rand index (ARI)~\citep{hubert1985comparing}. ARI is invariant to permutations of cluster labels. One indicates perfect agreement, zero is the chance baseline, and negative values are possible.
With the fitting pixels fixed, we compared assignments of all unperturbed valid pixels and averaged ARI over the three repetition pairs per specimen.
Because the pairs were dependent, we did not compute a between-repetition standard deviation.
ARI was undefined for fewer than two valid pixels and set to one if both partitions had a single cluster or both assigned each pixel to its own cluster.
We did not impute missing values. At $K=8$, every pair had a defined ARI in all 49 specimens.
Tables~\ref{tab:oof-k8-ari} and~\ref{tab:global-k8-ari} report the equally weighted mean of the specimen averages.
Gray values after $\pm$ give the between-specimen sample standard deviation. Boldface marks the highest mean in each table.
The interpretation of stability depends on which steps are repeated~\citep{vonluxburg2010stability}.

\begin{table}[t]
  \centering
  \caption{Agreement across Cosine-KMeans initializations with representations held fixed after full-dataset fitting.}
  \label{tab:global-k8-ari}
  \setlength{\tabcolsep}{8pt}
\begin{tabular}{@{}lc@{}}
    \toprule
    Condition & ARI $\uparrow$ \\
    \midrule
    B0 & $0.798$\textcolor{black!55}{$\pm0.0440$} \\
    B1 & $\mathbf{0.857}$\textcolor{black!55}{$\pm0.0635$} \\
    A0 & $0.711$\textcolor{black!55}{$\pm0.163$} \\
    A1 & $0.545$\textcolor{black!55}{$\pm0.148$} \\
    M00 & $0.769$\textcolor{black!55}{$\pm0.121$} \\
    M11 & $0.744$\textcolor{black!55}{$\pm0.114$} \\
    \bottomrule
  \end{tabular}
\end{table}

\paragraph{Cross-Validation}
The three repetitions used the same folds and sampled pixels (Section~\ref{sec:cv-design}).
For neural networks, we repeated representation learning and clustering.
Only clustering varied for B0 (fixed input spectra) and B1 (fixed fold-specific PCA transformations).
In OOF evaluation, all pixels of each held-out specimen were excluded from fitting.

Mean ARI for M11 exceeded those for A0, A1, and M00 but fell below those for B0, B1, M10, and M01 (Table~\ref{tab:oof-k8-ari}).
M11(-50), the main comparison condition with 50\% masking, had the highest mean ARI across mask ratios. M11-75 had lower ARI despite higher LLA (Appendix~\ref{app:mask-sweep}).

\paragraph{Full-Dataset Fitting}
Across the six main conditions (Section~\ref{sec:full-fit-protocol}), we fixed the representations and repeated Cosine-KMeans with three initializations.
We compared assignments of all valid pixels in all 49 specimens, including pixels outside the fitting set.
Cross-validation used different specimens for fitting and repeated different steps, so ARI differences cannot isolate variability from representation learning.
A1 had the highest mean LLA-5 and LLA-9 but the lowest mean ARI. M11 had lower ARI than M00 (Tables~\ref{tab:global-k8} and~\ref{tab:global-k8-ari}).
Thus, spatial coherence within a segmentation and partition agreement across repetitions are distinct.
Since ARI disregards spatial coordinates, we also compared internal boundary positions~\citep{arbelaez2011contour}.

\subsection{Spatial Map Reproducibility in M11}
\label{app:spatial-map-reproducibility}

For M11 at $K=8$, we fixed the fitting set at 8,192 pixels from each of the 49 specimens (401,408 pixels in total).
We trained M11 three times for 800 epochs per run, varying the random draws for initialization, pixel order, masking, and additional perturbations.
We ran Cosine-KMeans on each final representation with the same three initialization seeds, yielding nine maps per specimen.
We compared assignments of all 3,902,250 valid pixels, including pixels not used for fitting, without postprocessing or image registration.
These repetitions used the same observations and fitting pixels, so they exclude variability from reacquiring images or resampling pixels.

We indexed each map by its representation learning and clustering repetitions $(e,k)$.
Of the 36 map pairs, nine shared a representation and 27 used different representations.
We also examined the subset of eight pairs comparing the first map $(1,1)$, used for interpretation, with each of the other maps.
Comparisons across representations include variation from both learning and clustering.

\onecolumn
\endgroup
\appFGPage{%
We compared partitions using ARI and variation of information (VI)~\citep{meila2007comparing}.
For two partitions $A,B$, let $n_{ij}=|A_i\cap B_j|$ be the number of pixels shared by clusters $A_i$ and $B_j$, and let $N$ be the total number of pixels.
With marginal counts $n_i=\sum_j n_{ij}$ and $n_j=\sum_i n_{ij}$, we defined
\begin{equation}
  \begin{aligned}
    H(B\mid A)&=\sum_{i,j:n_{ij}>0}\frac{n_{ij}}{N}\log_2\frac{n_i}{n_{ij}},\\
    H(A\mid B)&=\sum_{i,j:n_{ij}>0}\frac{n_{ij}}{N}\log_2\frac{n_j}{n_{ij}},\\
    \mathrm{VI}(A,B)&=H(B\mid A)+H(A\mid B)
  \end{aligned}
  \label{eq:map-vi}
\end{equation}
VI is measured in bits and equals zero for identical partitions. The two conditional entropies measure how clusters in $A$ split across clusters in $B$ or merge in $B$, respectively. They do not count splits or mergers of spatially connected regions.
We assigned $A,B$ in lexicographic order of $(e,k)$, so $A=(1,1)$ in each first-map comparison.

We represented internal boundaries by the midpoints of edges shared by horizontally or vertically adjacent valid pixels with different labels.
We excluded edges on the specimen outline or adjacent to background, excluded pixels, or holes.
Let $\mathcal{B}_A,\mathcal{B}_B$ be the sets of boundary midpoints, and define the distance between midpoints $p=(r_p,c_p)$ and $q=(r_q,c_q)$ as $d_\infty(p,q)=\max\{|r_p-r_q|,|c_p-c_q|\}$.
For neighborhood widths $w=3,5,9$, we used a tolerance $\delta_w=(w-1)/2$ and defined the directed boundary agreement rate as
\begin{equation}
  C_{A\to B}(w)=\frac{1}{|\mathcal{B}_A|}\sum_{p\in\mathcal{B}_A}
  \mathbf{1}\!\left[\min_{q\in\mathcal{B}_B}d_\infty(p,q)\le\delta_w\right]
  \label{eq:boundary-agreement}
\end{equation}
For each pair, we recorded the lower and higher directed rates. Four-neighbor edges were compared within square neighborhoods at tolerances of 1, 2, and 4 pixels.
Many-to-one matches are allowed, irrespective of orientation or flanking cluster labels. This proximity measure differs from boundary-detection precision and recall.
The rate was undefined when the source set was empty and was zero when the source set was nonempty but the target set was empty.
We included the same specimens at all scales, requiring both directions to be defined for every planned pair at all three tolerances. We did not impute missing values.
Each metric was averaged over pairs per specimen, then summarized by the equally weighted mean and sample standard deviation across specimens. All 49 contributed to every metric and comparison group.

Internal boundary positions were similar across repetitions, including comparisons with the first map (Table~\ref{tab:spatial-map-reproducibility}).
Comparisons across representations had lower ARI and higher VI than those within a representation. Even at a one-pixel tolerance, their lower directed agreement rate averaged 0.879.
Agreement with the first map varied by specimen. The lowest specimen means over the eight pairs were 0.689, 0.743, and 0.810 at 1, 2, and 4 pixels, respectively.
These rates lack chance correction, and denser boundaries favor nearby matches. We therefore consider both directions and their variation across specimens.

Qualitative comparison of the maps across repetitions consistently showed characteristic regions interpreted as degradation-related and distributions following latewood in softwoods.
Assignments varied more visibly in interior regions considered less affected by surface degradation or aging, except where features such as latewood were distinct.
In regions with small spectral differences, quantization of the latent representations may have produced different partitions across repetitions, changing which pixels were grouped together and thereby contributing to the lower ARI.

These observations and the similarity of boundary configurations support the stability of the maps as a basis for interpreting local features in relation to surrounding distributions.
We examine the chemical meaning of each region using the material in Appendix~\ref{app:chemical-interpretation}, relating NIR spectral features and spatial distributions to literature band assignments and specimen information (Section~\ref{sec:results-spectra}).
}{%
\centering
  \caption{Agreement in M11 partitions and internal boundary configurations across repetitions ($K=8$, 49 specimens).
  Values are means $\pm$ between-specimen standard deviations.
  The lower and higher boundary agreement rates are the minimum and maximum of the two directed rates for each pair, not confidence limits.
  The eight pairs involving the first map are a subset of the other two groups combined.}
  \label{tab:spatial-map-reproducibility}
  \small
  \setlength{\tabcolsep}{5pt}
\begin{tabular}{@{}lccc@{}}
    \toprule
    Metric & \shortstack{Within representation\\(9 pairs)} & \shortstack{Across representations\\(27 pairs)} & \shortstack{First map\\(8 pairs)} \\
    \midrule
    ARI & $0.772\color{black!55}\pm0.119$ & $0.714\color{black!55}\pm0.096$ & $0.766\color{black!55}\pm0.076$ \\
    $H(B\mid A)$ [bits] & $0.379\color{black!55}\pm0.156$ & $0.480\color{black!55}\pm0.168$ & $0.409\color{black!55}\pm0.142$ \\
    $H(A\mid B)$ [bits] & $0.381\color{black!55}\pm0.152$ & $0.485\color{black!55}\pm0.168$ & $0.414\color{black!55}\pm0.154$ \\
    VI [bits] & $0.760\color{black!55}\pm0.307$ & $0.966\color{black!55}\pm0.332$ & $0.823\color{black!55}\pm0.286$ \\
    \midrule
    Boundary agreement, $w=3$, lower & $0.899\color{black!55}\pm0.102$ & $0.879\color{black!55}\pm0.085$ & $0.906\color{black!55}\pm0.067$ \\
    Boundary agreement, $w=3$, higher & $0.951\color{black!55}\pm0.043$ & $0.937\color{black!55}\pm0.039$ & $0.951\color{black!55}\pm0.030$ \\
    Boundary agreement, $w=5$, lower & $0.939\color{black!55}\pm0.089$ & $0.933\color{black!55}\pm0.074$ & $0.950\color{black!55}\pm0.054$ \\
    Boundary agreement, $w=5$, higher & $0.979\color{black!55}\pm0.031$ & $0.974\color{black!55}\pm0.027$ & $0.981\color{black!55}\pm0.020$ \\
    Boundary agreement, $w=9$, lower & $0.968\color{black!55}\pm0.069$ & $0.968\color{black!55}\pm0.056$ & $0.978\color{black!55}\pm0.039$ \\
    Boundary agreement, $w=9$, higher & $0.993\color{black!55}\pm0.017$ & $0.993\color{black!55}\pm0.015$ & $0.995\color{black!55}\pm0.011$ \\
    \bottomrule
  \end{tabular}
}
\onecolumn
\makeatletter
\global\@colroom\@colht
\global\vsize\@colroom
\makeatother
\begingroup
\newsavebox{\appHIntro}\newsavebox{\appHTable}\newsavebox{\appHNotes}
\newlength{\appHFree}\newlength{\appHColumns}
\setlength{\tabcolsep}{2pt}
\renewcommand{\arraystretch}{1.00}
\newcount\appHTenths
\newif\ifappHTooLarge
\makeatletter
\newcommand{\appHCaption}[1]{\def\@captype{table}\caption{#1}}
\makeatother
\newcommand{\appHFitPage}[3]{%
  \sbox{\appHIntro}{\begin{minipage}{\linewidth}#1\end{minipage}}%
  \sbox{\appHNotes}{\begin{minipage}{\linewidth}#3\end{minipage}}%
  \setlength{\appHFree}{\dimexpr\textheight-\ht\appHIntro-\dp\appHIntro
    -\ht\appHNotes-\dp\appHNotes-8pt\relax}%
  \appHTenths=100
  \loop
    \sbox{\appHTable}{\fontsize{\dimexpr1pt*\appHTenths/10\relax}%
      {\dimexpr1.1pt*\appHTenths/10\relax}\selectfont
      \settowidth{\appHColumns}{\hbox{No.}\hbox{Wavenumber}}%
      \setlength{\appHColumns}{\dimexpr\linewidth-\appHColumns-8\tabcolsep-.1pt\relax}%
      #2}%
    \appHTooLargefalse
    \ifdim\wd\appHTable>\linewidth \appHTooLargetrue\fi
    \ifdim\dimexpr\ht\appHTable+\dp\appHTable\relax>\appHFree \appHTooLargetrue\fi
    \ifnum\appHTenths=80 \appHTooLargefalse\fi
    \ifappHTooLarge \advance\appHTenths by -1
  \repeat
  \ifdim\wd\appHTable>\linewidth
    \sbox{\appHTable}{\resizebox{\linewidth}{!}{\usebox{\appHTable}}}%
  \fi
  \ifdim\dimexpr\ht\appHTable+\dp\appHTable\relax>\appHFree
    \sbox{\appHTable}{\resizebox*{!}{\appHFree}{\usebox{\appHTable}}}%
  \fi
  \noindent\vbox{%
    \hbox{\usebox{\appHIntro}}\nointerlineskip\vskip2pt
    \hbox to\linewidth{\hfil\usebox{\appHTable}\hfil}\nointerlineskip\vskip4pt
    \hbox{\usebox{\appHNotes}}}\par}
\appHFitPage{%
\section{NIR Band Assignments and Specimen Maps}
\label{app:chemical-interpretation}

\subsection{NIR Band Assignments}
\label{app:nir-band-assignments}

Table~\ref{tab:nir-band-assignments} gives the assignments corresponding to the numbered bands in Figures~\ref{fig:m11-nir-spectra} and~\ref{fig:chemical-atlas}.
We used Table~2 and the accompanying text in the review by \citet{schwanninger2011assignments} as a starting point and checked the assignments against original studies comparing wood with isolated lignin and cellulose~\citep{fackler2010polysaccharide}, assigning bands in cellulosic paper~\citep{ali2001cellulosic}, and examining water--polymer interactions~\citep{moll2022water}.
The assignment of band~7 to semicrystalline cellulose~\citep{yonenobu2003antique} and the description of the combination band at band~16~\citep{tsuchikawa2005ageing} follow the respective original studies.
At band~12, an assignment to CH$_3$ in the acetyl groups of xylan has also been reported~\citep{fackler2007decay}, leaving the specific vibrational assignment open.
At band~10, aromatic extractives may contribute alongside lignin~\citep{wagner2020larch}.
The table distinguishes the component assignments from the vibrational details and contributions of other components at nearby wavenumbers.

\par
  \appHCaption{Literature assignments for the numbered bands in Figures~\ref{fig:m11-nir-spectra} and~\ref{fig:chemical-atlas}.}
  \label{tab:nir-band-assignments}%
}{%
  \begin{tabular}{@{}r l p{.20\appHColumns}p{.34\appHColumns}p{.46\appHColumns}@{}}
  \toprule
  No. & \shortstack[l]{Wavenumber\\(cm$^{-1}$)} & Assigned constituents & Vibration & Interpretation \\
  \midrule
  1 & 8547 & Lignin & Second overtone of asymmetric unsaturated C--H stretching (HC=CH) & Lignin-related C--H band, distinct from the aromatic C--H band at band~10. \\
  \nirassignmentrowrule
  2 & 8420--8370 & Lignin, possible cellulose contribution & Second overtone of C--H stretching & A CH$_3$ assignment has been reported. This band is not specific to lignin. \\
  \nirassignmentrowrule
  3 & 8250--8160 & Cellulose & Second overtone of C--H stretching & Includes several C--H bands. Assignments to individual CH and CH$_2$ groups are tentative. \\
  \nirassignmentrowrule
  4 & 7321--7315 & Cellulose & Combination of the first overtone of C--H stretching and C--H bending & --- \\
  \nirassignmentrowrule
  5 & 7003--6993 & Water, amorphous polysaccharides & First overtone of O--H stretching (polysaccharides) & Free or weakly hydrogen-bonded O--H. Glucomannan and xylan contribute alongside cellulose. \\
  \nirassignmentrowrule
  6 & 6974--6954 & Water, amorphous polysaccharides & Combination of symmetric and asymmetric O--H stretching in water (main water contribution) and first overtone of polysaccharide O--H stretching & Absorption by water and amorphous polysaccharides overlaps. Interpret together with band~5 to assess differences in the hydrogen-bonding environment. \\
  \nirassignmentrowrule
  7 & 6757 & Cellulose (semicrystalline regions) & First overtone of O--H stretching & Weak hydrogen bonding in cellulose and adsorbed water have also been reported to contribute nearby. \\
  \nirassignmentrowrule
  8 & 6450 & Cellulose (crystalline regions) & First overtone of hydrogen-bonded O--H stretching & Nearby assignments to intrachain hydrogen bonding and to cellulose I$\alpha$ overlap. \\
  \nirassignmentrowrule
  9 & 6281 & Cellulose (crystalline regions) & First overtone of strongly hydrogen-bonded O--H stretching & Several intra- and interchain hydrogen bonds have been assigned in this region. \\
  \nirassignmentrowrule
  10 & 5980--5963 & Lignin, aromatic extractives & First overtone of aromatic C--H stretching & CH$_3$ in hemicellulose acetyl groups contributes on either side, near 6003 and 5950\,cm$^{-1}$. \\
  \nirassignmentrowrule
  11 & 5890 & Lignin & First overtone of C--H stretching & Also observed in isolated lignin. The type of C--H group remains uncertain. \\
  \nirassignmentrowrule
  12 & 5865 & Hemicellulose & First overtone of C--H stretching & Assignments to CH$_2$ and to CH$_3$ in xylan acetyl groups have both been reported. A cellulose C--H contribution has been reported nearby at 5872\,cm$^{-1}$. \\
  \nirassignmentrowrule
  13 & 5795 & Lignin & First overtone of C--H stretching & Also observed in isolated lignin. Hemicellulose contributes nearby at about 5800\,cm$^{-1}$. \\
  \nirassignmentrowrule
  14 & 5776 & Cellulose & First overtone of C--H stretching & The assignment to CH$_2$ is tentative. \\
  \nirassignmentrowrule
  15 & 5577 & Cellulose, lignin & First overtone of C--H stretching & In addition to cellulosic paper, isolated lignin shows absorption at 5578\,cm$^{-1}$. \\
  \nirassignmentrowrule
  16 & 5464 & Cellulose (semicrystalline and crystalline regions) & Combination of O--H stretching and the first overtone of C--O stretching & Interpret together with O--H bands~7--9 to assess features associated with crystalline and semicrystalline regions. \\
  \nirassignmentrowrule
  17 & 5220--5150 & Water & Combination of asymmetric O--H stretching and H--O--H bending & Reflects bound water in the cell wall under air-dry conditions. Hydrogen bonding with polymers affects its position and shape. \\
  \nirassignmentrowrule
  18 & 4392 & Cellulose & Combination of O--H and C--C stretching, and/or of C--H stretching and C--H bending & Both combinations have been proposed, leaving the specific vibrational assignment open. \\
  \bottomrule
  \end{tabular}%
}{%
The vertical lines in the figures mark literature wavenumbers associated with the troughs at which we compared changes. Wavenumber ranges encompass values reported in multiple sources.
Overlapping absorption bands and differentiation can affect the position and depth of a trough.
We therefore considered multiple bands associated with each component together, without assuming that neighboring band numbers correspond to independently resolved peaks.
We did not convert these assignments directly into component quantities or degrees of degradation.%
}
\endgroup
\clearpage
\subsection{Specimen Maps and NIR Spectra}
\label{app:specimen-nir-atlas}

\begin{figure}[!ht]
  \centering
  \includegraphics[page=1,width=\linewidth,height=.53\textheight,keepaspectratio]{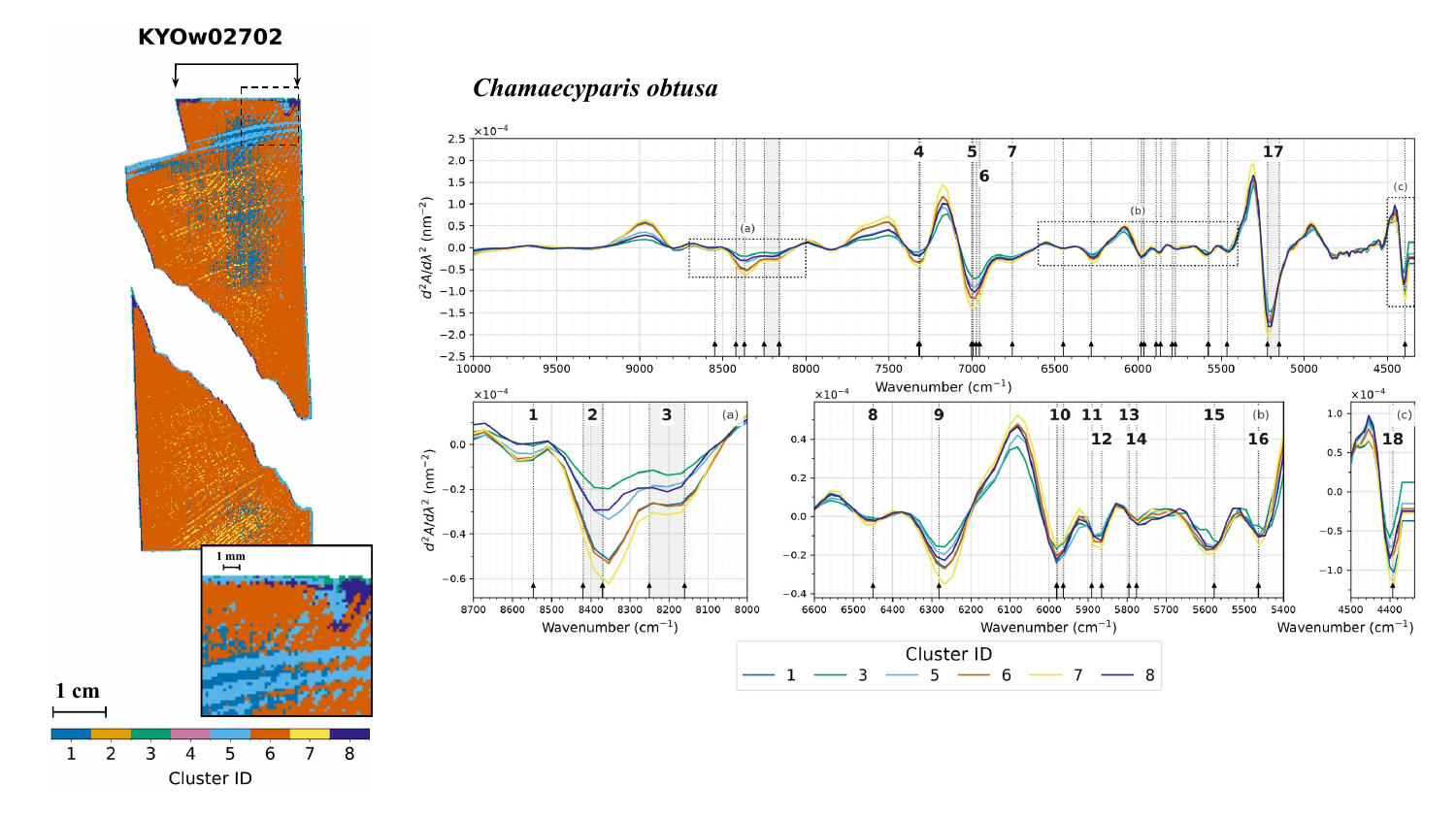}
  \caption{M11 spatial maps and second-derivative NIR spectra for all 49 specimens.
  Each specimen is shown with its map and local enlargement on the left and its spectra on the right.
  Map colors, color-bar numbers, and spectral line colors denote cluster IDs.
  On the measured transverse surfaces, the arrows mark perimeter segments corresponding to side surfaces of the original timber member that were exposed during service.
  Dotted boxes enclose the regions adjacent to those surfaces that are shown enlarged.
  Spectral band numbers 1--18 correspond to Table~\ref{tab:nir-band-assignments}. In the spectral plots, arrows and vertical dotted lines indicate the literature-assigned wavenumbers used for the comparisons.
  Gray bands indicate wavenumber ranges reported in the literature. Regions (a)--(c) are shown enlarged.
  The legend gives the specimen ID when a spectrum from another specimen is included for reference.
  Triangles mark spectral values outside the displayed vertical range.}
  \label{fig:chemical-atlas}
\end{figure}

The maps in Figure~\ref{fig:chemical-atlas} show all pixel assignments from full-dataset fitting of M11 for each of the 49 specimens.
For each specimen, we show second-derivative NIR spectra for clusters whose assigned pixels formed visually discernible regions in its map (Section~\ref{sec:nir-interpretation}), with any additional reference spectra labeled by source specimen.
Section~\ref{sec:results-spectra} relates the spectral features to the spatial distributions.
Appendix~\ref{app:global-map-comparison} compares the spatial maps from full-dataset fitting under the six main conditions.

\newlength{\arxivAtlasWidth}
\newlength{\arxivAtlasPanelWidth}
\newlength{\arxivAtlasPanelHeight}
\newsavebox{\arxivAtlasCaption}
\newcommand{\arxivAppendixAtlasPanel}[1]{%
  \includegraphics[page=\number\numexpr#1\relax,width=\arxivAtlasPanelWidth,
    height=\arxivAtlasPanelHeight,keepaspectratio]{Figure_H.1.pdf}}
\newcommand{\arxivAtlasRow}[2]{%
  \hbox to\arxivAtlasWidth{\hfil\arxivAppendixAtlasPanel{#1}%
    \hskip4pt\arxivAppendixAtlasPanel{#2}\hfil}}
\newcommand{\arxivAppendixAtlasSheet}[2]{%
  \clearpage
  \begingroup
    \sbox{\arxivAtlasCaption}{\footnotesize Figure~\ref{fig:chemical-atlas} (continued, #2/7).}%
    \setlength{\arxivAtlasWidth}{\dimexpr\paperwidth-.6in\relax}%
    \setlength{\arxivAtlasPanelWidth}{\dimexpr(\arxivAtlasWidth-4pt)/2\relax}%
    \setlength{\arxivAtlasPanelHeight}{\dimexpr(\textheight-\ht\arxivAtlasCaption
      -\dp\arxivAtlasCaption-15pt)/4\relax}%
    \noindent\makebox[\linewidth][c]{\vbox{\offinterlineskip
      \arxivAtlasRow{#1}{#1+1}\vskip3pt
      \arxivAtlasRow{#1+2}{#1+3}\vskip3pt
      \arxivAtlasRow{#1+4}{#1+5}\vskip3pt
      \arxivAtlasRow{#1+6}{#1+7}\vskip5pt
      \hbox to\arxivAtlasWidth{\hfil\usebox{\arxivAtlasCaption}\hfil}}}\par
  \endgroup}
\arxivAppendixAtlasSheet{2}{2}
\arxivAppendixAtlasSheet{10}{3}
\arxivAppendixAtlasSheet{18}{4}
\arxivAppendixAtlasSheet{26}{5}
\arxivAppendixAtlasSheet{34}{6}
\arxivAppendixAtlasSheet{42}{7}

\end{document}